\documentclass[11pt,a4paper]{article}
\usepackage[utf8]{inputenc}

\usepackage{mathrsfs,amsthm,xcolor,verbatim,bbm,amsmath,amsfonts,amssymb,nicefrac,enumitem,hyperref,bm,mathtools,xparse,etoolbox}
\usepackage[margin=1in]{geometry}
\usepackage[capitalise]{cleveref} % clever reference
\usepackage{tikz}
\usetikzlibrary{matrix,chains,positioning,decorations.pathreplacing,arrows,shapes,math,arrows.meta,calc, shapes.geometric}

\tikzset{
    neuron/.style={circle, draw, minimum size=0.5cm, text centered},
    input neuron/.style={neuron, fill=yellow!30},
    hidden neuron/.style={neuron, fill=blue!10, draw=blue, thick},
    output neuron/.style={neuron, fill=green!20},
    conn/.style={-stealth, thick},
    ellipsis/.style={font=\footnotesize},
}
\usepackage{capt-of}
\usepackage{stmaryrd} % for double brackets

\crefname{enumi}{item}{items}
\crefname{equation}{}{}
\crefname{subsection}{Subsection}{Subsections}
\crefname{figure}{Figure}{Figures}

\usepackage{xurl}

\hypersetup{
    colorlinks,
    linkcolor={red!80!black},
    citecolor={green},
    urlcolor={blue!80!black}
}

\theoremstyle{plain}
\newtheorem{theorem}{Theorem}[section]

\newtheorem{prop}[theorem]{Proposition}

\newtheorem{remark}[theorem]{Remark}
\newtheorem{definition}[theorem]{Definition}
\newtheorem{setting}[theorem]{Setting}

\theoremstyle{definition}

\DeclareMathAlphabet{\mathpzc}{OT1}{pzc}{m}{it}

\DeclareFontEncoding{LS1}{}{}
\DeclareFontSubstitution{LS1}{stix}{m}{n}
\DeclareMathAlphabet{\mathscr}{LS1}{stixscr}{m}{n}

\renewcommand{\P}{\mathbb{P}}

\newcommand{\R}{\mathbb{R}}
\newcommand{\N}{\mathbb{N}}

\renewcommand{\d}{ \mathrm{d}}

\newcommand{\w}[1]{\mathfrak{w}^{#1}}
\renewcommand{\b}[1]{\mathfrak{b}^{#1}}

\renewcommand{\c}[1]{\mathfrak{c}^{#1}}

\newcommand{\element}{r}
\newcommand{\approximate}{n}

\newcommand{\with}{\curvearrowleft}

\newcommand{\func}{\mathbf{g}}

\newcommand{\cB}{\mathcal{B}}

\newcommand{\cD}{\mathcal{D}}
\newcommand{\cE}{\mathcal{E}}

\newcommand{\cG}{\mathcal{G}}

\newcommand{\cL}{\mathcal{L}}

\newcommand{\cN}{\mathcal{N}}

\newcommand{\bfd}{\mathbf{d}}
\newcommand{\bfe}{\mathbf{e}}

\newcommand{\bfx}{\mathbf{x}}

\newcommand{\bfC}{\mathbf{C}}

\newcommand{\bfE}{\mathbf{E}}

\newcommand{\scrA}{\mathscr{A}}

\newcommand{\activate}{A}
\newcommand{\scra}{\mathscr{a}}

\newcommand{\g}{\|\mathfrak{g}\|_{Lip}}

\newcommand{\fC}{\mathfrak{C}}

\newcommand{\fG}{\mathcal{G}}

\newcommand{\fb}{\mathfrak{b}}

\newcommand{\fd}{\mathfrak{d}}

\newcommand{\fg}{\mathfrak{g}}

\newcommand{\fw}{\mathfrak{w}}

\newcommand{\scrd}{\mathscr{d}}

\newcommand{\ZZ}{z}

\def\mN{\mathcal N}

\renewcommand{\emptyset}{\varnothing}

\newcommand{\const}{\fC}

\newcommand\restr[2]{{% we make the whole thing an ordinary symbol
  \left.\kern-\nulldelimiterspace % automatically resize the bar with \right
  #1 % the function
  \vphantom{|} % pretend it's a little taller at normal size
  \right|_{#2} % this is the delimiter
  }}

\DeclarePairedDelimiter{\norm}{\lVert}{\rVert}
\DeclarePairedDelimiter{\abs}{\lvert}{\rvert}
\DeclarePairedDelimiter{\rbr}{(}{)}
\DeclarePairedDelimiter{\br}{[}{]}
\DeclarePairedDelimiter{\cu}{\{}{\}}
\DeclarePairedDelimiter{\spro}{\langle}{\rangle}

\newcommand{\qandq}{\quad \text{and} \quad }
\newcommand{\qqandqq}{\qquad\text{and}\qquad}

\newcommand{\indicator}[1]{\mathbbm{1}_{\smash{#1}}}

\usepackage{cleveref}

\ExplSyntaxOn

\seq_new:N \g_cflist_loaded
\seq_new:N \g_cflist_pending

\NewDocumentCommand{\cfadd} { m } {
  \seq_if_in:NnF \g_cflist_loaded { #1 } {
    \seq_if_in:NnF \g_cflist_pending { #1 } {
      \seq_gput_right:Nn \g_cflist_pending { #1 }
    }
  }
}

\NewDocumentCommand{\cfconsiderloaded} { m } {
  \seq_gput_right:Nn \g_cflist_loaded {#1}
}

\NewDocumentCommand{\cfremove} { m } {
  \seq_gremove_all:Nn \g_cflist_pending { #1 }
}

\NewDocumentCommand{\cfload} { o } {
  \seq_if_empty:NTF \g_cflist_pending {
    \IfValueTF{#1}{\ignorespaces}{\unskip}
  } {
    (cf.\ \cref{\seq_use:Nn \g_cflist_pending {,}})\IfValueTF{#1}{#1~}{\unskip}
    \seq_gconcat:NNN \g_cflist_loaded \g_cflist_loaded \g_cflist_pending
    \seq_gclear:N \g_cflist_pending
    \IfValueT{#1}{\ignorespaces}
  }
}

\NewDocumentCommand{\cfclear} {} {
  \seq_gclear:N \g_cflist_loaded
  \seq_gclear:N \g_cflist_pending
}

\NewDocumentCommand{\cfout} { o } {
  \seq_if_empty:NTF \g_cflist_pending {\unskip\IfValueT{#1}{\ignorespaces}} {
    (cf.\ \cref{\seq_use:Nn \g_cflist_pending {,}})\IfValueTF{#1}{#1~}{\unskip}
    \seq_gclear:N \g_cflist_pending
    \IfValueT{#1}{\ignorespaces}
  }
}

\NewDocumentCommand{\ifnocf} { m } {
  \seq_if_empty:NT \g_cflist_pending { #1 }
}

\ExplSyntaxOff

\ExplSyntaxOn

\bool_new:N \g_noteobserve

\NewDocumentCommand{\setnote}{}{
  \bool_gset_true:N \g_noteobserve
}

\NewDocumentCommand{\setobserve}{}{
  \bool_gset_false:N \g_noteobserve
}

\NewDocumentCommand{\nobs}{ o }{
  \IfValueT{#1}{
    \str_if_eq:noTF {note} {#1} {
      \bool_gset_true:N \g_noteobserve
    } {
      \str_if_eq:noTF {Note} {#1} {
        \bool_gset_true:N \g_noteobserve
      } {
        \bool_gset_false:N \g_noteobserve
      }
    }
  }
  \bool_if:nTF { \g_noteobserve } {
    \bool_gset_false:N \g_noteobserve
    note
  } {
    \bool_gset_true:N \g_noteobserve
    observe
  }
  \IfValueF{#1}{~}
}

\NewDocumentCommand{\Nobs}{ o }{
  \IfValueT{#1}{
    \str_if_eq:noTF {note} {#1} {
      \bool_gset_true:N \g_noteobserve
    } {
      \str_if_eq:noTF {Note} {#1} {
        \bool_gset_true:N \g_noteobserve
      } {
        \bool_gset_false:N \g_noteobserve
      }
    }
  }
  \bool_if:nTF { \g_noteobserve } {
    \bool_gset_false:N \g_noteobserve
    Note
  } {
    \bool_gset_true:N \g_noteobserve
    Observe
  }
  \IfValueF{#1}{~}
}

\ExplSyntaxOff

\ExplSyntaxOn

\bool_new:N \g_hencetherefore

\NewDocumentCommand{\hence}{ o }{
  \IfValueT{#1}{
    \str_if_eq:noTF {hence} {#1} {
      \bool_gset_true:N \g_hencetherefore
    } {
      \str_if_eq:noTF {Hence} {#1} {
        \bool_gset_true:N \g_hencetherefore
      } {
        \bool_gset_false:N \g_hencetherefore
      }
    }
  }
  \bool_if:nTF { \g_hencetherefore } {
    \bool_gset_false:N \g_hencetherefore
    hence
  } {
    \bool_gset_true:N \g_hencetherefore
    therefore
  }
  \IfValueF{#1}{~}
}

\NewDocumentCommand{\Hence}{ o }{
  \IfValueT{#1}{
    \str_if_eq:noTF {hence} {#1} {
      \bool_gset_true:N \g_hencetherefore
    } {
      \str_if_eq:noTF {Hence} {#1} {
        \bool_gset_true:N \g_hencetherefore
      } {
        \bool_gset_false:N \g_hencetherefore
      }
    }
  }
  \bool_if:nTF { \g_hencetherefore } {
    \bool_gset_false:N \g_hencetherefore
    Hence,~we~obtain
  } {
    \bool_gset_true:N \g_hencetherefore
    Therefore,~we~obtain
  }
  \IfValueF{#1}{~}
}

\ExplSyntaxOff

\ExplSyntaxOn

\seq_const_from_clist:Nn \g_prove_mru {
  establish,
  demonstrate,
  prove,
  show,
  imply,
  ensure
}

\prop_new:N \l__verbs
\prop_put:Nnn \l__verbs {show} {shows}
\prop_put:Nnn \l__verbs {imply} {implies}
\prop_put:Nnn \l__verbs {demonstrate} {demonstrates}
\prop_put:Nnn \l__verbs {prove} {proves}
\prop_put:Nnn \l__verbs {establish} {establishes}
\prop_put:Nnn \l__verbs {ensure} {ensures}
\prop_put:Nnn \l__verbs {assure} {assures}

\tl_new:N \g_wordtmp
\seq_new:N \l_mytmps

\cs_generate_variant:Nn \str_if_in:nnTF { nVTF }
\cs_generate_variant:Nn \str_if_in:nnTF { xVTF }

\NewDocumentCommand{\prove}{ o }{
  \IfValueTF{#1}{
    \seq_clear:N \l_mytmps
    \seq_map_inline:Nn \g_prove_mru {
      \str_if_eq:nnTF {##1} {ensure} {
        \str_set:Nn \l_temps {n}
      } {
        \str_set:Nx \l_temps {\str_head_ignore_spaces:n {##1}}
      }
      \str_if_in:xVTF {#1} \l_temps {
        \seq_put_right:Nn \l_mytmps {##1}
      } { }
    }
    \seq_get_right:NN \l_mytmps \g_wordtmp
  } {
    \seq_get_right:NN \g_prove_mru \g_wordtmp
  }
  \tl_use:N \g_wordtmp
  \IfValueTF{#1}{}{~}
  \seq_gput_left:NV \g_prove_mru \g_wordtmp
  \seq_gremove_duplicates:N \g_prove_mru
}

\NewDocumentCommand{\proves}{ o }{
  \IfValueTF{#1}{
    \seq_clear:N \l_mytmps
    \seq_map_inline:Nn \g_prove_mru {
      \str_if_eq:nnTF {##1} {ensure} {
        \str_set:Nn \l_temps {n}
      } {
        \str_set:Nx \l_temps {\str_head_ignore_spaces:n {##1}}
      }
      \str_if_in:xVTF {#1} \l_temps {
        \seq_put_right:Nn \l_mytmps {##1}
      } { }
    }
    \seq_get_right:NN \l_mytmps \g_wordtmp
  } {
    \seq_get_right:NN \g_prove_mru \g_wordtmp
  }
  \str_set:NV \l_tmpa_str \g_wordtmp
  \prop_get:NVN \l__verbs \l_tmpa_str \l_tmpa_tl
  \tl_use:N \l_tmpa_tl
  \IfValueTF{#1}{}{~}
  \seq_gput_left:NV \g_prove_mru \g_wordtmp
  \seq_gremove_duplicates:N \g_prove_mru
}

\newcommand{\llabel}[1]{\savelabel{#1}\label{\loc.#1}\ignorespaces}

\clist_new:N \l_localreflist
\clist_new:N \l_reflist

\NewDocumentCommand{\lref} { m } {
  \clist_set:No \l_localreflist {#1}
  \clist_clear:N \l_reflist
  \clist_map_inline:Nn \l_localreflist { \clist_put_right:Nn \l_reflist {\loc.##1} }
  \cref{\l_reflist}
}

\NewDocumentCommand{\Lref} { m } {
  \clist_set:No \l_localreflist {#1}
  \clist_clear:N \l_reflist
  \clist_map_inline:Nn \l_localreflist { \clist_put_right:Nn \l_reflist {\loc.##1} }
  \Cref{\l_reflist}
}

\NewDocumentCommand{\itref}{ m m }{
  \clist_set:No \l_localreflist {#2}
  \clist_clear:N \l_reflist
  \clist_map_inline:Nn \l_localreflist { \clist_put_right:Nn \l_reflist {#1.##1} }
  \cref{\l_reflist}~in~\cref{#1}
}

\seq_new:N \l_enum_seq
\int_new:N \l_num_items

\bool_new:N \g_commaused_bool

\providecommand{\comma}{}

\cs_new:Nn \enum_it:nn {
  \int_case:nnF {\l_num_items - #1} {
    {0} {
      \renewcommand{\comma}{}
      #2\space
    }
    {1} {
      \bool_gset_false:N \g_commaused_bool
      \renewcommand{\comma}{,~\bool_gset_true:N \g_commaused_bool}
      #2
      \bool_if:NTF \g_commaused_bool {} {,~}
      and~
    }
  } {
    \bool_gset_false:N \g_commaused_bool
    \renewcommand{\comma}{,~\bool_gset_true:N \g_commaused_bool}
    #2
    \bool_if:NTF \g_commaused_bool {} {,~}
  }
}

\cs_new:Nn \enum_it_U:nn {
  \int_case:nnF {\l_num_items - #1} {
    {0} {
      \renewcommand{\comma}{}
      #2
      \space
    }
    {1} {
      \bool_gset_false:N \g_commaused_bool
      \renewcommand{\comma}{,~\bool_gset_true:N \g_commaused_bool}
      #2
      \bool_if:NTF \g_commaused_bool {} {,~}
      and~
    }
  } {
    \bool_gset_false:N \g_commaused_bool
    \renewcommand{\comma}{,~\bool_gset_true:N \g_commaused_bool}
    \int_compare:nTF {#1=1} {
      \text_titlecase_first:n {#2}
    } {
      #2
    }
    \bool_if:NTF \g_commaused_bool {} {,~}
  }
}

\cs_generate_variant:Nn \tl_if_eq:nnTF {onTF}

\cs_new:Nn \enum:nnnn {
  \seq_set_split:Nnn \l_enum_seq ; {#1}
  \seq_remove_all:Nn \l_enum_seq { }
  \seq_remove_all:Nn \l_enum_seq {#2}
  \seq_log:N \l_enum_seq
  \int_set:Nn \l_num_items {\seq_count:N \l_enum_seq}
  \int_log:N \l_num_items
  \int_case:nnF {\l_num_items} {
    { 0 } { 0 }
    { 1 } {
      \IfBooleanTF{#4} {
        \tl_set:Nn \l_text_case_exclude_arg_tl {\cref}
        \bool_set_false:N \l_text_titlecase_check_letter_bool
        \text_titlecase_first:n {\seq_use:Nn \l_enum_seq {}}
      } {
        \seq_use:Nn \l_enum_seq {}
      }
      \space
      \tl_if_eq:onTF{#3}{-}{}{
        \bool_if:NTF \l_plural_bool {
          \prove[#3]~
        } {
          \proves[#3]~
        }
      }
    }
    { 2 } {
      \IfBooleanTF{#4} {
        \tl_set:Nn \l_text_case_exclude_arg_tl {\cref}
        \bool_set_false:N \l_text_titlecase_check_letter_bool
        \text_titlecase_first:n {\seq_item:Nn \l_enum_seq {1}}
        {} ~and~
        \seq_item:Nn \l_enum_seq {2}
      } {
        \seq_use:Nn \l_enum_seq {~and~}
      }
      \space
      \tl_if_eq:onTF{#3}{-}{}{
        \prove[#3]~
      }
    }
  } {
    \IfBooleanTF{#4} {
      \tl_set:Nn \l_text_case_exclude_arg_tl {\cref}
      \seq_indexed_map_function:NN \l_enum_seq \enum_it_U:nn
    } {
      \seq_indexed_map_function:NN \l_enum_seq \enum_it:nn
    }
    \tl_if_eq:onTF{#3}{-}{}{
      \prove[#3]~
    }
  }
}

\cs_generate_variant:Nn \enum:nnnn {nxnn}
\cs_generate_variant:Nn \enum:nnnn {nxxn}

\NewDocumentCommand{\enum}{O{} m O{-} s}{
  \IfBooleanTF{#4}{
    \enum:nxnn {#2} {#1} {sindep} \BooleanFalse
  } {
    \enum:nxxn {#2} {#1} {#3} \BooleanFalse
  }
}

\NewDocumentCommand{\dott}{}{\ifnocf{.}\space}

\bool_new:N \g_arg_start_bool
\bool_gset_true:N \g_arg_start_bool

\NewDocumentCommand{\startnewargseq}{}{\bool_gset_true:N \g_arg_start_bool \tl_set:Nn \g_label_tl {}}

\cs_generate_variant:Nn \seq_if_in:NnTF {NxTF}
\cs_generate_variant:Nn \seq_remove_all:Nn {Nx}

\int_new:N \l_random_int

\cs_generate_variant:Nn \tl_if_head_eq_catcode:nNTF {oNTF}
\cs_generate_variant:Nn \tl_if_head_eq_catcode:nNTF {VNTF}

\cs_generate_variant:Nn \tl_if_head_eq_catcode:nNTF {eNTF}
\cs_generate_variant:Nn \tl_log:n {o}
\cs_generate_variant:Nn \tl_log:n {f}
\cs_generate_variant:Nn \tl_log:n {x}
\cs_generate_variant:Nn \tl_log:n {e}

\cs_generate_variant:Nn \tl_if_in:nnTF {onTF}
\cs_generate_variant:Nn \tl_if_in:NnTF {NeTF}

\cs_generate_variant:Nn \tl_if_head_eq_meaning:nNTF {VNTF}

\seq_const_from_clist:Nn \g_arg_mru_this {
  Ahpr,
  Tapr,
  Ctapr,
  H
}

\seq_const_from_clist:Nn \g_arg_mru_nothis {
  Ia,
  Nwc,
  N,
  Itns,
  Fm,
  Itnswc,
  Mo
}

\seq_new:N \l_arg_seq
\tl_new:N \l_cons_tl
\tl_new:N \l_dummy_tl

\bool_new:N \g_debug_bool
\bool_gset_false:N \g_debug_bool

\bool_new:N \l_insidearg_bool

\bool_new:N \g_firstargletter_bool

\sys_gset_rand_seed:n {0903}

\bool_new:N \l_plural_bool
\tl_new:N \l_arg_verbs_tl

\NewDocumentCommand{\argument}{mom}{
\color{black}
  \bool_set_false:N \l_plural_bool
  \tl_set:Nn \l_arg_verbs_tl {sindep}
  \keys_define:nn { benno/argument } {
    plural .value_forbidden:n = true,
    plural .code:n = {\bool_set_true:N \l_plural_bool},
    verbs .value_required:n = false,
    verbs .tl_set:N = \l_arg_verbs_tl,
  }
  \IfValueT{#2}{
    \keys_set:nn { benno/argument } {#2}
  }
  \bool_log:N \l_plural_bool
  \bool_gset_true:N \l_insidearg_bool
  \seq_set_split:Nnn \l_arg_seq ; {#1}
  \seq_remove_all:Nn \l_arg_seq { }
  \seq_log:N \l_arg_seq
  \tl_set:Nn \l_cons_tl {#3}
  \tl_trim_spaces:N \l_cons_tl
  \seq_if_in:NxTF \l_arg_seq {\lref{\g_label_tl}} {
    \seq_remove_all:Nx \l_arg_seq {\lref{\g_label_tl}}
    \seq_get_left:NNTF \l_arg_seq \l_dummy_tl {
      \tl_trim_spaces:N \l_dummy_tl
      \bool_gset_false:N \g_firstargletter_bool
      \tl_if_head_eq_catcode:VNTF \l_dummy_tl a {
        \bool_gset_true:N \g_firstargletter_bool
      } {
        \tl_if_head_eq_meaning:VNTF \l_dummy_tl {\cref} {
          \tl_set:Nx \l_tmpa_tl {\tl_tail:N \l_dummy_tl}
          \tl_set:Nx \l_tmpb_tl {\tl_head:N \l_tmpa_tl}
          \bool_gset_true:N \g_firstargletter_bool
          \tl_if_in:NeTF \l_tmpb_tl {lem\c_colon_str} {} {
            \tl_if_in:NeTF \l_tmpb_tl {thm\c_colon_str} {} {
              \tl_if_in:NeTF \l_tmpb_tl {prop\c_colon_str} {} {
                \tl_if_in:NeTF \l_tmpb_tl {cor\c_colon_str} {} {
                  \bool_gset_false:N \g_firstargletter_bool
                }
              }
            }
          }
        } {
        }
      }
      \bool_if:NTF \g_firstargletter_bool {
        \seq_set_eq:NN \l_tmpa_seq \g_arg_mru_this
        \seq_remove_all:Nn \l_tmpa_seq {H}
        \seq_get_right:NN \l_tmpa_seq \l_tmpa_tl
        \int_case:nnF {\seq_count:N \l_arg_seq} {
          {1} {
            \str_case:VnF {\l_tmpa_tl} {
              {Ahpr} {
                \bool_if:NT \g_debug_bool {C1.1}
                \seq_gput_left:Nn \g_arg_mru_this {Ahpr}
                \seq_gremove_duplicates:N \g_arg_mru_this
                \enum:nxnn {#1} {\lref{\g_label_tl}} {-} {\BooleanTrue}
                \hence~
                \bool_if:NTF \l_plural_bool {
                  \prove[\l_arg_verbs_tl]~\ignorespaces #3
                } {
                  \proves[\l_arg_verbs_tl]~\ignorespaces #3
                }
              }
              {Tapr} {
                \bool_if:NT \g_debug_bool {C1.2}
                \seq_gput_left:Nn \g_arg_mru_this {Tapr}
                \seq_gremove_duplicates:N \g_arg_mru_this
                \enum[\lref{\g_label_tl}]{
                  This;
                  #1
                }[\l_arg_verbs_tl]\ignorespaces #3
              }
              {Ctapr} {
                \bool_if:NT \g_debug_bool {C1.3}
                \seq_gput_left:Nn \g_arg_mru_this {Ctapr}
                \seq_gremove_duplicates:N \g_arg_mru_this
                Combining~
                \enum[\lref{\g_label_tl}]{
                  this;
                  #1
                } \proves[\l_arg_verbs_tl]~\ignorespaces #3
              }
            } {}
          }
        } {
          \str_case:VnF {\l_tmpa_tl} {
             {Ahpr} {
              \bool_if:NT \g_debug_bool {C2.1}
              \seq_gput_left:Nn \g_arg_mru_this {Ahpr}
              \seq_gremove_duplicates:N \g_arg_mru_this
              \enum:nxnn {#1} {\lref{\g_label_tl}} {-} {\BooleanTrue}
              \hence~
              \prove[\l_arg_verbs_tl]~\ignorespaces #3
            }
            {Tapr} {
              \bool_if:NT \g_debug_bool {C2.2}
              \seq_gput_left:Nn \g_arg_mru_this {Tapr}
              \seq_gremove_duplicates:N \g_arg_mru_this
              \enum[\lref{\g_label_tl}]{
                This;
                #1
              }[\l_arg_verbs_tl]\ignorespaces #3
            }
            {Ctapr} {
              \int_case:nn {\int_rand:nn {0} {1}} {
                {0} {
                  \bool_if:NT \g_debug_bool {C2.3}
                  \seq_gput_left:Nn \g_arg_mru_this {Ctapr}
                  \seq_gremove_duplicates:N \g_arg_mru_this
                  Combining~
                  \enum[\lref{\g_label_tl}]{
                    this;
                    #1
                  } \proves[\l_arg_verbs_tl]~\ignorespaces #3
                }
                {1} {
                  \bool_if:NT \g_debug_bool {C2.4}
                  \seq_gput_left:Nn \g_arg_mru_this {Ctapr}
                  \seq_gremove_duplicates:N \g_arg_mru_this
                  Combining~
                  \enum:nxnn {#1} {\lref{\g_label_tl}} {-} {\BooleanFalse}
                  \hence~
                  \proves[\l_arg_verbs_tl]~\ignorespaces #3
                }
              }
            }
          } {}
        }
      } {
        \seq_set_eq:NN \l_tmpa_seq \g_arg_mru_this
        \seq_remove_all:Nn \l_tmpa_seq {H}
        \seq_remove_all:Nn \l_tmpa_seq {Ahpr}
        \seq_get_right:NN \l_tmpa_seq \l_tmpa_tl
        \int_case:nnF {\seq_count:N \l_arg_seq} {
          {1} {
            \str_case:VnF {\l_tmpa_tl} {
              {Tapr} {
                \bool_if:NT \g_debug_bool {C3.1}
                \seq_gput_left:Nn \g_arg_mru_this {Tapr}
                \seq_gremove_duplicates:N \g_arg_mru_this
                \enum[\lref{\g_label_tl}]{
                  This;
                  #1
                }[\l_arg_verbs_tl]\ignorespaces #3
              }
              {Ctapr} {
                \bool_if:NT \g_debug_bool {C3.2}
                \seq_gput_left:Nn \g_arg_mru_this {Ctapr}
                \seq_gremove_duplicates:N \g_arg_mru_this
                Combining~
                \enum[\lref{\g_label_tl}]{
                  this;
                  #1
                } \proves[\l_arg_verbs_tl]~\ignorespaces #3
              }
            } {}
          }
        } {
          \str_case:VnF {\l_tmpa_tl} {
            {Tapr} {
              \bool_if:NT \g_debug_bool {C4.1}
              \seq_gput_left:Nn \g_arg_mru_this {Tapr}
              \seq_gremove_duplicates:N \g_arg_mru_this
              \enum[\lref{\g_label_tl}]{
                This;
                #1
              }[\l_arg_verbs_tl]\ignorespaces #3		
            }
            {Ctapr} {
              \int_case:nn {\int_rand:nn {0} {1}} {
                {0} {
                  \bool_if:NT \g_debug_bool {C4.2}
                  \seq_gput_left:Nn \g_arg_mru_this {Ctapr}
                  \seq_gremove_duplicates:N \g_arg_mru_this
                  Combining~
                  \enum[\lref{\g_label_tl}]{
                    this;
                    #1
                  } \proves[\l_arg_verbs_tl]~\ignorespaces #3		
                }
                {1} {
                  \bool_if:NT \g_debug_bool {C4.3}
                  \seq_gput_left:Nn \g_arg_mru_this {Ctapr}
                  \seq_gremove_duplicates:N \g_arg_mru_this
                  Combining~
                  \enum:nxnn {#1} {\lref{\g_label_tl}} {-} {\BooleanFalse}
                  \hence~
                  \proves[\l_arg_verbs_tl]~\ignorespaces #3    
                }
              }
            }
          } {}
        }
      }
    } {
      \tl_if_head_eq_catcode:oNTF \l_cons_tl a {
        \seq_set_eq:NN \l_tmpa_seq \g_arg_mru_this
        \seq_remove_all:Nn \l_tmpa_seq {Ctapr}
        \seq_remove_all:Nn \l_tmpa_seq {Ahpr}
        \seq_get_right:NN \l_tmpa_seq \l_tmpa_tl
        \str_case:VnF {\l_tmpa_tl} {
          {H} {
            \bool_if:NT \g_debug_bool {C5.1}
            \seq_gput_left:Nn \g_arg_mru_this {H}
            \seq_gremove_duplicates:N \g_arg_mru_this
            Hence,~we~obtain~\ignorespaces #3
          }
          {Tapr} {
            \bool_if:NT \g_debug_bool {C5.2}
            \seq_gput_left:Nn \g_arg_mru_this {Tapr}
            \seq_gremove_duplicates:N \g_arg_mru_this
            This~\proves[\l_arg_verbs_tl]~\ignorespaces #3
          }
        } {}
      } {
        \bool_if:NT \g_debug_bool {C6.1}
        \seq_gput_left:Nn \g_arg_mru_this {Tapr}
        \seq_gremove_duplicates:N \g_arg_mru_this
        This~\proves[\l_arg_verbs_tl]~\ignorespaces #3
      }
    } 
  } {
    \int_compare:nNnTF {\seq_count:N \l_arg_seq} = {0} {
      \bool_if:NTF \g_arg_start_bool {
        \bool_if:NT \g_debug_bool {C7.1}
        \Nobs\unskip
        #3
      } {
        \bool_if:NT \g_debug_bool {C7.2}
        \Moreover~
        #3
      }
    } {
      \bool_if:NTF \g_arg_start_bool {
        \bool_if:NT \g_debug_bool {C8.1}
        \tl_log:N \l_arg_verbs_tl
        \Nobs~that~
        \enum{
          #1
        }[\l_arg_verbs_tl]\ignorespaces #3
      } {
        \int_compare:nNnTF {\seq_count:N \l_arg_seq} = {1} {
          \seq_set_eq:NN \l_tmpa_seq \g_arg_mru_nothis
          \seq_remove_all:Nn \l_tmpa_seq {Nwc}
          \seq_remove_all:Nn \l_tmpa_seq {Itnswc}
          \seq_get_right:NN \l_tmpa_seq \l_tmpa_tl
        } {
          \seq_get_right:NN \g_arg_mru_nothis \l_tmpa_tl
        }
        \str_case:VnF {\l_tmpa_tl} {
          {Mo} {
            \bool_if:NT \g_debug_bool {C9.1}
            \seq_gput_left:Nn \g_arg_mru_nothis {Mo}
            \seq_gremove_duplicates:N \g_arg_mru_nothis
            Moreover,~\nobs~that~
            \enum{
              #1
            }[\l_arg_verbs_tl]\ignorespaces #3		
          }
          {Fm} {
            \bool_if:NT \g_debug_bool {C9.2}
            \seq_gput_left:Nn \g_arg_mru_nothis {Fm}
            \seq_gremove_duplicates:N \g_arg_mru_nothis
            Furthermore,~\nobs~that~
            \enum{
              #1
            }[\l_arg_verbs_tl]\ignorespaces #3		
          }
          {Ia} {
            \bool_if:NT \g_debug_bool {C9.3}
            \seq_gput_left:Nn \g_arg_mru_nothis {Ia}
            \seq_gremove_duplicates:N \g_arg_mru_nothis
            In~addition,~\nobs~that~
            \enum{
              #1
            }[\l_arg_verbs_tl]\ignorespaces #3		
          }
          {N} {
            \bool_if:NT \g_debug_bool {C9.4}
            \seq_gput_left:Nn \g_arg_mru_nothis {N}
            \seq_gremove_duplicates:N \g_arg_mru_nothis
            Next,~\nobs~that~
            \enum{
              #1
            }[\l_arg_verbs_tl]\ignorespaces #3		
          }
          {Itns} {
            \bool_if:NT \g_debug_bool {C9.5}
            \seq_gput_left:Nn \g_arg_mru_nothis {Itnswc}
            \seq_gput_left:Nn \g_arg_mru_nothis {Itns}
            \seq_gremove_duplicates:N \g_arg_mru_nothis
            In~the~next~step~we~\nobs~that~
            \enum{
              #1
            }[\l_arg_verbs_tl]\ignorespaces #3		
          }
          {Nwc} {
            \bool_if:NT \g_debug_bool {C9.6}
            \seq_gput_left:Nn \g_arg_mru_nothis {Nwc}
            \seq_gremove_duplicates:N \g_arg_mru_nothis
            Next~we~combine~
            \enum{
              #1
            }to~obtain~\ignorespaces #3
          }
          {Itnswc} {
            \bool_if:NT \g_debug_bool {C9.7}
            \seq_gput_left:Nn \g_arg_mru_nothis {Itns}
            \seq_gput_left:Nn \g_arg_mru_nothis {Itnswc}
            \seq_gremove_duplicates:N \g_arg_mru_nothis
            In~the~next~step~we~combine~
            \enum{
              #1
            }to~obtain~\ignorespaces #3
          }
        } {}
      }
    }
  }
  \bool_gset_false:N \g_arg_start_bool
  \bool_gset_false:N \l_insidearg_bool
  \cfload[.]%~\unskip
  \color{black}
}

\tl_new:N \g_label_tl
\tl_gset:Nn \g_label_tl { }

\NewDocumentCommand{\savelabel}{m}{
  \bool_if:NTF \l_insidearg_bool {
    \tl_gset:Nn \g_label_tl {#1}
  } {
    \tl_gset:Nn \g_label_tl { }
  }
}

\ExplSyntaxOff

\ExplSyntaxOn

\NewDocumentEnvironment {athm} {m m o} {
\str_if_eq:noTF {example} {#1} {
  \bool_gset_true:N \g_example_bool
} {
  \bool_gset_false:N \g_example_bool
}
\cfclear
\IfNoValueTF{#3}{
\begin{#1}\label{#2}\global\def\loc{#2}
}{
\begin{#1}[#3]\label{#2}\global\def\loc{#2}
}
}{
\end{#1}
}

\NewDocumentEnvironment {adef} {m} {
\begin{definition}\label{#1}\global\def\loc{#1}
}{
\end{definition}
}

\NewDocumentEnvironment{aproof} {} {
\bool_if:NTF \g_example_bool {
  \bool_gset_true:N \g_arg_start_bool
  \begin{proof}[Proof~for~\cref{\loc}]
} {
  \bool_gset_true:N \g_arg_start_bool
  \begin{proof}[Proof~of~\cref{\loc}]
}
\bool_gset_false:N \g_finishproof_bool
}{
\bool_if:NTF \g_finishproof_bool {}
{\finishproofthus}
\end{proof}
}

\NewDocumentCommand{\finishproofthus} {} {
  \bool_gset_true:N \g_finishproof_bool 
  \bool_if:NTF \g_example_bool {
    The~proof~for~\cref{\loc}~is~thus~complete.
  } {
    The~proof~of~\cref{\loc}~is~thus~complete.
  }
}
\NewDocumentCommand{\finishproofthis} {} {
  \bool_gset_true:N \g_finishproof_bool 
  \bool_if:NTF \g_example_bool {
    This~completes~the~proof~for~\cref{\loc}.
  } {
    This~completes~the~proof~of~\cref{\loc}.
  }
}

\ExplSyntaxOff

\NewDocumentEnvironment{cproof}{m}
{\begin{proof}[Proof of \cref{#1}]}%
{\noindent The proof of \cref{#1} is thus complete.
\end{proof}}

\NewDocumentEnvironment{cproof2}{m}
{\begin{proof}[Proof of \cref{#1}]}%
{\noindent This completes the proof of \cref{#1}.
\end{proof}}

\hypersetup{
    colorlinks,
    linkcolor={blue!80!black},
    citecolor={green},
    urlcolor={blue!80!black}
}

\makeatletter
\ExplSyntaxOn
\seq_new:N \g_abbrs
\prop_new:N \g_abbr_counts
\tl_new:N \l_abbr_count_tl

\ExplSyntaxOn

\bool_new:N \g_forexample

\NewDocumentCommand{\eg}{ o }{
	\IfValueT{#1}{
		\str_if_eq:noTF {fe} {#1} {
			\bool_gset_true:N \g_forexample
		} {\bool_gset_false:N \g_forexample}
	}
	\bool_if:nTF { \g_forexample } {
		\bool_gset_false:N \g_forexample
		for~example
	}{
		\bool_gset_true:N \g_forexample
		for~instance
	}
}

\NewDocumentCommand{\abbr}{m m O{#1} m m O{#4} m}{
	\expandafter\newcommand\csname#3\endcsname[1][]{
		\seq_if_in:NnTF \g_abbrs {#1} {
			\prop_get:NnN \g_abbr_counts {#1} \l_abbr_count_tl
			\prop_gput:Nnx \g_abbr_counts {#1} {\int_eval:n {\l_abbr_count_tl + 1}}
			\hyperref[#1]{#7}
		} {
			\seq_gput_left:Nn \g_abbrs {#1}
			\prop_gput:Nnn \g_abbr_counts {#1} {1}
			\expandafter\gdef\csname#1@def\endcsname{#2}
			\phantomsection\label{#1}
			\str_if_eq:nnTF{##1}{}{\emph{#2}}{##1}~(\hyperref[#1]{#7})
		}
	}
	\expandafter\newcommand\csname#6\endcsname[1][]{
		\seq_if_in:NnTF \g_abbrs {#1} {
			\prop_get:NnN \g_abbr_counts {#1} \l_abbr_count_tl
			\prop_gput:Nnx \g_abbr_counts {#1} {\int_eval:n {\l_abbr_count_tl + 1}}
			\hyperref[#1]{#4}
		} {
			\expandafter\gdef\csname#1@def\endcsname{#5}
			\seq_gput_left:Nn \g_abbrs {#1}
			\prop_gput:Nnn \g_abbr_counts {#1} {1}
			\phantomsection\label{#1}
			\str_if_eq:nnTF{##1}{}{\emph{#5}}{##1}~(\hyperref[#1]{#4})
		}
	}
}

\ExplSyntaxOff
\makeatother
\abbr{GD}{gradient descent}{GDs}{gradient descents}{GD}
\abbr{ANN}{artificial neural network}{ANNs}{artificial neural networks}{ANN}
\abbr{DNN}{deep artificial neural network}{DNNs}{deep artificial neural networks}{DNN}
\abbr{SGD}{stochastic gradient descent}{SGDs}{stochastic gradient descent}{SGD}
\abbr{iid}{independent and identically distributed}{i.i.d}{independent and identically distributed}{i.i.d.}
\abbr{DL}{Deep learning}{DLs}{Deep learning}{DL}
\abbr{AI}{artificial intelligence}{AIs}{artificial intelligence}{AI}
\abbr{LLM}{large language model}{LLMs}{large language models}{LLM}
\abbr{PDE}{partial differential equation}{PDEs}{partial differential equations}{PDE}
\abbr{as}{almost surely}{ass}{almost surely}{a.s.}
\abbr{pas}{$\P$-almost surely}{pass}{almost surely}{$\P$-a.s.}
\abbr{ReLUs}{rectified linear unit}{ReLU}{rectified linear unit}
\makeatother

\title{
Mathematical analysis of the gradients in
deep learning
}
\author{Steffen Dereich$^{1}$, Thang Do$^{2,3}$, Arnulf Jentzen$^{4,5}$, and Frederic Weber$^{6}$
	\bigskip
	\\
	\small{$^1$ Institute for Mathematical Stochastics, Faculty of Mathematics and Computer Science,}\vspace{-0.1cm}\\
\small{University of M\"unster, Germany, e-mail: \texttt{steffen.dereich@uni-muenster.de}}
\smallskip
\\
    	\small{$^2$ School of Data Science, The Chinese University of Hong Kong, Shenzhen}
	\vspace{-0.1cm}\\
	\small{ (CUHK-Shenzhen), China, e-mail: \texttt{minhthangdo@link.cuhk.edu.cn}}
 \smallskip
	\\
 \small{$^3$ Deparment of Probability and Statistic, Institute of Mathematics,}
	\vspace{-0.1cm}\\
	\small{Vietnam Academy of Science and Technology, Vietnam, e-mail: \texttt{dmthang@math.ac.vn}}
	\smallskip
	\\
	\small{$^4$ School of Data Science and Shenzhen Research Institute of Big Data, The Chinese University}
	\vspace{-0.1cm}\\
	\small{of Hong Kong, Shenzhen (CUHK-Shenzhen), China, e-mail: \texttt{ajentzen@cuhk.edu.cn}}
	\smallskip
	\\
 \small{$^5$ Applied Mathematics: Institute for Analysis and Numerics, Faculty of Mathematics and}
	\vspace{-0.1cm}\\
	\small{Computer Science, University of M{\"u}nster, Germany, e-mail: \texttt{ajentzen@uni-muenster.de}}
	\smallskip
	\\
    \small{$^6$ Applied Mathematics: Institute for Analysis and Numerics, Faculty of Mathematics and}
	\vspace{-0.1cm}\\
	\small{Computer Science, University of M{\"u}nster, Germany, e-mail: \texttt{frederic.weber@alumni.uni-ulm.de}}
	\smallskip
	\\
}

\date{\today}
\allowdisplaybreaks 
\begin{document}
\maketitle
\begin{abstract}
    Deep learning algorithms -- typically consisting of a class of deep artificial neural networks (ANNs) trained by a stochastic gradient descent (SGD) optimization method -- are nowadays an integral part in many areas of science, industry, and also our day to day life. Roughly speaking, in their most basic form, ANNs can be regarded as functions that consist of a series of compositions of affine-linear functions with multidimensional versions of so-called activation functions. One of the most popular of such activation functions is the rectified linear unit (ReLU) function $\R \ni x \mapsto \max\{ x, 0 \} \in \R$. The ReLU function is, however, not differentiable and, typically, this lack of regularity transfers to the cost function of the supervised learning problem under consideration. Regardless of this lack of differentiability issue, deep learning practioners apply SGD methods based on suitably generalized gradients in standard deep learning libraries like {\sc TensorFlow} or {\sc Pytorch}. In this work we reveal an accurate and concise mathematical description of such generalized gradients in the training of deep fully-connected feedforward ANNs and we also study the resulting generalized gradient function analytically. Specifically, we provide an appropriate approximation procedure that uniquely describes the generalized gradient function, we prove that the generalized gradients are limiting Fréchet subgradients of the cost functional, and we conclude that the generalized gradients must coincide with the standard gradient of the cost functional on every open sets on which the cost functional is continuously differentiable.
\end{abstract}
\pagebreak
\tableofcontents
\newpage
\section{Introduction}
 Deep learning algorithms are nowadays an integral part in many areas of science, industry, and also our day to day life. In particular, the training of \ANNs\ is highly important for applications like, for example, autonomous driving, face recognition, language translation, and credit scoring.
In supervised learning problems, the aim of this training procedure is to minimize the error that the \ANN\ makes in estimating the output for the given input. This error is measured by using a so-called loss function and optimization algorithms that are based on \SGD\ methods with the normalized summed loss function, sometimes also referred to as cost function, as the objective function are most commonly used in the training of \ANNs\ (cf., \eg, the survey articles and monographs \cite{bach2024learning,Cucker2001OnTM,WeChaoSteLei2020,ArBePhi2024,PhilipJa2024,ruder2017overviewgradientdescentoptimization,Sun2019OptimizationFD}).

Roughly speaking, in their most basic form, \ANNs\ can be seen as a series of compositions
of affine-linear functions with multidimensional versions of so-called activation functions. 
One of the most popular of such activation functions is the \ReLU\ function
\begin{equation}
\R \ni x \mapsto \max\{x,0\} \in \R.
\end{equation}
The \ReLU\ function is, however, not differentiable and, typically, this lack of regularity transfers to the cost functional of the supervised learning problem under consideration. Regardless of this lack of differentiability issue, deep learning practioners apply SGD methods based on suitably generalized gradients in standard deep learning libraries like {\sc TensorFlow} or {\sc Pytorch} (cf., \eg, \cite{TensorFlow2016,Adam2017automatic,pytorch2019}) even when they employ \ANNs\ with non-differentiable activation functions such as the popular \ReLU\ function and, for example, the leaky \ReLU\ activation function $ \R \ni x \mapsto \max\{\gamma x,x\} \in \R $ with leaky parameter $ \gamma \in [0,1]$.

In this work we reveal an accurate and concise mathematical description of such generalized gradients in the training 
of deep fully-connected feedforward \ANNs\ and we also study the resulting generalized gradient function analytically. Specifically, in \cref{Theorem X} in \cref{sec: regularity} we provide an appropriate approximation procedure that uniquely describes the generalized gradient function (see \cref{item 1: Theorem X} in \cref{Theorem X} and also \cref{thm:mainthm_sequence: part 1} in \cref{sec: explicit representation}), we prove that the generalized gradients are limiting Fr\'{e}chet subgradients (see \cref{item 2: Theorem X} in \cref{Theorem X}), and we conclude that the generalized gradients must coincide with the standard/usual gradient/derivative of the cost functional on open sets on which the cost functional is continuously differentiable (see \cref{item 3: Theorem X} in \cref{Theorem X}). The specific approximation approach that we employ to specify the generalized gradient function (see \cref{item 1: Theorem X} in \cref{Theorem X}) is based on the approximation approaches in \cite{CheriditoJentzenRiekert2021,ScarpaMultiLayers,ArAd2022aproofofconvergence} in which, loosely speaking, specialized and weakened variants of the considered approximation procedure have been proposed and used. Moreover, our proofs of \cref{item 2: Theorem X,item 3: Theorem X} in \cref{Theorem X} are based on the arguments in \cite[Section 3]{Aradexistenceofglobmin} in which special cases of \cref{item 2: Theorem X,item 3: Theorem X} in \cref{Theorem X} have been established. We also refer to \cref{subsec: literature} below for a more detailed review of related research findings in the scientific literature.

\begin{figure}[ht!] 
	\centering
	\newcommand{\layerdistance}{6}
	\newcommand{\neurondistance}{3.0}
	\newcommand{\ANNInput}{3}
	\newcommand{\ANNHidden}{5}
	\newcommand{\ANNHiddenTwo}{4} % Second hidden layer neurons
	\newcommand{\ANNOutput}{3}
    \resizebox{\textwidth}{!}{
	\begin{tikzpicture}[shorten >=2pt, -latex, draw=black!50, auto]
		\tikzstyle{neuron}=[circle, draw=black!75, minimum size=80pt, inner sep=2pt, thick, align=center];
		\tikzstyle{neuron0}=[neuron, draw={rgb:black,10;black,0}, fill={rgb:yellow,1;white,5}, thick];
		\tikzstyle{neuron1}=[neuron, draw={rgb:purple,10;black,0}, fill={rgb:blue,0.3;white,5}, ultra thick];
		\tikzstyle{neuron2}=[neuron, draw={rgb:black,10;black,0}, fill={rgb:green,1;white,5}, thick];
				\tikzstyle{arrow} = [thick,->,>=stealth, line width=1.2pt];

		\tikzstyle{arrow0} = [arrow, black]
		
		% Input layer
		\foreach \i in {1,...,\ANNInput}{
			\tikzmath{
				\y = (\ANNInput-1)*\neurondistance/2 - (\i-1)*\neurondistance;
			}
			\node[neuron0] (layer-0-neuron-\i) at ( 0, \y ) {\fontsize{9pt}{9pt}\selectfont Input layer\\ \fontsize{9pt}{9pt}\selectfont Neuron $ \i $};
		}
		
		% Hidden layer 1
		\foreach \i in {1,...,\ANNHidden}{
			\tikzmath{
				\y = (\ANNHidden-1)*\neurondistance/2 - (\i-1)*\neurondistance;
			}
			\node[neuron1] (layer-1-neuron-\i) at ( \layerdistance, \y ) {\fontsize{9pt}{9pt}\selectfont Hidden layer 1\\ \fontsize{9pt}{9pt}\selectfont Neuron $ \i $};
		}
		
		% Hidden layer 2
		\foreach \i in {1,...,\ANNHiddenTwo}{
			\tikzmath{
				\y = (\ANNHiddenTwo-1)*\neurondistance/2 - (\i-1)*\neurondistance;
			}
			\node[neuron1] (layer-2-neuron-\i) at ( 2*\layerdistance, \y ) {\fontsize{9pt}{9pt}\selectfont Hidden layer 2\\ \fontsize{9pt}{9pt}\selectfont Neuron $ \i $};
		}
		
		% Output layer
		\foreach \i in {1,...,\ANNOutput}{
			\tikzmath{
				\y = (\ANNOutput-1)*\neurondistance/2 - (\i-1)*\neurondistance;
			}
			\node[neuron2] (layer-3-neuron-\i) at ( 3*\layerdistance, \y ) {\fontsize{9pt}{9pt}\selectfont Output layer\\\fontsize{9pt}{9pt}\selectfont Neuron $ \i $};
		}
		
		% Connections
		\foreach \i in {1,...,\ANNInput}{
			\foreach \j in {1,...,\ANNHidden}{
				\draw [arrow0,arrows = {-Latex[width=10pt, length=10pt]}] (layer-0-neuron-\i) -- (layer-1-neuron-\j);
			}
		}
		\foreach \i in {1,...,\ANNHidden}{
			\foreach \j in {1,...,\ANNHiddenTwo}{
				\draw [arrow0,arrows = {-Latex[width=10pt, length=10pt]}] (layer-1-neuron-\i) -- (layer-2-neuron-\j);
			}
		}
		\foreach \i in {1,...,\ANNHiddenTwo}{
			\foreach \j in {1,...,\ANNOutput}{
				\draw [arrow0,arrows = {-Latex[width=10pt, length=10pt]}] (layer-2-neuron-\i) -- (layer-3-neuron-\j);
			}
		}
	\end{tikzpicture}
    }
    \captionof{figure}{Graphical illustration of an \ANN\ architecture with $L = 3$ affine transformations, $L - 1 = 2$ hidden layers, and $L + 1 = 4$ layers overall (with  $\ell_0 = 3$ neurons on the input layer, $\ell_1 =5$ neurons on the $1\textsuperscript{st}$ hidden layer, $\ell_2 = 4$ neurons on the $2\textsuperscript{nd}$ hidden layer, and $\ell_3 =3$ neurons on the output layer) within the class of \ANNs\ considered in \cref{main theorem}}
\label{Figure XX}
\end{figure}
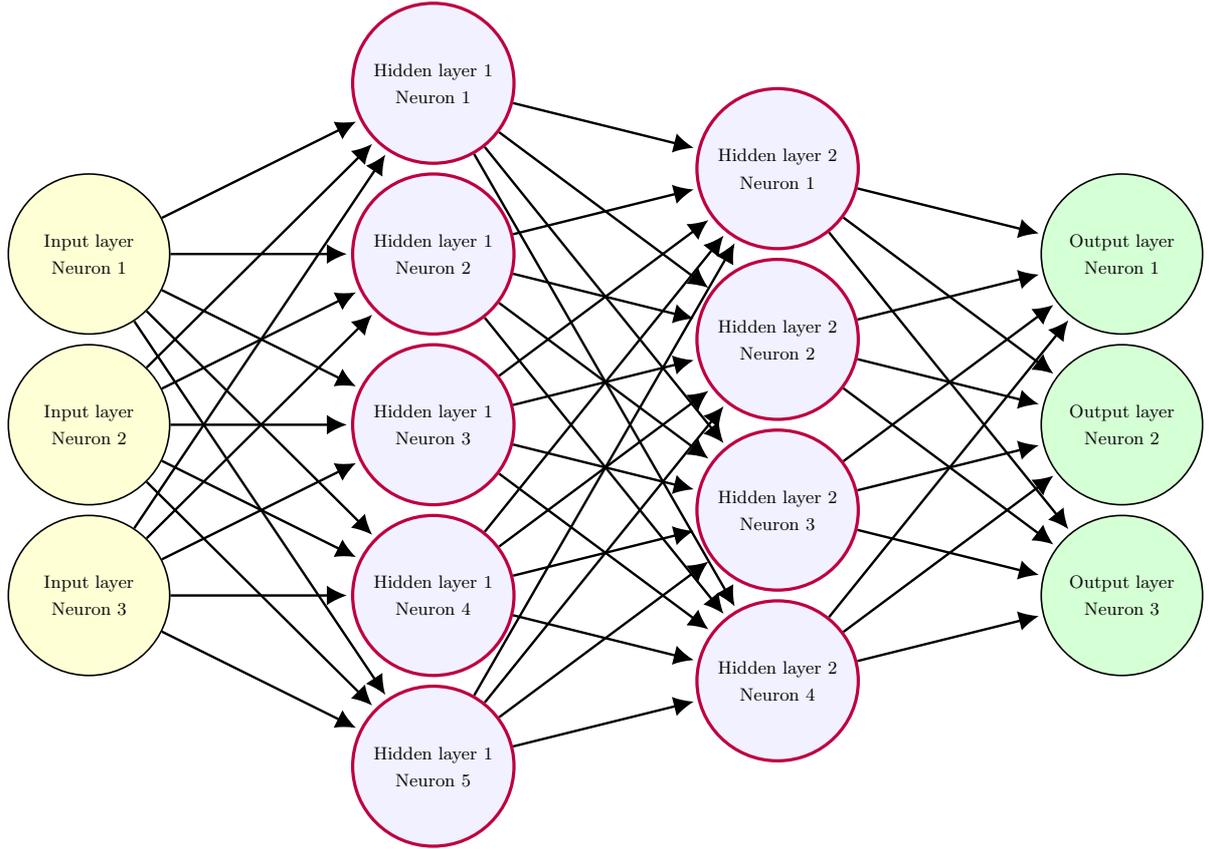

To briefly sketch the contribution of this work within this introductory, we now present in \cref{main theorem} a special case of \cref{Theorem X} in which we also omit the conclusion that generalized gradients are limiting Fr\'{e}chet subgradients for simplicity and we refer to \cref{sec: regularity} below for the formulation of \cref{Theorem X} in its full generality. In \cref{main theorem} we have that $d \in \N = \{ 1, 2, 3, \dots \} $ represents the overall number of one-dimensional parameters of the considered \ANNs, we have that $(L - 1) \in \N_0 = \{ 0, 1, 2, 3, \dots\}$ describes the number of hidden layers of the considered \ANNs, and we have that $\ell_0, \ell_1, \dots, \ell_L\in \N$ specify the number of neurons on the layers of the \ANNs\ (we also refer to \cref{Figure XX} for a graphical illustration of an example \ANN\ architecture within the class of \ANNs\ considered in \cref{main theorem}). We now present \cref{main theorem} with all mathematical details, thereafter, explain the contribution of \cref{main theorem} and the mathematical objects appearing in \cref{main theorem} in words and, thereafter, also connect the conclusion of \cref{main theorem} to related research findings in the scientific literature (see \cref{subsec: literature} below).
\renewcommand{\fd}{d}
\begin{theorem}\label{main theorem}
%For every $n \in \N$, $f \colon \R^n \to \R$ let $\mathbf{T}_f \colon \R^n \to \R^n$ denote the realization of the Tensorflow gradient of $f$
Let $ \fd, L\in \N $,
$ \ell_0,\ell_1,\dots,\ell_L\in \N $
%$\delta \in (0,\infty)$,
satisfy 
$
  \fd = \sum_{k=1}^L \ell_k ( \ell_{k-1} + 1 )
$, 
for every $\activate\in C(\R,\R)$,
$ 
  \theta = ( \theta_1, \dots, \theta_{ \fd } ) \in \R^{ \fd }$
 let $\mN^{ k, \theta }_{\activate}=(\mN^{ k, \theta }_{\activate,1},\dots,\mN^{ k, \theta }_{\activate,\ell_k}) \colon \R^{ \ell_0 }\allowbreak \to \R^{ \ell_k} $, $k\in \{0,1,\dots,L\}$, satisfy for all $k\in \{0,1,\dots,L-1\}$, $x=(x_1,\dots, x_{\ell_0})\in \R^{\ell_0}$, $i\in\{1,2,\dots,\ell_{k+1}\}$ that
\begin{equation}\label{relization multi}
\begin{split}
  \mN^{ k+1, \theta }_{\activate,i}( x ) &= \theta_{\ell_{k+1}\ell_{k}+i+\sum_{h=1}^{k}\ell_h(\ell_{h-1}+1)}\\
  &+\textstyle\sum_{j=1}^{\ell_{k}}\theta_{(i-1)\ell_{k}+j+\sum_{h=1}^{k}\ell_h(\ell_{h-1}+1)}\big(x_j\indicator{\{0\}}(k) 
  +\activate(\mN^{k,\theta}_{\activate,j}(x))\indicator{\N}(k)\big),
  \end{split}
\end{equation}
let $\mu \colon \cB(\R^{\ell_0}\times \R^{\ell_L})\to [0,\infty]$ be a finite measure with compact support, 
let $H \in C^1(\R^{\ell_L} \times \R^{\ell_L}\times\R^{\fd},\R)$, for every $\activate \in C(\R,\R)$ let $\cL_\activate \colon \R^\fd \to \R$ satisfy for all $\theta \in \R^\fd$ that 
\begin{equation}\label{eq3: maintheorem}
\textstyle 
  \cL_\activate( \theta ) 
  = 
  \int_{\R^{\ell_0}\times \R^{\ell_L}}
  H(\mathcal{N}_{\activate}^{L,\theta}(x),y,\theta)\, \mu (\mathrm{d}x,\mathrm{d}y)
  ,
\end{equation}
let $S\subseteq \R$ be finite, let $\scrA\in C(\R,\R)$ satisfy $\restr{\scrA}{\R \backslash S} \in C^1(\R \backslash S,\R)$, let $\scra \colon \R \to \R$ satisfy $\min_{ z \in \{ -1, 1 \}} \bigl( \sum_{ x \in S } \limsup_{ h \searrow 0 }\allowbreak | \scra( x + z h ) - \scra( x ) | \bigr) = 0$, and assume $\restr{\scra}{\R \backslash S}=\nabla(\restr{\scrA}{\R \backslash S})$.
Then 
\begin{enumerate}[label=(\roman*)]
\item \label{item 0: main theorem} it holds for all $A\in C^1(\R,\R)$ that $\cL_A\in C^1(\R^\fd,\R)$,
\item \label{item 1: main theorem} there exists a unique $\cG \colon \R^\fd \to \R^\fd$ which satisfies for all $\theta \in \R^\fd$ and all $(A_n)_{n\in \N} \subseteq C^1(\R,\R)$ with $\forall \, x \in \R \colon \exists \, m \in \N \colon \sum_{n=m}^\infty (|\scrA(x)-A_n(x)|+|\scra(x)-(A_n)'(x)|) = 0$ and $\forall \,m \in \N \colon \sup_{n \in \N} \sup_{x \in [-m,m]}\bigl[( |A_n(x)|  + |(A_n)'(x)| )\bigr]<\infty$ that  
\begin{equation}\label{eq4: main theorem}
\textstyle\limsup_{n \to \infty} \norm{(\nabla\cL_{A_n})(\theta) - \cG(\theta)}=0,
\end{equation}
and 
    \item \label{item 2: main theorem} it holds for all open $U\subseteq \R^d$ with $\cL_\scrA|_{U}\in  C^1( U, \R )$ that
    \begin{equation}\label{eq5: main theorem}
        \cG|_{U}=\nabla (\cL_\scrA|_{U}).
        \end{equation}
\end{enumerate}
\end{theorem}

\cref{main theorem} is an immediate consequence of \cref{Theorem X}, which is the main result of \cref{sec: regularity}. 
In \cref{main theorem} we consider fully-connected feedforward \ANNs\ with $L+1 \in \N$ layers 
\begin{itemize}
\item
with $\ell_0 \in \N$ neurons on the input layer (with an $\ell_0$-dimensional input layer), 
\item
with $\ell_1\in\N$ neurons on the $1\textsuperscript{st}$ hidden layer (with an $\ell_1$-dimensional $1\textsuperscript{st}$ hidden layer), 
\item 
with $\ell_2\in \N$ neurons on the $2\textsuperscript{nd}$ hidden layer (with an $\ell_2$-dimensional $2\textsuperscript{nd}$ hidden layer),
\item $\dots$,
\item 
with $\ell_{ L - 1 }\in\N$ neurons on the $(L-1)\textsuperscript{th}$ hidden layer (with an $\ell_{ L-1 }$-dimensional $(L-1)\textsuperscript{th}$ hidden layer), 
and 
\item 
with $\ell_L\in\N$ neurons on the output layer (with an $\ell_L$-dimensional output layer).
\end{itemize}
The set $C( \R, \R )$ in \cref{main theorem} represents the set of continuous functions from $\R$ to $\R$
and we note that for every continuous function $\activate \colon \R \to \R$ 
and every $d$-dimensional vector $\theta = ( \theta_1,\dots, \theta_d ) \in \R^d$ we have that 
the function \begin{equation}
\mN^{ L, \theta }_\activate \colon \R^{ \ell_0 } \to \R^{ \ell_L }
\end{equation}
in \cref{relization multi} in \cref{main theorem} is the realization function 
of the fully-connected feedforward \ANN\ with architecture $( \ell_0, \ell_1,\dots, \ell_L )$ 
and \ANN\ parameter vector $\theta$ with respect to the activation function $a$ (we also refer to \cref{Figure XX} for a graphical illustration of an example \ANN\
architecture within the class of \ANNs\ considered in \cref{main theorem}).

The set $\mathcal B( \R^{ \ell_0 } \times \R^{ \ell_L } )$ in \cref{main theorem} is the Borel sigma-algebra of $\R^{ \ell_0 } \times \R^{ \ell_L }$
and the measure $\mu \colon \mathcal B( \R^{ \ell_0 } \times \R^{ \ell_L }) \to [0,\infty]$ in \cref{main theorem} is typically a probability measure that can be used to describe the joint probability distribution on the $\ell_0$-dimensional input data and the $\ell_L$-dimensional 
output data of the considered supervised learning problem (cf., \eg, \cite{Cucker2001OnTM}). The measure $\mu$ in \cref{main theorem} could also 
be a discrete measure that consists of a linear combination of Dirac measures described through pairs of input-output data pairs. 

The set $C^1( \R^{\ell_L} \times \R^{\ell_L} \times \R^d, \R )$ in \cref{main theorem} is the set of all continuously differentiable functions from $\R^{\ell_L} \times \R^{\ell_L} \times \R^d$ to $\R$ and the continuously differentiable function $H \colon \R^{\ell_L} \times \R^{\ell_L} \times \R^d \to \R$ is the loss function of the considered supervised learning problem that may for every \ANN\ parameter vector $\theta \in \R^\fd$ and every activation $\activate \in C(\R,\R)$ depend on the \ANN\ approximation $\cN^{ L, \theta }_\activate( x ) \in \R^{\ell_L}$ of the output $y \in \R^{\ell_L}$ given the input $x \in\R^{\ell_0}$ (first argument of $H$), on the output $y \in \R^{ \ell_L }$ (second argument of $H$), and on the \ANN\ parameter vector $\theta \in \R^\fd$ (third argument of $H$).

In \cref{eq3: maintheorem} in \cref{main theorem} we describe for every activation $\activate \in C(\R,\R)$ the function $\cL_\activate \colon\R^\fd\to\R$ that one intends to minimize (the objective function) associated to the considered supervised learning problem. The function $\cL_\activate$ could be the true risk function of the considered supervised learning problem or, if the measure $\mu$ is a discrete measure consisting of a linear combination of Dirac measures described through pairs of input-output data pairs, then the function $\cL_\activate$ in \cref{main theorem} reduces to the empirical risk function. 

The continuous function $\scrA \colon \R \to \R$ in \cref{main theorem} is the considered activation function, which is assumed to be continuously differentiable on an open subset of the set of real numbers that agrees with the set of real numbers except of finitely many points, collected in the set $S$, and the function $\scra \colon \R\to\R$ in \cref{main theorem} is assumed to coincide on $\R\backslash S$ with the derivative of $\scrA$. In this sense the function $\scra$ serves as a generalized derivative of the considered activation function $\scrA$. For example, $\scrA$ may coincide with the \ReLU\ activation function $\R \ni x \mapsto \max\{x,0\}\in \R$ in which case we choose $S = \{0\}$ and in which case we can choose $\scra$ to be, \eg, the left derivative 
$\R \ni x \mapsto \mathbbm 1_{(0,\infty)}(x) \in \R$ of the \ReLU\ activation. This selection of the left derivative corresponds 
to the choice considered in {\sc Pytorch} and {\sc Tensorflow} (cf., \eg, \cite{TensorFlow2016, pytorch2019}). We note, however, that even if $\scra$ is chosen as the left derivative of $\scrA$ it does not hold that the generalized gradient function $\cG \colon \R^\fd\to\R^\fd$ in \cref{item 1: main theorem} in \cref{main theorem} coincides with the left derivative of $\cL_{\scrA}$. 

In \cref{item 1: main theorem} in \cref{main theorem} we reveal an accurate and concise description of the generalized gradient used in the training of deep fully-connected feedforward \ANNs\ and in \cref{item 2: main theorem} we demonstrate that the generalized gradient function coincides on the largest open set on which the objective function is continuously differentiable with the gradient of the objective function. In \cref{Theorem X}, which is the main result of \cref{sec: regularity}, we present a generalized and strengthened variant of \cref{main theorem} in which also the notion of limiting Fr\'{e}chet subgradients (cf.\ \cref{def:limit:subdiff} in \cref{subsec: frechet definition} below) appears. In \cref{thm:mainthm_sequence: part 1}, which is the main result of \cref{sec: explicit representation}, we present a generalization of \cref{item 1: main theorem} in \cref{main theorem} and \cref{Theorem X}, respectively.
\subsection{Literature review on related research findings}\label{subsec: literature}
Developing a rigorous mathematical theory to  justify the remarkable success of deep learning has become a dynamic and highly active area of research. An integral part of the success of supervised learning algorithms in practice is their training procedure, which typically involves a programmatic calculation of the gradient of the risk function of an \ANN. A major challenge  to mathematically justify the success of deep learning comes from the fact that the risk function of an \ANN\ is in many practical scenarios neither convex nor differentiable. Nonetheless, the training of \ANNs\ is typically build on \GD\ type optimization algorithms.

\GD\ type optimization is essentially well understood when the objective function is convex or strongly convex, respectively (cf., \eg, the review articles and monographs \cite{Guro2023handbook,ArBePhi2024,Nesterov2004} and the references mentioned therein). However, in the absence of global assumptions on the objective function like convexity, gradient based methods may diverge or converge to saddle points or non-global local minima; cf., \eg, \cite{gallon2022blowphenomenagradientdescent,HannibalJentzenThang2024,ArAd2024,QuocTung2023,MR4243432} (we also refer, \eg, to \cite{BLXZ2024,CheriditoJentzenRossmanek2022,DingLiSun2019,SafranShamir2017,SwirszczCzarneckiParcanu,LuAfBru2019,ZZLX2022} for works studying critical points of the risk function in the training of \ANNs). In particular, the analysis on the convergence properties of the optimization processes in the training of \ANNs\ has received a lot of attention in recent years. We refer, \eg, to \cite{CheriditoJentzenRossmanek2021,DeRoAr2024nonconvergence,gallon2022blowphenomenagradientdescent,HannibalJentzenThang2024,ArAd2024,QuocTung2023,LSSK2020,MR4243432,ReddiKale2019,ShinKarniadakis_2020} for non-convergence results for gradient based optimization methods and we refer, \eg, to \cite{MR4056927,DereichKassingconvergence2021,Gentile2022,Ibragimov2022,Aradexistenceofglobmin,Welper2024,MR4765349} and the references therein for convergence results for gradient based optimization methods.

% Non-convergence results can, \eg, be found in  \cite{CheriditoJentzenRossmanek2021,HannibalJentzenThang2024,ArAd2024,LSSK2020,ShinKarniadakis_2020}.
% For convergence results, we refer, \eg , to \cite{MR4056927,DereichKassingconvergence2021,Gentile2022,Ibragimov2022,Aradexistenceofglobmin,Welper2024,MR4765349}.

In many practically relevant scenarios, the risk function of an \ANN\ is not differentiable. For instance, this occurs when the popular \ReLU\ function is employed as the activation function. This challenge is bypassed in practice by making use of the automatic differentiation principle: Roughly speaking, in automatic differentiation the gradient of the risk function of an \ANN\ is calculated by defining generalized derivatives, even for non-differentiable functions, and employing the chain rule in the concept of so-called computational graphs. For instance, in {\sc Pytorch} and {\sc Tensorflow} (cf., \eg, \cite{TensorFlow2016, pytorch2019}) the left derivative is used as the generalized derivative of the \ReLU\ function in the implementation of the automatic differentiation algorithm. A mathematical model for automatic differentiation has been proposed, \eg, in \cite{BolPau}. We also refer, \eg, to \cite{BBPP2022,BPV22,Kakade2018ProvablyCA,LYRY_PAP} for results that motivate the usage of automatic differentiation in practice from a theoretical point of view. In particular, \cite{BP2021} introduced the notion of conservative set-valued fields as a generalization of the Clarke subgradient in order to describe the output of automatic differentiation algorithms. Several articles followed to study geometrical properties of the conservative set-valued fields of definable functions, \eg, in \cite{Da2022conservative,LeTian2021structure,Pauwels2023conservative}
 and to employ the concept of conservative set-valued fields in order to establish convergence results of \SGD\ algorithm, \eg, in \cite{BoPauSi2024differentiating,KuNaKim2023adamfamily,NaXiXinKim2023adamfamily,NaXiKim2023convergence}. 
An article particularly worth mentioning in connection with the content of our article, which makes use of the concept of conservative set-valued fields from 
\cite{BP2021}, is the recent work \cite{Sholom2024gradientlimit}
in which the author proves that the limit of conservative set-valued fields of definable functions is under appropriate assumptions a conservative set-valued field (see \cite[Theorem 1]{Sholom2024gradientlimit}). 

In this article we specify generalized gradients of the risk function of \ANNs\ through an approximation approach and, in contrast to \cite{BP2021}, 
not as a set-valued mapping due to the uniqueness that we establish in \cref{item 1: main theorem} of \cref{main theorem}. The specific approximation approach that we employ to specify the generalized gradient function in this work is based on the approximation approaches in \cite{CheriditoJentzenRiekert2021,ScarpaMultiLayers,ArAd2022aproofofconvergence} in which, loosely speaking, specialized and weakened variants of the considered approximation procedure have been proposed and used; cf., \eg, also \cite{Aradexistenceofglobmin,JentzenRiekert_Flow}. In the mentioned articles \cite{CheriditoJentzenRiekert2021,ScarpaMultiLayers,Aradexistenceofglobmin,ArAd2022aproofofconvergence,JentzenRiekert_Flow} the activation function is given by the \ReLU\ function and the loss function is given by the mean squared error loss. In \cite{CheriditoJentzenRiekert2021} the sequence of functions that approximates the \ReLU\ function is given explicitly (see \cite[Setting 2.1]{CheriditoJentzenRiekert2021}). In contrast, in  \cite{ScarpaMultiLayers,Aradexistenceofglobmin,JentzenRiekert_Flow} the sequence of functions that approximates  the \ReLU\ function is not specified explicitly. In \cite{ScarpaMultiLayers,Aradexistenceofglobmin,ArAd2022aproofofconvergence,JentzenRiekert_Flow} the approximating sequence of continuously differentiable functions converges to the \ReLU\ function in the pointwise sense and the derivatives of the elements of the approximating sequence converge to the left derivative of the \ReLU\ function in the pointwise sense.

A central motivation to establish a rigorous mathematical framework to describe gradients in deep learning is to prove convergence results for gradient based methods in the training of \ANNs. The results in this work thus motivate future research efforts that aim to establish convergence results in the training of \ANNs\ applicable to the generalized gradients in this work.

\subsection{Structure of this article}
The remaining article is organized as follows. In the main result of \cref{sec: explicit representation}, \cref{thm:mainthm_sequence: part 1} in \cref{sec: explicit representation}, we establish a generalized and strengthened variant of \cref{item 1: main theorem} in \cref{main theorem} above. In \cref{sec: regularity} we establish in \cref{prop:loss:gradient:subdiff} that the generalized gradient function of the objective function in \cref{item 1: main theorem} in \cref{main theorem} consists of limiting Fr\'{e}chet subgradients of the objective function. This fact then allows us to conclude that the generalized gradient function of the objective function in \cref{item 1: main theorem} of \cref{main theorem} coincides on the largest open set on which the objective function is continuously differentiable with the usual gradient of the objective function, which we reveal and summarize in \cref{cor:cG_equal_to_gradient} and \cref{Theorem X} in \cref{sec: regularity}, respectively. \Cref{item 1: main theorem,item 2: main theorem} in \cref{main theorem} above are direct consequences of \cref{item 1: Theorem X,item 2: Theorem X} in \cref{Theorem X}.

%giu cho ky
% Finally, in \cref{sec: example} we illustrate the conclusions of \cref{main theorem} and \cref{Theorem X}, respectively, in the situation of several concrete loss functions such as the cross-entropy loss function.
\newpage

\section{Generalized gradients of the objective function}\label{sec: explicit representation}
\renewcommand{\fg}{A}
In the main result of this section, \cref{thm:mainthm_sequence: part 1} in \cref{ssec:rep_formula_gengrads}, we present an appropriate approximation procedure that uniquely describes the generalized gradient function in the training of deep fully-connected feedforward \ANNs\ (see \cref{item 2: Theorem X part 1} in \cref{thm:mainthm_sequence: part 1} in \cref{sec: explicit representation}). The conclusion of \cref{thm:mainthm_sequence: part 1} is a generalized and strengthened variant of the conclusion in \cref{item 1: main theorem,item 2: main theorem} in \cref{main theorem} above. All the other frameworks, concepts, and results in \cref{subsec: mathematical framework,,subsec: existence of A_n,,ssec:rep_formula_gengrads} within this section are basically preparations and auxiliary results in order to establish \cref{thm:mainthm_sequence: part 1}. In the following we briefly sketch some of these preparations for \cref{thm:mainthm_sequence: part 1}. The arguments in this section are in parts based on the arguments in \cite{CheriditoJentzenRiekert2021,ScarpaMultiLayers,ArAd2022aproofofconvergence} in which, roughly speaking, specialized and weakened variants of the approximation procedure in \cref{thm:mainthm_sequence: part 1} have been proposed and used. 

In \cref{subsec: mathematical framework} we introduce in \cref{setting:activationapprox} our mathematical framework for \ANNs\ in which we approximate the activation function $ \fg_0 \colon \R \to \R $ through a sequence of continuously differentiable functions $ \fg_\approximate \colon \R \to \R $, $ n \in \N $. We assume that the functions $ \fg_\approximate \colon \R \to \R $, $\approximate \in \N$, converge in the pointwise sense to the activation function $ \fg_0 $ with respect to the discrete topology and we assume that the derivative functions $ ( \fg_\approximate )' \colon \R \to \R $, $n \in \N$, converge in the pointwise sense to a generalized derivative of the activation function $ \fg_0 $ with respect to the discrete topology; see \cref{def: g_r} in \cref{setting:activationapprox}. This generalized derivative is the subject of \cref{def:LKB_left_right_derivative}. In \cref{setting:activationapprox} we specify for every $ n \in \N_0 $ and every $\theta = (\theta_1,\dots,\theta_\fd) \in \R^\fd$ the realization function $ \mN^{L,\theta}_{n} \colon \R^{\ell_0} \to\R^{\ell_L}$ of the \ANN\ with the activation function $ \fg_\approximate \colon \R \to \R $, the \ANN\ parameter vector $ \theta $, and the \ANN\ architecture vector $( \ell_0, \ell_1, \dots, \ell_L ) \in \R^{ L + 1 }$. The natural number $\fd \in \N$ in \cref{setting:activationapprox} describes the number of real parameters $\fd = \sum_{n=1}^{L} \ell_n(\ell_{n-1}+1)$ of the \ANNs. In addition, in \cref{eq:def_risk_function} in \cref{setting:activationapprox} for every $ \approximate \in \N_0 $ we denote by $ \mathcal{L}_\approximate \colon \R^\fd \to \R $ the risk function associated to the \ANNs\ with the activation function $ \fg_\approximate \colon \R \to \R $ and the \ANN\ architecture vector $( \ell_0, \ell_1, \dots, \ell_L )$. 

The aim of \cref{subsec: existence of A_n} is to supply sufficient conditions that ensure the existence of continuously differentiable functions $ \fg_\approximate \colon \R \to \R $, $ n \in \N $, approximating the activation function $\fg_0 \colon \R \to \R$ in the sense of \cref{def: g_r} in \cref{setting:activationapprox}; cf.\ \cref{lem: existence of A_n} and \cref{cor: equivalent sequence} in \cref{subsec: existence of A_n}.

In \cref{ssec:rep_formula_gengrads} we prove several auxiliary results for \cref{thm:mainthm_sequence: part 1} such as, \eg, the representation formulas for the partial derivatives of the risk functions $\mathcal{L}_n \colon \R^\fd \to \R$, $n \in \N$, in \cref{it2:formulaLrgradient} and \cref{it3:formulaLrgradient} in \cref{lem:rep_formulaLrgradient}. Another central tool to establish \cref{thm:mainthm_sequence: part 1} is formulated in \cref{prop:gcNconvergence} in \cref{ssec:rep_formula_gengrads} below. \cref{prop:gcNconvergence} employs the assumption that the functions $ \fg_\approximate \colon \R \to \R $, $ n \in \N $, converge in the pointwise sense to the activation function $ \fg_0 $ with respect to the discrete topology and employs the assumption that the derivative functions $ ( \fg_\approximate )' \colon \R \to \R $, $ n \in \N $, converge in the pointwise sense to the generalized derivative of the activation function $ \fg_0 $ with respect to the discrete topology to establish, among other things, pointwise convergence of the realization functions of the corresponding \ANNs\ with respect to the discrete topology.

\subsection{Mathematical framework for deep artificial neural networks (ANNs)}\label{subsec: mathematical framework}
\begin{definition}\label{def:LKB_left_right_derivative}
    Let $A\subseteq\R$ be closed and let $f\colon \R\to\R$ and $g\colon A\to\R$ be functions which  satisfy $f|_{\R\backslash A}\in C^1(\R\backslash A,\R)$. Then we denote by $\scrd_g f\colon \R\to\R$ the function which satisfies for all $x\in \R$ that
    \begin{equation}
        (\scrd_g f)(x)=\begin{cases}
            f'(x)&\colon x\in \R\backslash A\\
            g(x)&\colon x\in A.
        \end{cases}
    \end{equation}
\end{definition}
\begin{setting}\label{setting:activationapprox}
Let  
$ ( \ell_k )_{ k \in \N_0 } \subseteq \N $, for every $k \in \N_0$ let $\mathbf{d}_k \in \N_0$ satisfy $\mathbf{d}_k = \sum_{n=1}^k \ell_n ( \ell_{n-1} + 1 )$, let
%$\delta \in (0,\infty)$,
$ L, \fd \in \N $
satisfy 
$
  \fd = \mathbf{d}_L 
$, 
for every 
$ 
  \theta = ( \theta_1, \dots, \theta_{ \fd } ) \in \R^{ \fd } 
$
let 
$
  \fb^{ k, \theta } 
  = 
  ( \fb^{ k, \theta }_1, \dots, \fb^{ k, \theta }_{ \ell_k} )
  \in \R^{ \ell_k } 
$,
$ k \in \N $, 
and
$ 
  \fw^{ k, \theta } = 
  ( \fw^{ k, \theta }_{ i, j } )_{ 
    (i,j) \in \{ 1, \ldots, \ell_k \} \times \{ 1, \ldots, \ell_{ k - 1 } \} 
  }
  \in \R^{ \ell_k \times \ell_{ k - 1 } }
$, 
$
  k \in \N 
$, 
satisfy for all 
$ k \in \{ 1, \dots, L \} $, 
$ i \in \{ 1, \ldots, \ell_k \} $,
$ j \in \{ 1, \ldots, \ell_{ k - 1 } \} $ 
that
\begin{equation}
\label{wb}
  \fw^{ k, \theta }_{ i, j }
  = 
  \theta_{ ( i - 1 ) \ell_{ k - 1 } + j 
  + 
  \sum_{ h = 1 }^{ k - 1 } \ell_h ( \ell_{ h - 1 } + 1 ) }
\qqandqq
  \fb^{ k, \theta }_i 
  =
  \theta_{ \ell_k \ell_{ k - 1 } + i 
  + 
  \sum_{ h = 1 }^{ k - 1 } \ell_h ( \ell_{ h - 1 } + 1 ) } 
  ,
\end{equation}
let $\element\in \N$, $y_1,y_2,\ldots,y_\element \in \R$, let $\func \colon \{y_1,y_2,\ldots,y_\element\}\to \R$ and $\fg_\approximate \colon \R \to \R$, $\approximate \in \N_0$, satisfy for all $\approximate \in \N$ that $\fg_\approximate \in C^1(\R,\R)$ and $ \restr{\fg_0}{\R \backslash \{y_1,y_2,\ldots,y_\element\}} \in C^1(\R \backslash \{y_1,y_2,\ldots,y_\element\},\R)$, 
for every $\approximate\in \N_0$, $\theta \in \R^\fd$ let $\mathcal{N}_{\approximate}^{k,\theta}=\big(\mathcal{N}_{\approximate,1}^{k,\theta},\ldots,\mathcal{N}_{\approximate,\ell_k}^{k,\theta}\big) \colon \R^{\ell_0} \to \R^{\ell_k}$, $ k \in \N_0$, 
satisfy 
for all 
$k\in \N_0$, $x\in \R^{\ell_0}$,  $i\in \{1,2,\dots,\ell_k\}$
that
\begin{equation}
\label{eq:def_NN_realization}
  \mathcal{N}_{\approximate,i}^{k+1,\theta} 
  = \fb^{k+1,\theta}_i + 
  \sum_{j=1}^{\ell_{k}}\fw^{k+1,\theta}_j \bigl(x_j\mathbbm 1_{\{0\}}(k)+\fg_\approximate\big(\mathcal{N}^{k,\theta}_{\approximate,j}(x)\bigr)\mathbbm 1_{\N}(k)\big) ,
\end{equation}
assume for all $x \in \R$ that there exists $m \in \N$ such that
\begin{equation}\label{def: g_r}
\textstyle\sum\nolimits_{n=m}^\infty\big[|\fg_\approximate(x)-\fg_0(x)| + |(\fg_\approximate)'(x) - (\mathscr{d}_\func \fg_0)(x)|\big]=0, \cfadd{def:LKB_left_right_derivative} 
\end{equation}
let $\mu \colon \cB(\R^{\ell_0}\times \R^{\ell_L})\to [0,\infty]$ be a finite measure, 
let $H \in C^1(\R^{\ell_L} \times\R^{\fd}\times \R^{\ell_L},\R)$, for every $\approximate \in \N_0 $ let $\cL_\approximate \colon \R^\fd \to \R$ satisfy for all $\theta \in \R^\fd$  that $\int_{\R^{\ell_0} \times \R^{\ell_L}}
  |H(\mathcal{N}_{\approximate}^{L,\theta}(x),\theta,y)| \, \mu(\mathrm{d}x,\d y)<\infty$ and
\begin{equation}
\label{eq:def_risk_function}
\textstyle 
  \cL_\approximate( \theta ) 
  = 
  \int_{\R^{\ell_0} \times \R^{\ell_L}}
  H\big(\mathcal{N}_{\approximate}^{L,\theta}(x),\theta,y\big) \, \mu(\mathrm{d}x,\d y)
  ,
\end{equation}
assume for all $m\in \N$ that $\sup_{\approximate \in \N}\sup_{x\in[-m,m]}\allowbreak\bigl[(|\fg_\approximate(x)|+|(\fg_\approximate)'(x)|)\mathbbm 1_{\{\infty\}}(|\mathrm{supp}(\mu)|)\bigr]<\infty$, assume that $\{x\in \R^{\ell_0}\colon (\exists\, y\in \R^{\ell_L}\colon (x,y)\in \mathrm{supp}(\mu))\}$ is bounded, assume for all $r\in (0,\infty)$, $X\in [-r,r]^{\ell_0}$, $\Theta\in [-r,r]^{\fd}$ that
\begin{equation}\label{setting: assume}
   \textstyle \sup\limits_{x\in [-r,r]^{\ell_0}}\sup\limits_{y\in \R^{\ell_L}}\sup\limits_{\theta\in [-r,r]^\fd}\Bigl(\frac{(\|\nabla_x H(x,\theta,y)\|+\|\nabla_\theta H(x,\theta,y)\|)\mathbbm 1_{\mathrm{supp}(\mu)}(x,y)}{1+|H(X,\Theta,y)|}\Bigr)<\infty,
\end{equation} 
 and let $\cG=\big(\cG_1,\ldots,\cG_\fd \big)  \colon \R^\fd \to\R^\fd $ satisfy for all   $\theta \in 
  \{
    \vartheta \in \R^{ \fd } \colon 
    ( ( \nabla\cL_\approximate )( \vartheta ) )_{ \approximate \in \N }
    \text{ is convergent} 
  \}$ that $\cG(\theta) = \lim_{\approximate \to \infty} (\nabla \cL_\approximate) (\theta)$   \cfadd{def:LKB_left_right_derivative}\cfout.
\end{setting}
\newcommand{\G}{G}
\newcommand{\inndex}{\mathbf{i}}
\subsection{On the existence of regular approximations of the activation function}\label{subsec: existence of A_n}
\begin{athm}{prop}{lem:existence_approxsequence}
Let $\element \in \N$,  $y_1,y_2,\ldots,y_\element\in \R$ satisfy $y_1<y_2<\ldots<y_\element$, let $S\subseteq\R$ satisfy $S=\{y_1,y_2,\ldots,y_\element\}$, let $h \colon S\to \R$ and $g \colon \R \to \R$ satisfy $\restr{g}{\R \backslash S}\in C^1(\R \backslash S,\R)$, for every $\varepsilon\in (0,\infty)$ let $O_\varepsilon\subseteq\R$ satisfy $O_\varepsilon=\cup_{i=1}^\element (y_i-\varepsilon,y_i+\varepsilon)$, let $\delta \in (0,\infty)$ satisfy 
$
\delta = \frac{1}{2}\min([\cup_{v,w\in S,\, v\neq w}\{|v-w|\}]\cup\{1\})$, for every $x \in  O_\delta$ let $\inndex_x \in \N$ satisfy $\inndex_x= \mathrm{argmin}_{i \in \{1,2,\dots,\element\}} |x-y_i|$, let $\eta \in C^1(\R,\R)$ satisfy 
\begin{equation}\llabel{def: eta}
\eta(0)=\eta'(0)=\eta'(1)=0,\qquad \eta(1)=1, \qqandqq \eta((0,1))\subseteq [0,1],
\end{equation}
and for every $\approximate \in \N$ let  $\G_\approximate \colon \R \to \R$ satisfy for all $u \in \R \backslash O_{\delta \approximate^{-1}}$, $v \in \overline{O_{\delta (2\approximate)^{-1}}}$, $w \in O_{\delta \approximate^{-1}}\backslash \overline{O_{\delta (2\approximate)^{-1}}}$ that
\begin{align}\llabel{assume}
\G_\approximate(u)&=g(u),\qquad \G_\approximate(v) = h(y_{\inndex_v})  (x-y_{\inndex_v}) + g(y_{\inndex_v}), \qqandqq \\ \nonumber
\G_\approximate(w)&= \Big( 1 -\eta \Big(\frac{2\approximate|w-y_{\inndex_w}| - \delta}{\delta}\Big)\Big)  \big(h(y_{\inndex_w}) (w-y_{\inndex_w}) + g(y_{\inndex_w})\big) + \eta\Big( \frac{2\approximate|w-y_{\inndex_w}| - \delta}{\delta} \Big) g(w).
\end{align}
Then
\begin{enumerate}[label=(\roman*)]
\item\label{it1:existence_approxsequence} for all $\approximate \in \N$ it holds that $\G_\approximate \in C^1(\R,\R)$ and
\item\label{it2:existence_approxsequence} for all $x \in \R$ there exists $m \in \N$ such that for all $\approximate \in \N \cap [m , \infty)$ it holds that
\begin{equation}
\G_\approximate(x)=g(x) \qqandqq (\G_\approximate)'(x) = (\mathscr{d}_\func  g ) (x).
\end{equation} 
\end{enumerate}
\end{athm}
\begin{aproof}
Throughout this proof let $\gamma_1,\gamma_2,\ldots,\gamma_\element \in \R$ satisfy for all $i \in \{1,2,\dots,\element\}$ that $h(y_i)=\gamma_i$ and for every $\approximate \in \N$ let $D_\approximate^-\subseteq \R$ and $D_\approximate^+\subseteq \R$ satisfy
\begin{equation}\llabel{def: D_r}
D_\approximate^- = \Big\{ x \in \R \colon \lim_{y \nearrow x} \frac{\G_\approximate(y)-\G_\approximate(x)}{y-x} \text{ exists}\Big\} 
\end{equation}
and
\begin{equation}\llabel{def: D_r2}
D_\approximate^+ = \Big\{ x \in \R \colon \lim_{y \searrow x} \frac{\G_\approximate(y)-\G_\approximate(x)}{y-x} \text{ exists}\Big\}
\end{equation}
and let $(\partial_- \G_\approximate )\colon D_\approximate^-\to \R$ and $(\partial_+ \G_\approximate)\colon D_\approximate^+ \to \R$ satisfy for all $x \in D_\approximate^-$, $y \in D_\approximate^+$ that
\begin{equation}\llabel{def: partial}
(\partial_- g)(x) = \lim_{z \nearrow x} \frac{g(z)-g(x)}{z-x} \qqandqq (\partial_+ g)(y) = \lim_{z \searrow y} \frac{g(z)-g(y)}{z-y}.
\end{equation}
\argument{\lref{assume};\lref{def: partial}}{that for all $\approximate \in \N$, $x \notin \bigl(\cup_{i=1}^n \{y_i-\delta \approximate^{-1}, y_i-\delta (2\approximate)^{-1},y_i+\delta (2\approximate)^{-1},y_i+\delta \approximate^{-1} \}\bigr)$ it holds that 
\begin{equation}\label{eq:left_eq_right}
x \in D_\approximate^- \cap D_\approximate^+\qqandqq\partial_- \G_\approximate(x)= \partial_+ \G_\approximate(x)\dott
\end{equation}}
\argument{\lref{def: eta}; the fact that for all $x\in O_\delta$ it holds that $\inndex_x= \mathrm{argmin}_{i \in \N \cap [1,n]} |x-y_i|$}{ that for all $\approximate \in \N$, $x \in O_{\delta \approximate^{-1}}\backslash \overline{O_{\delta (2\approximate)^{-1}}}$ it holds that 
\begin{equation}\llabel{eq:etapart_derivative}
\begin{split}
\frac{\partial}{\partial x} &\Big[\Big( 1 -\eta \Big(\frac{2\approximate|x-y_{\inndex_x}| - \delta}{\delta}\Big)\Big)  \big(\gamma_{\inndex_x} (x-y_{\inndex_x}) +g(y_{\inndex_x})\big) + \eta\Big( \frac{2\approximate|x-y_{\inndex_x}| - \delta}{\delta} \Big) g(x) \Big] \\
&=\Big( 1- \eta\Big(\frac{2\approximate|x-y_{\inndex_x}|-\delta}{\delta}\Big)\Big) \gamma_{\inndex_x} \\
&- \eta'\Big(\frac{2\approximate|x-y_{\inndex_x}|-\delta}{\delta} \Big) (-1)^{\indicator{(0,\infty)}{(y_{\inndex_x}-x)}} \frac{2\approximate}{\delta}\big(\gamma_{\inndex_x} (x-y_{\inndex_x})+g(y_{\inndex_x})\big) \\
& + \eta'\Big(\frac{2\approximate|x-y_i|-\delta}{\delta}\Big)(-1)^{\indicator{(0,\infty)}{(y_{\inndex_x}-x)}} g(x) \frac{2\approximate}{\delta} + g'(x) \eta\Big(\frac{2\approximate|x-y_{\inndex_x}|-\delta}{\delta}\Big) \\
&= \frac{2\approximate}{\delta}(-1)^{\indicator{(0,\infty)}{(y_{\inndex_x}-x)}} \eta'\Big(\frac{2\approximate|x-y_{\inndex_x}|-\delta}{\delta}\Big) \big(g(x)-g(y_{\inndex_x}) -\gamma_{\inndex_x} (x-y_{\inndex_x}) \big) \\
& + \Big( 1- \eta\Big(\frac{2\approximate|x-y_{\inndex_x}|-\delta}{\delta}\Big)\Big) \gamma_{\inndex_x}+ g'(x) \eta\Big(\frac{2\approximate|x-y_{\inndex_x}|-\delta}{\delta}\Big)\dott
\end{split}
\end{equation}}
\argument{\lref{def: eta};;\lref{assume};\lref{eq:etapart_derivative}}{for all $\approximate \in \N$ that
\begin{equation}\label{eq:grC1_1}
\begin{split}
(\partial_- \G_\approximate )\big(y_{\inndex_x}-\delta (2\approximate)^{-1}\big) &=  -\frac{2\approximate}{\delta} \eta'(0)\big(g(y_{\inndex_x} - \delta (2\approximate)^{-1})- g(y_{\inndex_x})+ \gamma_{\inndex_x} \delta (2\approximate)^{-1}\big) \\
&\qquad+ \big( 1- \eta(0)\big) \gamma_{\inndex_x} + g'(y_{\inndex_x}-\delta (2\approximate)^{-1})\eta(0) \\
&=\gamma_{\inndex_x} = (\partial_+  \G_\approximate)(y_{\inndex_x}-\delta (2\approximate)^{-1})
\end{split}
\end{equation}
and 
\begin{equation}\label{eq:grC1_2}
\begin{split}
(\partial_+ \G_\approximate )\big(y_{\inndex_x}+\delta (2\approximate)^{-1}\big) &=  \frac{2\approximate}{\delta} \eta'(0)\big(g(y_{\inndex_x} + \delta (2\approximate)^{-1})- g(y_{\inndex_x})- \gamma_{\inndex_x} \delta (2\approximate)^{-1}\big) \\
&\qquad+ \big( 1- \eta(0)\big) \gamma_{\inndex_x} + g'(y_{\inndex_x}+\delta (2\approximate)^{-1})\eta(0) \\
&=\gamma_{\inndex_x}= (\partial_-  \G_\approximate)(y_{\inndex_x}+\delta (2\approximate)^{-1}).
\end{split}
\end{equation}}
\argument{\lref{def: eta};\lref{assume}}{for all $\approximate \in \N$ that
\begin{equation}\label{eq:grC1_3}
\begin{split}
(\partial_+ \G_\approximate )\big(y_{\inndex_x}-\delta \approximate^{-1}\big) &=  -\frac{2\approximate}{\delta} \eta'(1)\big(g(y_{\inndex_x} - \delta \approximate^{-1})- g(y_{\inndex_x})+ \gamma_{\inndex_x} \delta \approximate^{-1}\big) \\
&\qquad+ \big( 1- \eta(1)\big) \gamma_{\inndex_x} + g'(y_{\inndex_x}-\delta (\approximate)^{-1})\eta(1) \\
&=g'(y_{\inndex_x}-\delta \approximate^{-1}) = (\partial_-  \G_\approximate)(y_{\inndex_x}-\delta \approximate^{-1})
\end{split}
\end{equation}
and
\begin{equation}\label{eq:grC1_4}
\begin{split}
(\partial_- \G_\approximate )\big(y_{\inndex_x}+\delta \approximate^{-1}\big) &=  \frac{2\approximate}{\delta} \eta'(1)\big(g(y_{\inndex_x} +\delta \approximate^{-1})- g(y_{\inndex_x})- \gamma_{\inndex_x} \delta \approximate^{-1}\big) \\
&\qquad+ \big( 1- \eta(1)\big) \gamma_{\inndex_x} + g'(y_{\inndex_x}+\delta (\approximate)^{-1})\eta(1) \\
&=g'(y_{\inndex_x}+\delta \approximate^{-1}) = (\partial_+  \G_\approximate)(y_{\inndex_x}+\delta \approximate^{-1}).
\end{split}
\end{equation}}
\argument{\cref{eq:left_eq_right}; \cref{eq:grC1_1}; \cref{eq:grC1_2}; \cref{eq:grC1_3}; \cref{eq:grC1_4}}
{that for all $\approximate \in \N$ it holds that \llabel{arg1} $D_\approximate^-=D_\approximate^+=\R$ and $(\partial_- \G_\approximate) = (\partial_+ \G_\approximate)$\dott}
\argument{\lref{arg1};}{\cref{it1:existence_approxsequence}\dott} 
\startnewargseq
\argument{\lref{assume};}{for all $\approximate\in \N$, $i \in \{1,2,\dots,\element\}$ that 
\begin{equation}\label{eq:discreteconv_atyi}
\G_\approximate(y_i)=g(y_i) \qqandqq (\G_\approximate)'(y_i) = \big(\mathscr{d}_\func g\big)(y_i).
\end{equation}}
\argument{\lref{assume};}{for all $x \in \R \backslash \{y_1,y_2,\ldots,y_\element\}$ that there exists $m\in\N$ such that for all $\approximate\in \N\cap [m,\infty)$ it holds that
\begin{equation}\llabel{eq2}
\G_\approximate(x)=x \qqandqq (\G_\approximate)'(x) = (\mathscr{d}_\func g)(x).
\end{equation}}
\argument{\lref{eq2}; \cref{eq:discreteconv_atyi}}{\cref{it2:existence_approxsequence}\dott}
\startnewargseq
\end{aproof}
\begin{athm}{prop}{lem:existence_approxsequence2}
Let $\element \in \N$,  $y_1,y_2,\ldots,y_\element\in \R$ satisfy $y_1<y_2<\ldots<y_\element$, let $S\subseteq\R$ satisfy $S=\{y_1,y_2,\ldots,y_\element\}$, let $g \in C(\R,\R)$ satisfy $\restr{g}{\R \backslash S}\in C^1(\R \backslash S,\R)$, let $\func \colon S\to \R$ be a function, assume that $\scrd_\func g$ is locally bounded, for every $\varepsilon\in (0,\infty)$ let $O_\varepsilon\subseteq\R$ satisfy $O_\varepsilon=\cup_{i=1}^n (y_i-\varepsilon,y_i+\varepsilon)$, let $\delta \in (0,\infty)$ satisfy $
\delta = \frac{1}{2}\min([\cup_{v,w\in S,\, v\neq w}\{|v-w|\}]\cup\{1\})$, for every $x \in  O_\delta$ let $\inndex_x \in \N$ satisfy $\inndex_x= \mathrm{argmin}_{i \in \{1,2,\dots,\element\}} |x-y_i|$, let $\eta \in C^1(\R,\R)$ satisfy 
\begin{equation}\llabel{def: eta}
\eta(0)=\eta'(0)=\eta'(1)=0,\qquad \eta(1)=1, \qqandqq \eta((0,1))\subseteq [0,1],
\end{equation}
and for every $\approximate \in \N$ let  $\G_\approximate \colon \R \to \R$ satisfy for all $u \in \R \backslash O_{\delta \approximate^{-1}}$, $v \in \overline{O_{\delta (2\approximate)^{-1}}}$, $w \in O_{\delta \approximate^{-1}}\backslash \overline{O_{\delta (2\approximate)^{-1}}}$ that
\begin{align}\llabel{assume}
&\G_\approximate(u)=g(u),\qquad \G_\approximate(v) = h(y_{\inndex_v})  (x-y_{\inndex_v}) + g(y_{\inndex_v}), \qquad \text{and}\\
&\nonumber
\G_\approximate(w)= \Big( 1 -\eta \Big(\frac{2\approximate|w-y_{\inndex_w}| - \delta}{\delta}\Big)\Big)  \big(h(y_{\inndex_w}) (w-y_{\inndex_w}) + g(y_{\inndex_w})\big) + \eta\Big( \frac{2\approximate|w-y_{\inndex_w}| - \delta}{\delta} \Big) g(w).
\end{align}
Then it holds for all $m\in \N$ that
\begin{equation}\label{it3:existence_approxsequence}
\sup_{\approximate \in \N} \sup_{x \in [-m,m]} \big( |\G_\approximate(x)|  + |(\G_\approximate)'(x)| \big) <\infty.
\end{equation}
\end{athm}
\begin{aproof}
    \argument{the assumption that $g$ is continuous; the assumption that $\mathscr{d}_\func g$ is locally bounded}{that  for all $m\in \N$ it holds that
\begin{equation}\llabel{eq:g_itself_dominated}
\sup_{x \in [-m,m]} \big( |g(x)| + |(\mathscr{d}_\func  g)(x)|\big) <\infty\dott
\end{equation}}
\argument{\lref{eq:g_itself_dominated};\lref{assume}}{for all $m\in \N$ that
\begin{equation}\label{eq:4dom_noOcase}
\sup_{\approximate \in \N} \sup_{x \in [-m,m] \backslash O_{\delta \approximate^{-1}}} \big(|\G_\approximate(x)| + |(\G_\approximate)'(x)|\big)<\infty\dott
\end{equation}}
\argument{\lref{assume};}{ for all $\approximate \in \N$, $x \in \overline{O_{\delta (2\approximate)^{-1}}}$ that
\begin{equation}\label{eq:4dom_g_O2r}
\big|\G_\approximate(x) \big| = \big| \gamma_{\inndex_x} (x-y_{\inndex_x}) + g(y_{\inndex_x}) \big| \leq \frac{\delta}{2\approximate} \max_{i \in \{1,2,\dots,\element\}} |\gamma_{i}| + \max_{i \in \{1,2,\dots,\element\}} | g(y_i)|  
\end{equation}
and
\begin{equation}\llabel{eq:4dom_dg_O2r}
|(\G_\approximate)'(x)| = | \gamma_{\inndex_x}| \leq \max_{i \in \{1,2,\dots,\element\}} |\gamma_i|. 
\end{equation}}
\argument{\lref{eq:4dom_dg_O2r};}{for all $m\in \N$ that
\begin{equation}\label{eq:4dom_O2r}
\sup_{\approximate \in \N} \sup_{x \in [-m,m] \cap \overline{O_{\delta (2\approximate)^{-1}}}} \big(|\G_\approximate(x)| + |(\G_\approximate)'(x)|\big)<\infty.
\end{equation}}
\argument{\lref{def: eta};}{ for all $\approximate \in \N$, $x \in O_{\delta \approximate^{-1}} \backslash \overline{O_{\delta (2\approximate)^{-1}}}$ that
\begin{equation}\label{eq:4dom_g_Or}
\begin{split}
\big| \G_\approximate(x)  \big|  &= \Big| \Big( 1 -\eta \Big(\frac{2\approximate|x-y_{\inndex_x}| - \delta}{\delta}\Big)\Big)  \big(\gamma_{\inndex_x} (x-y_{\inndex_x})+g(y_{\inndex_x}) \big)\\ 
&\qquad\quad+ \eta\Big( \frac{2\approximate|x-y_{\inndex_x}| - \delta}{\delta} \Big) g(x)\Big| \\
&\leq\big|\gamma_{\inndex_x} (x-y_{\inndex_x})+g(y_{\inndex_x})\big| + \big| g(x) \big| \\
&\leq |g(x)|+\frac{\delta}{\approximate} \biggl[\max_{i \in \{1,2,\dots,\element\}} |\gamma_i|\biggr] + \Bigl[\max_{i \in \{1,2,\dots,\element\}} |g (y_i)| \Bigr]
\end{split} 
\end{equation}
and 
\begin{equation}\label{eq:4dom_dg_Or}
\begin{split}
\big| (\G_\approximate)'(x) \big|  &= \biggl|\frac{2\approximate}{\delta}(-1)^{\indicator{(0,\infty)}{(y_{\inndex_x}-x)}} \eta'\Big(\frac{2\approximate|x-y_{\inndex_x}|-\delta}{\delta}\Big) \big(g(x)-g(y_{\inndex_x}) -\gamma_{\inndex_x} (x-y_{\inndex_x}) \big) \\
&+ \Big( 1- \eta\Big(\frac{2\approximate(x-y_{\inndex_x})-\delta}{\delta}\Big)\Big) \gamma_{\inndex_x}+ g'(x) \eta\Big(\frac{2\approximate|x-y_{\inndex_x}|-\delta}{\delta}\Big)\bigg| \\
&\leq \biggl[\sup_{y \in (0,1)}|\eta'(y)|\biggr] \frac{2\approximate}{\delta} \Big( \big| g(x)-g(y_{\inndex_x})\big| + \Bigl[\max_{i \in \{1,2,\dots,\element\}} |\gamma_i|\Bigr] |x-y_{\inndex_x}|\Big) \\
&+ \biggl[\sup_{y \in (0,1)} |\eta(y)|\biggr] \Big( \Bigl[\max_{i \in \{1,2,\dots,\element\}} |\gamma_i|\Bigr] + |g'(x)| \Big) \\
&\leq 2\biggl[\sup_{y \in (0,1)} |\eta'(y)|\biggl] \Big(\frac{|g(x)-g(y_{\inndex_x})|}{|x-y_{\inndex_x}|} +\Bigl[ \max_{i \in \{1,2,\dots,\element\}} |\gamma_i|\Bigr]\Big) \\
&+ \biggl[\sup_{y \in (0,1)} |\eta(y)|\biggr] \biggl( \Bigl[\max_{i \in \{1,2,\dots,\element\}} |\gamma_i|\Bigr] + |g'(x)| \biggr)\dott
\end{split}
\end{equation}}
\argument{the assumption that $\restr{g}{\R \backslash  \{y_1,\ldots,y_\element\}} \in C^1(\R \backslash  \{y_1,y_2,\ldots,y_\element\},\R)$; the assumption that $g$ is continuous}{that \llabel{arg3} $g$ is locally Lipschitz continuous\dott}
\argument{\lref{arg3};} {that there exists $L \in (0,\infty)$, $R \in \N$ such that for all $\approximate \in \N \cap [R,\infty)$, $x \in O_{\delta \approximate^{-1}} \backslash \overline{O_{\delta(2\approximate)^{-1}}}$ it holds that
\begin{equation}\llabel{eq3}
\frac{|g(x)-g(y_{\inndex_x})|}{|x-y_{\inndex_x}|} \leq L\dott
\end{equation}}
\argument{\lref{eq3};\lref{eq:g_itself_dominated};\cref{eq:4dom_g_Or};\cref{eq:4dom_dg_Or}}{for all $m\in \N$ that
\begin{equation}\llabel{eq:4dom_OrminusO2r}
\sup_{\approximate \in \N} \sup_{x \in [-m,m] \cap (O_{\delta \approximate^{-1}}\backslash\overline{O_{\delta (2\approximate)^{-1}}})} \big(|\G_\approximate(x)| + |(\G_\approximate)'(x)|\big)<\infty.
\end{equation}}
\argument{\cref{eq:4dom_noOcase}; \cref{eq:4dom_O2r}; \lref{eq:4dom_OrminusO2r}}{\cref{it3:existence_approxsequence}\dott}
\end{aproof}
\begin{remark}
Note that $\eta \in C^1(\R)$ which satisfies $\eta(0)=\eta'(0)=\eta'(1)=0$, $\eta(1)=1$ and $\eta((0,1))\subseteq [0,1]$ exists. Indeed, consider for instance the function $\eta \colon \R \to \R$ which is for all $x \in \R$ given by
\begin{equation}
\eta (x) = \begin{cases}
\mathrm{exp}\Big(1-\frac{1}{1-(x-1)^2} \Big) &\colon x \in (0,2) \\
0 &\colon x \notin (0,2).
\end{cases}
\end{equation}
\end{remark}
\begin{remark}
Let $g \colon \R \to \R$ be given by
\begin{equation}
g(x) = \begin{cases}
x \sin\big(x^{-1} \big) &\colon x>0\\
0 &\colon x\leq 0.
\end{cases}
\end{equation}
Then it holds for all $x\in \R$ that $|g(x)|\leq |x|$. This and the fact that $\restr{g}{\R \backslash \{0\}} \in C^1(\R \backslash \{0\},\R)$ demonstrate that $g$ is locally Lipschitz continuous. However, for all $x>0$ it holds that
\begin{equation}\label{eq:sin_counter_derivative}
g'(x) = \sin\big(x^{-1}\big) - \frac{\cos\big(x^{-1}\big)}{x}. 
\end{equation} 
For every $k \in \N$ let  $x_k \in \R$ satisfy $x_k=(k\pi)^{-1}$. Then \cref{eq:sin_counter_derivative} shows that
$
|g'(x_k)| = k\pi
$ and consequently it holds that $\lim_{k \to \infty} |g'(x_k)| =\infty$.
 This demonstrates that in \cref{it3:existence_approxsequence} of \cref{lem:existence_approxsequence2} the assumption that $g$ is locally Lipschitz continuous does not imply in general that it holds for all $M\in \N$ that 
\begin{equation}\label{eq:sup_der_condition}
\sup_{x \in [-M,M] } |(\mathscr{d}_\func g)(x)| < \infty.
\end{equation} 
Moreover   \cref{eq:sup_der_condition} does not imply in general that $g$ is continuous. Hence the condition that for all $M\in \N$ it holds \cref{eq:sup_der_condition} does not imply in general that $g$ is locally Lipschitz continuous.
\end{remark}

\begin{athm}{cor}{lem: existence of A_n}
    Let $S\subseteq\R$ be finite, let $\scrA\in C(\R,\R)$ satisfy $\restr{\scrA}{\R \backslash S} \in C^1(\R \backslash S,\R)$, let $\scra \colon \R \to \R$ satisfy $\restr{\scra}{\R \backslash S}=\nabla(\restr{\scrA}{\R \backslash S})$, and let $\ZZ\in \N_0$ satisfy for all $m\in \N$ that $\sup_{x\in [-m,m]} \bigl[\ZZ|\scra(x)|\bigr]\allowbreak<\infty$. Then there exist $A_n \in C^1(\R,\R)$, $n\in \N$, such that
    \begin{enumerate}[label=(\roman*)]
        \item it holds for all $ x \in \R$ that there exists $m \in \N$ such that $\sum_{n=m}^\infty (|\scrA(x)-A_n(x)|+|\scra(x)-(A_n)'(x)|) = 0$ and
        \item it holds for all $m \in \N $ that $\sup_{n \in \N} \sup_{x \in [-m,m]} \bigl[\ZZ( |A_n(x)|  + |(A_n)'(x)| )\bigr] \allowbreak<\infty$.
 \end{enumerate}
\end{athm}
\begin{aproof}
Throughout this proof assume without loss of generality that $\ZZ>0$ (otherwise \cref{lem:existence_approxsequence} proves that there exist $(A_n)_{n\in \N} \subseteq C^1(\R,\R)$ which satisfy $\forall \, x \in \R \colon \exists \, m \in \N \colon \sum_{n=m}^\infty (|\scrA(x)-A_n(x)|+|\scra(x)-(A_n)'(x)|) = 0$).
\argument{the assumption that $\ZZ>0$; the assumption that for all $m\in \N$ it holds that $\sup_{x\in [-m,m]} [\ZZ\allowbreak|\scra(x)|]<\infty$}{that \llabel{arg0} $\scra$ is locally bounded\dott}
    \argument{the fact that $\restr{\scra}{\R \backslash S}=\nabla(\restr{\scrA}{\R \backslash S})$; \lref{arg0}}{that \llabel{arg1} $\scrA$ is locally Lipschitz continuous\dott}
    \argument{\lref{arg1};\cref{lem:existence_approxsequence};\cref{lem:existence_approxsequence2}; the fact that $\restr{\scra}{\R \backslash S}=\nabla(\restr{\scrA}{\R \backslash S})$; the fact that $\scra$ is locally bounded}{that there exists $A_n\colon \R\to\R$ such that
    \begin{enumerate}[label=(\Roman*)]
        \item it holds for all $n \in \N$ that $A_n \in C^1(\R,\R)$,
\item it holds for all $x \in \N$ that there exists $R \in \N$ which satisfies for all $n \in \N \cap [N , \infty)$ that
\begin{equation}
A_n(x)=\scrA(x) \qandq (A_n)'(x) = \scra(x),
\end{equation} and
\item it holds for all $m\in \N$ that
\begin{equation}
\sup_{n \in \N} \sup_{x \in [-m,m]} \big( |A_n(x)|  + |(A_n)'(x)| \big) <\infty.
\end{equation}
    \end{enumerate}}
\end{aproof}
\begin{athm}{lemma}{scrA is continuous}
        Let $S\subseteq\R$ be finite, let $\scrA\colon \R\to\R$ satisfy $\restr{\scrA}{\R \backslash S} \in C^1(\R \backslash S,\R)$, let $\scra \colon \R \to \R$ satisfy $\restr{\scra}{\R \backslash S}=\nabla(\restr{\scrA}{\R \backslash S})$, and let $(A_n)_{n\in \N} \subseteq C^1(\R,\R)$ satisfy $\forall \, x \in \R \colon \exists \, m \in \N \colon \sum_{n=m}^\infty (|\scrA(x)-A_n(x)|+|\scra(x)-(A_n)'(x)|) = 0$ and $\forall \,m \in \N \colon \sup_{n \in \N} \sup_{x \in [-m,m]} (
        \allowbreak|A_n(x)|  + |(A_n)'(x)| ) <\infty$. Then 
        \begin{enumerate}[label=(\roman*)]
            \item \llabel{item 1} it holds that $\scra$ is locally bounded and
            \item \llabel{item 2} it holds that $\scrA\in C(\R,\R)$.
        \end{enumerate}
\end{athm}
\begin{aproof}
\argument{the fact that $\forall \, x \in \R \colon \exists \, m \in \N \colon \sum_{n=m}^\infty (|\scrA(x)-A_n(x)|+|\scra(x)-(A_n)'(x)|) = 0$;}{that for all $x\in \R$ it holds that \llabel{arg1} there exists $m\in \N$ such that for all $n\in \N\cap[m,\infty)$ it holds that $(A_n)'(x)=\scra(x)$\dott}
    \argument{\lref{arg1};the fact that for all $m\in \N$ it holds that $\sup_{n \in \N} \sup_{x \in [-m,m]} ( |A_n(x)|  + |(A_n)'(x)| ) <\infty$}{that for all $m\in \N$ it holds that
    \begin{equation}\llabel{eq1}
        \sup_{x\in [-m,m]} |\scra(x)|\leq \sup_{n \in \N} \sup_{x \in [-m,m]}  |(A_n)'(x)| \leq \sup_{n \in \N} \sup_{x \in [-m,m]} ( |A_n(x)|  + |(A_n)'(x)| )<\infty\dott
    \end{equation}}
    \argument{\lref{eq1};}[verbs=ep]{\lref{item 1}\dott}
    \startnewargseq
    \argument{\lref{item 1}; the fact that $\restr{\scra}{\R \backslash S}=\nabla(\restr{\scrA}{\R \backslash S})$}[verbs=ep]{\lref{item 2}\dott}
\end{aproof}
\begin{athm}{cor}{cor: equivalent sequence}
    Let $S\subseteq\R$ be finite, let $\scrA\colon \R\to\R$ satisfy $\restr{\scrA}{\R \backslash S} \in C^1(\R \backslash S,\R)$, and let $\scra \colon \R \to \R$ satisfy $\restr{\scra}{\R \backslash S}=\nabla(\restr{\scrA}{\R \backslash S})$. Then the following two statements are equivalent:
    \begin{enumerate}[label=(\roman*)]
        \item  \label{item 2: equivalent} There exist $A_n \in C^1(\R,\R)$, $n\in \N$, such that $\forall \, x \in \R \colon \exists \, m\in \N \colon \sum_{n=m}^\infty (|\scrA(x)-A_n(x)|+|\scra(x)-(A_n)'(x)|) = 0$ and $\forall \,m \in \N \colon \sup_{n \in \N} \sup_{x \in [-m,m]} (
        \allowbreak|A_n(x)|  + |(A_n)'(x)| ) <\infty$.
         \item \label{item 1: equivalent} It holds that $\scrA\in C(\R,\R)$ and that $\scra(x)$ is locally bounded.
    \end{enumerate}
\end{athm}
\begin{aproof}
    \argument{\cref{lem: existence of A_n};}{that (\ref{item 1: equivalent} $\rightarrow$ \ref{item 2: equivalent})\dott}
    \startnewargseq
    \argument{\cref{scrA is continuous};}{that (\ref{item 2: equivalent} $\rightarrow$ \ref{item 1: equivalent})\dott}
\end{aproof}
%\begin{remark}
%Let $g \colon \R \to \R$ be given by
%\begin{equation}
%g(x) = \begin{cases}
%x^2 \sin\big(x^{-2} \big) &\colon x>0\\
%0 &\colon x\leq 0
%\end{cases}.
%\end{equation}
%Then it holds that $D_g^{-}=D_g^{+} = \R$. However, for all $x>0$ it holds that
%\begin{equation}
%g'(x) = 2x \sin\big(x^{-2}\big) - \frac{\cos\big(x^{-2}\big)}{2x}. 
%\end{equation} 
%Consequently, $\sup_{x>0} |g'(x)|=\infty$. This demonstrates that in \cref{it3:existence_approxsequence} of \cref{lem:existence_approxsequence} the assumption $D_g^{-}=D_g^{+}=\R$ does not imply in general that it holds for $M>0$ that 
%\begin{equation}\label{eq:sup_der_condition}
%\sup_{x \in [-M,M] \backslash \{y_1,\ldots,y_n\}} |g'(x)| < \infty.
%\end{equation} 
%Moreover that continuity of a function is not a necessary condition for the property that for all $M>0$ it holds \cref{eq:sup_der_condition}. Hence the condition that for all $M>0$ it holds \cref{eq:sup_der_condition} does not imply in general that $D_g^{-} = D_g^{+}=\R$.
%\end{remark}

\subsection{Representations for generalized gradients of the objective function}\label{ssec:rep_formula_gengrads}

\begin{athm}{prop}{prop:gcNconvergence}
Assume \cref{setting:activationapprox} and let $x \in \R^{\ell_0}$, $\theta \in \R^{\fd}$. Then
\begin{enumerate}[label=(\roman*)]
\item\label{it1:gcNconvergence} it holds for all $K \in \N$, $i \in \{1,2,\dots,\ell_K\}$ that there exists $m \in \N$ which satisfies for all $\approximate \in \N \cap [m,\infty)$ that
\begin{equation}
\cN_{\approximate,i}^{K,\theta}(x) = \cN_{0,i}^{K,\theta}(x)
\end{equation}
and
\item\label{it2:gcNconvergence} it holds for all $K \in \N$, $i \in  \{1,2,\dots,\ell_K\}$ that there exists $R \in \N$ which satisfies for all $\approximate \in \N \cap [R,\infty)$ that
\begin{equation}
\fg_\approximate \big( \cN_{\approximate,i}^{K,\theta}(x)\big) = \fg_0\big( \cN_{0,i}^{K,\theta}(x)\big) \qqandqq (\fg_\approximate)'\big(\cN_{\approximate,i}^{K,\theta}(x)\big) = \big(\mathscr{d}_\func \fg_0 \big)\big( \cN_{0,i}^{K,\theta}(x) \big)
\end{equation}
\end{enumerate}
\cfadd{def:LKB_left_right_derivative}\cfout.
\end{athm}

\begin{aproof}
 We prove \cref{it1:gcNconvergence} by induction on $K \in \N$. For the base case $K=1$ note that \cref{eq:def_NN_realization} proves  for all $\approximate \in \N$ that 
 \begin{equation}\llabel{base}
     \cN_\approximate^{1,\theta}(x) = \cN_0^{1,\theta}(x).
 \end{equation} This establishes \cref{it1:gcNconvergence} in the base case $K=1$. For the induction we assume that there exists $K \in \N$ which satisfies that there exists $\widetilde{R} \in \N$ which satisfies for all $\approximate \in \N \cap [\widetilde{R},\infty)$ that 
\begin{equation}\label{eq:gcN_IV}
\cN_{\approximate,j}^{K,\theta}(x) = \cN_{0,j}^{K,\theta}(x).
\end{equation}
\argument{\cref{eq:def_NN_realization};}{that for all $i \in \{1,2,\dots,\ell_{K+1}\}$, $\approximate\in \N$ it holds that
\begin{equation}\label{eq:cNrK+1formula}
\cN_{\approximate,i}^{K+1,\theta}(x) = \fb_i^{K+1,\theta} + \sum_{j=1}^{\ell_{K}} \fw_{i,j}^{K+1,\theta} \fg_\approximate\big( \cN_{\approximate,j}^{K,\theta}(x) \big).
\end{equation}}
\argument{\cref{def: g_r}}{ for all $j \in \{1,2,\dots,\ell_K\}$ that there exists $\widehat{R}_j \in \N$ which satisfies for all $\approximate \in \N \cap [\widehat{R}_j,\infty)$ that
\begin{equation}\llabel{eq:gcN_onKlayer}
\fg_\approximate\big( \cN_{0,j}^{K,\theta}(x) \big) =  \fg_0\big( \cN_{0,j}^{K,\theta}(x) \big).
\end{equation}}
\argument{\cref{eq:gcN_IV};\lref{eq:gcN_onKlayer};\cref{eq:cNrK+1formula}}{that for all $i \in \{1,2,\dots,\ell_{K+1}\}$, $\approximate \in \N \cap [\max \big\{ \widetilde{R} , \allowbreak\max_{j \in \{1,2,\dots,\ell_K\}} \widehat{R}_j\bigr\},\infty)$ it holds that
\begin{equation}\llabel{eq1}
\cN_{\approximate,i}^{K+1,\theta}(x) = \fb_i^{K+1,\theta} + \sum_{j=1}^{\ell_{K}} \fw_{i,j}^{K+1,\theta} \fg_0\big( \cN_{0,j}^{K,\theta}(x) \big) = \cN_{0,i}^{K+1,\theta}(x).
\end{equation} }
\argument{\lref{eq1};\lref{base};induction}{ \cref{it1:gcNconvergence}\dott} 
\startnewargseq
\argument{\cref{it1:gcNconvergence};}{\cref{it2:gcNconvergence}\dott}
\end{aproof}

\begin{athm}{lemma}{lem:derivsofcN}
 Assume \cref{setting:activationapprox} and let
 $\approximate \in \N$, $x=(x_1,\ldots,x_{\ell_0}) \in \R^{\ell_0}$, $\theta=(  \theta_1,\ldots,\theta_\fd)\in \R^{\fd}$. 
  Then
  \begin{enumerate}[label=(\roman*)]
  \item 
  \label{it1:derivsofcN}
  it holds for all 
  $ k \in \{1,2,\dots,L\}  $
  that $(\R^\fd\times\R^{\ell_0}\ni(\vartheta,y)\mapsto \mathcal N_n^{k,\vartheta}(y)\in \R^{\ell_k})\in C^1(\R^\fd\times\R^{\ell_0},\R^{\ell_k})$,
  \item
  \label{it2:derivsofcN}
  it holds for all 
  $K \in \N $,
  $ k \in \{1,2,\dots,K\}$,
  $ i \in \{1,2,\dots,\ell_k\}$,
  $ j \in \{1,2,\dots,\ell_{k-1}\}$,
  $ h \in \{1,2,\dots,\ell_K\}$
  that
  \begin{equation}
  \label{N_der1}
  \begin{multlined}
    \frac{ \partial }{ \partial\theta_{ (i-1) \ell_{k-1} + j + \mathbf{d}_{k-1}} 
    }
    \bigl( 
     \cN_{\approximate,h}^{K,\theta}(x)
    \bigr)
\\
    = 
    \sum_{
      \substack{v_k,v_{k+1}, \ldots,v_K \in \N, \\ \forall w\in \N\cap[k, K ]\colon v_w\leq\ell_w} 
    }
    \Bigl[
  \fg_{\approximate}  \big(\cN_{\approximate,j}^{\max\{k-1,1\},\theta}(x)\big) \indicator{ (1, K ]}(k)+    
      x_j\indicator{\{1\} }(k)
    \Bigr]
\\
    \cdot
    \Bigl[ \indicator{ \{ i \} }( v_k ) \Bigr]
    \Bigl[\indicator{ \{ h \} }( v_K ) \Bigr]
    \Bigl[
      \textstyle{\prod}_{p=k+1}^K
      \big( \fw^{p, \theta }_{v_p,v_{p-1} } 
      (\fg_\approximate)'\big(\cN_{\approximate,v_{p-1}}^{p-1,\theta}(x) \big)
      \big)
    \Bigr],
  \end{multlined}  
  \end{equation}
  and
  \item
  \label{it3:derivsofcN}
  it holds for all 
  $K \in \N $,
  $ k \in \{1,2,\dots,K\}$,
  $ i \in \{1,2,\dots,\ell_k\}$,
  $ h \in \{1,2,\dots,\ell_K\}$
  that
    \begin{equation}
  \label{N_der2}
  \begin{split}
  &
    \frac{ \partial }{ \partial \theta_{ \ell_k \ell_{ k - 1 } + i + \mathbf{d}_{ k - 1 } } 
    }
    \bigl( 
     \cN_{\approximate,h}^{k,\theta}(x) 
    \bigr)
\\
    &= \sum_{\substack{v_k,v_{k+1}, \ldots,v_K \in \N, \\ \forall w\in \N\cap[k, K ]\colon v_w\leq\ell_w} }
    \Big[\indicator{\{ i \} }(v_k)\Big]
    \Big[\indicator{\{ h \} }(v_K)\Big]
\\   
  &\qquad\qquad  \Big[\textstyle{\prod}_{p=k+1}^K
    \big( \fw^{p, \theta }_{v_p,v_{p-1} } 
   (\fg_\approximate)'(\cN_{\approximate,v_{p-1}}^{p-1,\theta}(x))
    \big)\Big].
  \end{split}  
  \end{equation}
  \end{enumerate}
\end{athm}
\begin{aproof}
\argument{\cref{eq:def_NN_realization}; the fact that for all $\approximate\in\N$ it holds that $ \fg_\approximate \in C^1(\R,\R)$}{\cref{it1:derivsofcN}\dott}
We prove \cref{it2:derivsofcN} and \cref{it3:derivsofcN} by induction on $K\in \N $. For the base case $K=1$ note that for all  $i,h \in \{1,2,\dots,\ell_1\}$, $j\in \{1,2,\dots,\ell_0\} $ it holds that
\begin{equation}\llabel{base}
\frac{\partial}{\partial \theta_{(i-1)\ell_0 + j}} \cN_{\approximate,h}^{1,\theta}(x) = x_j \indicator{\{h\}}(i) \qqandqq \frac{\partial}{\partial \theta_{\ell_0 \ell_1 + i}} \cN_{\approximate,h}^{1,\theta}(x) = \indicator{\{h\}}(i).
\end{equation}
This establishes \cref{it2:derivsofcN} and \cref{it3:derivsofcN} in the base case $K=1$.
For the induction step we assume that there exists $K \in \N $ which satisfies for all
$ k \in \{1,2,\dots,K\} $,
  $ i \in \{1,2,\dots,\ell_k\} $,
  $ j \in \{1,2,\dots,\ell_{k-1}\} $,
  $ h \in \{1,2,\dots,\ell_K\} $
  that
  \begin{equation}
  \label{N_der11}
  \begin{multlined}
    \frac{ \partial }{ \partial\theta_{ (i-1) \ell_{k-1} + j + \mathbf{d}_{k-1}} 
    }
    \bigl( 
     \cN_{\approximate,h}^{K,\theta}(x)
    \bigr)
\\
    = 
    \sum_{
      \substack{v_k,v_{k+1}, \ldots,v_K \in \N, \\ \forall w\in \N\cap[k, K ]\colon v_w\leq\ell_w} 
    }
    \Bigl[
\fg_\approximate \big( \cN_{\approximate,j}^{\max\{k-1,1\},\theta}(x)\big) \indicator{ (1, K ]}(k)+    
      x_j\indicator{\{1\} }(k)
    \Bigr]
\\
    \cdot
    \Bigl[ \indicator{ \{ i \} }( v_k ) \Bigr]
    \Bigl[\indicator{ \{ h \} }( v_K ) \Bigr]
    \Bigl[
      \textstyle{\prod}_{p=k+1}^K
      \big( \fw^{p, \theta }_{v_p,v_{p-1} } 
      (\fg_\approximate)'\big( \cN_{\approximate,v_{p-1}}^{p-1,\theta}(x) \big)
      \big)
    \Bigr]
  \end{multlined}  
  \end{equation}
  and
    \begin{equation}
  \label{N_der22}
  \begin{split}
  &
    \frac{ \partial }{ \partial \theta_{ \ell_k \ell_{ k - 1 } + i + \mathbf{d}_{ k - 1 } } 
    }
    \bigl( 
      \cN_{\approximate,h}^{K,\theta}(x)
    \bigr)
\\
    &= \sum_{\substack{v_k,v_{k+1}, \ldots,v_K \in \N, \\ \forall w\in \N\cap[k, K ]\colon v_w\leq\ell_w} }
    \Big[\indicator{\{ i \} }(v_k)\Big]
    \Big[\indicator{\{ h \} }(v_K)\Big]
    \Big[\textstyle{\prod}_{p=k+1}^K
    \big( \fw^{p, \theta }_{v_p,v_{p-1} } 
    (\fg_\approximate)'\big( \cN_{\approximate,v_{p-1}}^{p-1,\theta}(x)\big)
    \big)\Big].
  \end{split}  
  \end{equation}
  \startnewargseq
  \argument{\cref{relization multi};\cref{N_der11}}{that for all
$ k \in \{1,2,\dots,K\} $,
$ i \in \{1,2,\dots,\ell_k\} $, $ j \in \{1,2,\dots,\ell_{k-1}\} $,
$ h \in \{1,2,\dots,\ell_{K+1}\} $
it holds that
\begin{equation}\llabel{eq1}
\begin{split}
    &\frac{\partial}{\partial\theta_{ (i-1)\ell_{k-1} + j+ \mathbf{d}_{k-1} } }
    \bigl( \cN_{\approximate,h}^{K+1,\theta}(x)\bigr)\\
    &=
    \frac{ 
      \partial 
    }{ 
      \partial\theta_{ (i-1) \ell_{ k - 1 } + j + \mathbf{d}_{ k - 1 } } 
    }
    \biggl( 
      \fb^{ K + 1, \theta }_h 
      + 
      \sum_{ l = 1 }^{ \ell_K } 
      \fw^{ K+1, \theta }_{ h, l} 
\fg_\approximate  \big( 
        \cN_{\approximate,l}^{K,\theta}( x)
      \big)
    \biggr)
\\
&
  =
  \sum_{ l = 1 }^{ \ell_{K} }
  \Biggl[
    \fw^{ K + 1, \theta }_{ h, l }
    \bigl[ 
 (\fg_\approximate)' (\cN_{\approximate,l}^{K,\theta}(x) 
      )
    \bigr]
    \biggl( 
      \frac{ \partial }{ \partial\theta_{ (i-1) \ell_{ k - 1 } + j + \mathbf{d}_{k-1} } 
      }
      \bigl( 
       \cN_{\approximate,l}^{K,\theta}(x) 
      \bigr)
    \biggr)
  \Biggr]
\\  
&= 
  \sum_{ l = 1 }^{ \ell_K }
  \Biggl[
    \fw^{ K+1, \theta }_{ h, l }
    \bigl[
      (\fg_\approximate)'( \cN_{\approximate,l}^{K,\theta}(x)  )
    \bigr]
\\
&\quad\cdot
  \sum_{
    \substack{ v_k, v_{k+1}, \ldots, v_K \in \N, 
    \\ 
    \forall w \in \N\cap[k,K]\colon v_w\leq\ell_w} }
    \Big[\fg_\approximate\big( \cN^{\max\{k-1,1\},\theta}_{\approximate,j} (x)\big)\indicator{ (1,K]}(k)
    +x_j \indicator{\{1\} }(k)
    \Big] \\
    &\quad\cdot
    \Bigl[ \indicator{ \{ i \} }( v_k ) \Bigr]
    \Bigl[ \indicator{ \{ l \} }( v_K ) \Bigr]
    \Big[\textstyle{\prod}_{p=k+1}^K
    \big( \fw^{p, \theta }_{v_p,v_{p-1} } 
    \big[(\fg_\approximate)'\big(\cN_{\approximate,v_{p-1}}^{p-1,\theta}(x))\big]
    \big)\Big] \Bigg] \\
    &=\sum_{\substack{v_k,v_{k+1}, \ldots,v_{K+1} \in \N, \\\forall w\in \N\cap[k,K+1]\colon v_w\leq\ell_w} }
    \Big[\fg_\approximate \big( \cN_{\approximate,j}^{\max\{k-1,1\},\theta}(x) \big) \indicator{ (1,K]}(k)
    +x_j \indicator{\{1\} }(k)
    \Big]    
    \\
    &\quad\cdot
    \Big[\indicator{\{ i \} }(v_k )\Big]
    \Big[\indicator{\{ h  \} }(v_{K+1} )\Big]    
    \Big[\textstyle{\prod}_{p = k + 1}^{K+1}
    \big( \fw^{p, \theta }_{v_ p,v_{p-1} } \big[(\fg_\approximate)' \big(\cN_{\approximate,v_{p-1}}^{p-1,\theta}(x)\big)\big]
    \big)\Big]
  .
\end{split}
\end{equation}}
\argument{\cref{relization multi};}{for all
$ i,h \in \{1,2,\dots,\ell_{K+1}\}$, 
$ j \in \{1,2,\dots,\ell_K\} $
that
\begin{equation}\llabel{eq2}
\begin{split}
    \frac{\partial}{\partial\theta_{ (i-1)\ell_{K} + j + \mathbf{d}_K } }( \cN_{\approximate,  h  }^{K+1, \theta }(x))
    &= \frac{\partial}{\partial\theta_{ (i-1)\ell_{K} + j + \mathbf{d}_K } }
    \biggl( 
      \fb^{ K + 1, \theta }_h 
      + 
      \sum_{ l = 1 }^{ \ell_K } 
      \fw^{ K+1, \theta }_{ h, l} 
 \fg_\approximate  \big( \cN_{\approximate,l}^{K,\theta}( x) \big)
    \biggr)
\\
&
  =
 \fg_\approximate \big(  \cN_{\approximate,l}^{K,\theta}( x)
 \big) \indicator{ \{ h \} }( i ) .
\end{split}
\end{equation}}
\argument{\cref{relization multi};\cref{N_der22}}{for all
$ k \in \{1,2,\dots,K\}$, 
$ i \in \{1,2,\dots,\ell_k\}$, 
$  h  \in \{1,2,\dots,\ell_{K+1}\}$ that
\begin{equation}\llabel{eq3}
\begin{split}
&
  \frac{ \partial }{ \partial\theta_{ \ell_k \ell_{ k - 1 } + i + \mathbf{d}_{ k - 1 } } 
  }\bigl( 
  \cN_{\approximate,h}^{K+1,\theta}(x)  
  \bigr)
  = 
  \frac{ \partial }{ \partial\theta_{ \ell_k \ell_{ k - 1 } + i + \mathbf{d}_{ k - 1 } } }
  \biggl( 
    \fb^{ K + 1, \theta }_h 
    + 
    \sum_{ 
      l= 1 
    }^{ 
      \ell_K 
    } 
    \fw^{ K + 1, \theta }_{ h, l } 
\fg_\approximate \big( \cN_{\approximate,l}^{K,\theta}(x) \big)
  \biggr)
\\
    &=\sum_{ l = 1 }^{ \ell_{K} }
    \fw^{K+1, \theta }_{ h, l }
    \big[(\fg_\approximate)'\big(\cN_{\approximate,l}^{K,\theta}(x)\big) \big]
    \bigg( \frac{\partial}{\partial\theta_{\ell_k\ell_{k-1} + i +\mathbf{d}_{k-1} } }(\cN_{\approximate,l}^{K,\theta}(x))\bigg)\\  
    &= \sum_{ l= 1}^{ \ell_{K} }
    \Biggl[ 
      \fw^{K+1, \theta }_{ h, l }
      \bigl[
         (\fg_\approximate)'\big( \cN_{\approximate,l}^{K,\theta}(x) \big) 
      \bigr]
    \sum_{
      \substack{ v_k, v_{k+1}, \dots, v_K \in \N, 
      \\
      \forall w \in \N \cap [k,K] \colon v_w \leq \ell_w } 
    }
    \Bigl[
      \indicator{ \{ i \} }( v_k ) 
    \Bigr]
    \Bigl[
      \indicator{ \{ l \} }( v_K ) 
    \Bigr]
\\
    &\quad\cdot \Big[\textstyle{\prod}_{p=k+1}^K
    \big( \fw^{p, \theta }_{v_p,v_{p-1} } \big[(\fg_\approximate)'\big(\cN_{\approximate,v_{p-1}}^{p-1,\theta}(x)\big)\big]
    \big)\Big]
    \Bigg] \\
    &=
    \sum_{\substack{v_k,v_{k+1}, \ldots,v_{K+1} \in \N, \\\forall w\in \N\cap[k,K+1]\colon v_w\leq\ell_w} }
    \Big[\indicator{\{ i \} }(v_k )\Big]
    \Big[\indicator{\{ h  \} }(v_{K+1} )\Big]
    \Big[\textstyle{\prod}_{p=k+1}^{K+1}
    \fw^{p, \theta }_{v_p,v_{p-1} } 
    \big[(\fg_\approximate)'\big( \cN_{\approximate,v_{p-1}}^{p-1,\theta}(x)\big)\big]
    \Big] .
\end{split}
\end{equation}}
\argument{\cref{relization multi};}{for all
$ i,h \in \{1,2,\dots,\ell_{K+1}\} $ 
that
\begin{equation}\llabel{eq4}
\begin{split}
  \frac{ \partial }{ 
    \partial \theta_{ \ell_{ K + 1 } \ell_K + i + \mathbf{d}_K } 
  }( 
   \cN_{\approximate,h}^{K+1,\theta}(x) 
  )
&
  = \frac{\partial}{\partial\theta_{\ell_{K+1} \ell_{K} + i+\mathbf{d}_K } }
    \bigg( \fb^{K+1, \theta }_ h  + \sum_{ l = 1 }^{ \ell_K} \fw^{K+1, \theta }_{ h,l }  \fg_\approximate\big(\cN_{\approximate,l}^{K,\theta}(x)\big)\bigg)\\
&
  = \indicator{ \{ h \} }( i ) .
\end{split}
\end{equation}  }
\startnewargseq
\argument{\lref{base};\lref{eq1};\lref{eq2};\lref{eq3};\lref{eq4};induction}{\cref{it2:derivsofcN} and \cref{it3:derivsofcN}\dott}
\end{aproof}

\begin{athm}{lemma}{lem:uniformboundsforNandderN}
Assume \cref{setting:activationapprox} and let $K \subseteq \R^\fd$ and $V \subseteq \R^{\ell_0}$ be compact. Then 
\begin{enumerate}[label=(\roman*)]
\item\label{it1:uniformbounds} it holds for all  $k \in \N$, $i \in \{1,2,\dots,\ell_k\}$ that
\begin{equation}\label{eq:uniboundN}
\sup_{\theta \in K} \sup_{\approximate \in \N} \sup_{x \in V}\big[|\cN_{\approximate,i}^{k,\theta}(x)|\indicator{\{\infty\}}{(|\mathrm{supp}(\mu)|)}\big] <\infty,
\end{equation}
\item\label{it2:uniformbounds} it holds for all  $k \in \N$, $i \in \{1,2,\dots,\ell_k\}$ that
\begin{equation}
\sup\limits_{\theta \in K} \sup\limits_{\approximate\in \N} \sup_{x \in V}\big[ \big(|\fg_\approximate (\cN_{\approximate,i}^{k,\theta}(x))| + |(\fg_\approximate)' (\cN_{\approximate,i}^{k,\theta}(x))|\big)\indicator{\{\infty\}}{(|\mathrm{supp}(\mu)|)}\big] < \infty,
\end{equation}
and
\item\label{it3:uniformbounds} it holds for all  $k \in \N$, $i \in \{1,2,\dots,\ell_k\}$, $j \in \{1,2,\dots,\fd\}$ that
\begin{equation}\label{eq:uniboundDerN}
\sup_{\theta \in K} \sup_{\approximate \in \N} \sup_{x \in V}\Big[ \Big| \big(\tfrac{\partial}{\partial \theta_j} \cN_{\approximate,i}^{k,\theta}\big)(x)\Big|\indicator{\{\infty\}}{(|\mathrm{supp}(\mu)|)}\Big] <\infty.
\end{equation}
\end{enumerate}
\end{athm}

\begin{aproof}
Throughout this proof assume without loss of generality that $|\mathrm{supp}(\mu)|=\infty$ (otherwise \cref{it1:uniformbounds}, \cref{it2:uniformbounds}, and \cref{it3:uniformbounds} are obvious).
\argument{the assumption that $K$ is a compact;}{that there exists  $\const \in (0,\infty)$ which satisfies
\begin{equation}\label{eq:constforcompact}
\sup\limits_{\theta \in K}\textstyle\biggl[ \sum\limits_{i=1}^\fd|\theta_i|\biggr] < \const.
\end{equation}}
\startnewargseq
We prove \cref{it1:uniformbounds} by induction on $k \in \N$. For the base case $k=1$ note that \cref{it1:derivsofcN} in \cref{lem:derivsofcN} implies for all $\approximate \in \N$ that $\big(\R^\fd \ni \theta \mapsto \cN_{\approximate,i}^{1,\theta}(x)\big)$ and $\big(\R^{\ell_0} \ni x \mapsto \cN_{\approximate,i}^{1,\theta}(x)\big)$ are continuous. This and the fact for all $x\in \R^{\ell_0}$ it holds that  $\big( \N \ni \approximate \mapsto \cN_{\approximate,i}^{1,\theta}(x) \in \R^{\ell_1}\big)$ is a constant function establish that
\begin{equation}\llabel{base}
    \sup_{\theta \in K} \sup_{\approximate \in \N} \sup_{x \in V}\big[|\cN_{\approximate,i}^{1,\theta}(x)|\indicator{\{\infty\}}{(|\mathrm{supp}(\mu)|)}\big] <\infty.
\end{equation}
This establishes
\cref{it1:uniformbounds} in the base case $k=1$. For the induction step we assume that there exists $k\in \N$, $M\in \N$ which satisfies
\begin{equation}\label{eq:unobounds_IH}
\sup_{\theta \in K} \sup_{\approximate \in \N}\sup_{x \in V} |\cN_{\approximate,i}^{k,\theta}(x)| \leq M <\infty.
\end{equation}
\argument{\cref{eq:def_NN_realization};\cref{eq:constforcompact}; \cref{eq:unobounds_IH}}{that for all $i \in\{1,2,\dots,\ell_{k+1}\}$, $\approximate \in \N$ it holds that
\begin{equation}\llabel{eq1}
\begin{split}
\big|\cN_{\approximate,i}^{k+1,\theta}(x) \big| &\leq \big| \fb_i^{k+1,\theta} \big| + \sum\limits_{j=1}^{\ell_k} \big|\fw_{i,j}^{k+1,\theta}\big| \big|\fg_\approximate\big(\cN_{\approximate,j}^{k,\theta}(x)\big) \big|\\
&\leq \const \Big( 1+ \ell_k \sup\limits_{y \in [-M,M]} |\fg_\approximate(y)|\Big).
\end{split}
\end{equation}}
\argument{\lref{eq1}; the assumption that for all $m\in \N$ it holds that $\sup_{\approximate \in \N}\sup_{x\in[-m,m]}|\fg_\approximate(x)|<\infty$}{that
\begin{equation}\llabel{eq2}
    \sup_{\theta \in K} \sup_{\approximate \in \N} \sup_{x \in V}\big[|\cN_{\approximate,i}^{k,\theta}(x)|\indicator{\{\infty\}}{(|\mathrm{supp}(\mu)|)}\big] <\infty\dott
\end{equation}}
\argument{\lref{eq2};\lref{base};induction}
{\cref{it1:uniformbounds}\dott}
\startnewargseq
\argument{\cref{it1:uniformbounds}; the fact that for all $m\in \N$ it holds that $\sup_{\approximate \in \N}\sup_{x\in[-m,m]}\big(|\fg_\approximate(x)|+|(\fg_\approximate)'(x)|\big)<\infty$ }{\cref{it2:uniformbounds}\dott} 
\startnewargseq
\argument{\cref{lem:derivsofcN}; \cref{it2:uniformbounds}}{ \cref{it3:uniformbounds}\dott}
\end{aproof}

\begin{athm}{lemma}{lem:rep_formulaLrgradient}
Assume \cref{setting:activationapprox} and let $k \in \{1,2,\dots,L\}$, $i \in\{1,2,\dots,\ell_k\}$, $j \in \{1,2,\dots,\ell_{k-1}\}$,  $\approximate \in \N$, $\theta=(\theta_1,\dots,\theta_\fd) \in \R^\fd$. Then
\begin{enumerate}[label=(\roman*)]
\item\label{it1:formulaLrgradient} it holds that $\cL_\approximate \in C^1(\R^\fd,\R)$,
\item\label{it2:formulaLrgradient} it holds that
\begin{equation}
\begin{split}
 &\frac{\partial}{\partial \theta_{(i-1) \ell_{k-1}+j + \mathbf{d}_{k-1}}} \cL_\approximate(\theta)\\ &= \int_{\R^{\ell_0}\times \R^{\ell_L}} (\partial_{\ell_L+(i-1) \ell_{k-1}+j + \mathbf{d}_{k-1}} H) \big(\cN_\approximate^{L,\theta}(x),\theta,y\big)  + \sum\limits_{h=1}^{\ell_L} (\partial_h H) \big(\cN_\approximate^{L,\theta}(x),\theta, y\big)  \\
   &\sum_{
      \substack{v_k,v_{k+1}, \ldots,v_L \in \N, \\ \forall w\in \N\cap[k, L ]\colon v_w\leq\ell_w} 
    }
    \Bigl[
\fg_\approximate(\cN_{\approximate,j}^{\max\{k-1,1\},\theta}(x))\indicator{ (1, L ]}(k)+    
      x_j\indicator{\{1\} }(k)
    \Bigr]
\\
    &\cdot
    \Bigl[ \indicator{ \{ i \} }( v_k ) \Bigr]
    \Bigl[\indicator{ \{ h \} }( v_L ) \Bigr]
    \Bigl[
      \textstyle{\prod}_{p=k+1}^K
      \big( \fw^{p, \theta }_{v_p,v_{p-1} } 
      \big[(\fg_\approximate)'(\cN_{\approximate,v_{p-1}}^{p-1,\theta}(x))\big]
      \big)
    \Bigr]\mu(\d x , \d y),
\end{split}
\end{equation}
and 
\item\label{it3:formulaLrgradient} it holds that
\begin{equation}
\begin{split}
&\frac{\partial}{\partial \theta_{\ell_k \ell_{k-1}+i + \mathbf{d}_{k-1}}} \cL_\approximate (\theta) \\
&= \int_{\R^{\ell_0}\times \R^{\ell_L}} (\partial_{\ell_L+\ell_k \ell_{k-1}+i + \mathbf{d}_{k-1}} H) \big(\cN_\approximate^{L,\theta}(x),\theta ,y\big)+ \sum\limits_{h=1}^{\ell_L}  (\partial_h H) (\cN_\approximate^{L,\theta}(x),\theta ,y\big)  \\
&\sum_{\substack{v_k,v_{k+1}, \ldots,v_K \in \N, \\ \forall w\in \N\cap[k, K ]\colon v_w\leq\ell_w} }
    \Big[\indicator{\{ i \} }(v_k)\Big]
    \Big[\indicator{\{ h \} }(v_K)\Big]\\
    &\qquad\Big[\textstyle{\prod}_{p=k+1}^K
    \big( \fw^{p, \theta }_{v_p,v_{p-1} } 
    (\fg_\approximate)'( \cN_{\approximate,v_{p-1}}^{p-1,\theta}(x))
    \big)\Big]\mu(\d x , \d y).
\end{split}
\end{equation}
\end{enumerate}
\end{athm}
\begin{aproof}
Throughout this proof let $\bfe_1,\bfe_2,\dots,\bfe_\fd\in \R^\fd$ satisfy for all $i\in \{1,2,\dots,\fd\}$, $x=(x_1,\dots,x_\fd)\in \R^\fd$ that
\begin{equation}\llabel{def: e}
    \spro{\bfe_i,x}=x_i.
\end{equation}
\argument{\cref{it1:derivsofcN} in \cref{lem:derivsofcN};  the assumption that $H \in C^1(\R^{\ell_L}\times \R^\fd \times \R^{\ell_L},\R)$} {\cref{it1:formulaLrgradient}\dott}
\argument{chain rule;}{for all $l\in \{(i-1)\ell_{k-1}+j+\bfd_{k-1},\ell_k\ell_{k-1}+i+\bfd_{k-1}\}$ that
\begin{equation}\llabel{eq1}
\begin{split}
&\frac{\partial}{\partial \theta_l}\big[H(\cN_\approximate^{L,\theta}(x),\theta,y)\big]\\ &= \textstyle(\partial_{\ell_L+l} H)\big(\cN_\approximate^{L,\theta}(x),\theta,y\big)+ \sum\limits_{h=1}^{\ell_L} (\partial_h H) \big(\cN_\approximate^{L,\theta}(x),\theta,y\big) (\frac{\partial}{\partial \theta_l} \cN_{\approximate,h}^{L,\theta})(x).
\end{split}
\end{equation}}
\argument{\lref{eq1}; \cref{it1:derivsofcN} in \cref{lem:derivsofcN}; the assumption that $H \in C^1(\R^{\ell_L}\times\R^{\fd}\times \R^{\ell_L},\R)$}{that for all bounded $A,B\subseteq \R^{\ell_L}$, $C\subseteq \R^\fd$ and all $l\in \{(i-1)\ell_{k-1}+j+\bfd_{k-1},\ell_k\ell_{k-1}+i+\bfd_{k-1}\}$ it holds that
\begin{equation}\llabel{eq2}
\begin{split}
 &\sup_{(x,y,\vartheta) \in A\times B\times C}\Big|\textstyle(\frac{\partial}{\partial \vartheta_l}H)(\cN_\approximate^{L,\vartheta}(x),\vartheta,y)\Big| \\
&\leq \sup_{(x,y,\vartheta) \in A\times B\times C} \Big| \partial_{\ell_L+l} H\big(\cN_\approximate^{L,\vartheta}(x),\vartheta,y\big)\Big| \\
&+ \textstyle\sum\limits_{h=1}^{\ell_k}\textstyle\biggl[\biggl[\sup\limits_{(x,y,\vartheta) \in A\times B\times C} \bigl| \partial_h H(\cN_\approximate^{L,\vartheta}(x),\vartheta,y)\bigl|\biggr]\biggl[\sup\limits_{(x,y,\vartheta) \in A\times B\times C}\bigl|(\frac{\partial}{\partial \vartheta_l} \cN_{\approximate,h}^{L,\vartheta})(x)\bigr|\biggr]\biggr] < \infty.
\end{split}
\end{equation}}
\argument{the assumption that $\{x\in \R^{\ell_0}\colon (\exists\, y\in \R^{\ell_L}\colon (x,y)\in \mathrm{supp}(\mu))\}$ is bounded}{that there exist $a,b\in \R$ which satisfy
\begin{equation}\llabel{eq2.1}
    \mathrm{supp}(\mu)\subseteq [a,b]^{\ell_0}\times \R^{\ell_L}\dott
\end{equation}}
\startnewargseq
\argument{\lref{eq1};\lref{eq2.1};the assumption that $H \in C^1(\R^{\ell_L}\times\R^{\fd}\times \R^{\ell_L},\R)$}{that for all $\varepsilon\in \R\backslash \{0\}$, $l\in \{(i-1)\ell_{k-1}+j+\bfd_{k-1},\ell_k\ell_{k-1}+i+\bfd_{k-1}\}$ it holds that
\begin{equation}\llabel{eq2.2}
    \begin{split}
&\int_{\R^{\ell_0}\times\R^{\ell_L}}\varepsilon^{-1}|H(\mN^{L,\theta}_n(x),\theta,y)-H(\mN^{L,\theta+\varepsilon\bfe_l}(x),\theta+\varepsilon\bfe_{l},y)|\,\mu(\d x,\d y)\\
&\leq \int_{[a,b]^{\ell_0}\times\R^{\ell_L}}\biggl[\sup_{\delta\in [0,\varepsilon]} \bigl| \partial_{\ell_L+l} H\big(\cN_\approximate^{L,\theta+\delta \bfe_l}(x),\theta+\delta\bfe_l,y\big)\bigr|\biggr]\\
&+\sum\limits_{h=1}^{\ell_k}\textstyle\biggl[\biggl[\sup\limits_{\delta\in [0,\varepsilon]} \bigl| \partial_h H(\cN_\approximate^{L,\theta+\delta\bfe_l}(x),\theta+\delta\bfe_l,y)\bigl|\biggr]\biggl[\sup\limits_{\delta\in [0,\varepsilon]}\bigl|(\frac{\partial}{\partial \theta_l} \cN_{\approximate,h}^{L,\theta+\delta\bfe_l})(x)\bigr|\biggr]\biggr] \,\mu(\d x,\d y)\\
&= \int_{[a,b]^{\ell_0}\times\R^{\ell_L}}\biggl(\biggl[\sup_{\delta\in [0,\varepsilon]} \bigl| \partial_{\ell_L+l} H\big(\cN_\approximate^{L,\theta+\delta \bfe_l}(x),\theta+\delta\bfe_l,y\big)\bigr|\biggr]\\
&+\sum\limits_{h=1}^{\ell_k}\textstyle\biggl[\biggl[\sup\limits_{\delta\in [0,\varepsilon]} \bigl| \partial_h H(\cN_\approximate^{L,\theta+\delta\bfe_l}(x),\theta+\delta\bfe_l,y)\bigl|\biggr]\biggl[\sup\limits_{\delta\in [0,\varepsilon]}\bigl|(\frac{\partial}{\partial \theta_l} \cN_{\approximate,h}^{L,\theta+\delta\bfe_l})(x)\bigr|\biggr]\biggr]\biggr)\\
&\cdot\mathbbm 1_{\mathrm{supp}(\mu)}(x,y)\,\mu(\d x,\d y).
    \end{split}
\end{equation}}
\argument{\cref{eq:def_risk_function};the assumption that $\mu$ is a finite measure}{that for all $l\in \{(i-1)\ell_{k-1}+j+\bfd_{k-1},\ell_k\ell_{k-1}+i+\bfd_{k-1}\}$ it holds that
\begin{equation}\llabel{eq2.4}
\int_{\R^{\ell_0}\times\R^{\ell_L}}\bigl(|H(\mN^{L,\theta}_n(x),\theta,y)|+1\bigr)\,\mu(\d x,\d y)<\infty.
\end{equation}}
\argument{\lref{eq2.4};\cref{setting: assume};\lref{eq2};\lref{eq2.2};\cref{it1:uniformbounds} in \cref{lem:uniformboundsforNandderN}}{for all $l\in \{(i-1)\ell_{k-1}+j+\bfd_{k-1},\ell_k\ell_{k-1}+i+\bfd_{k-1}\}$ that
\begin{equation}\llabel{eq2.3}
    \begin{split}
\sup_{\varepsilon\in (-1,1)\backslash\{0\}}\int_{\R^{\ell_0}\times\R^{\ell_L}}\varepsilon^{-1}|H(\mN^{L,\theta}_n(x),\theta,y)-H(\mN^{L,\theta+\varepsilon\bfe_l}(x),\theta+\varepsilon\bfe_{l},y)|\,\mu(\d x,\d y)<\infty.
    \end{split}
\end{equation}}
\argument{\lref{eq2.3};the dominated convergence theorem }{for all  $l\in \{(i-1)\ell_{k-1}+j+\bfd_{k-1},\ell_k\ell_{k-1}+i+\bfd_{k-1}\}$ that
\begin{equation}\llabel{eq3}
\begin{split}
 \frac{\partial}{\partial \theta_l}\cL_\approximate(\theta) &= \int_{\R^{\ell_0}\times \R^{\ell_L}}(\partial_{\ell_L+l} H)\big(\cN_\approximate^{L,\theta}(x),\theta,y\big) \\
 &\quad+\sum\limits_{h=1}^{\ell_L} (\partial_h H) (\cN_\approximate^{L,\theta}(x),\theta,y\big) \big(\textstyle\frac{\partial}{\partial \theta_l} \cN_{\approximate,h}^{L,\theta}\big)(x)\mu(\d x , \d y).
\end{split}
\end{equation}}
\argument{\lref{eq3}; \cref{it2:derivsofcN} in \cref{lem:derivsofcN}}{\cref{it2:formulaLrgradient}\dott} 
\startnewargseq
\argument{\lref{eq3}; \cref{it3:derivsofcN} in \cref{lem:derivsofcN}}{\cref{it3:formulaLrgradient}\dott}
\end{aproof}

\begin{athm}{prop}{thm:generalizedgrad}
Assume \cref{setting:activationapprox} and let $k \in \{1,2,\dots,L\}$, $i \in \{1,2,\dots,\ell_k\}$, $j \in \{1,2,\dots,\ell_{k-1}\}$. Then
\begin{enumerate}[label=(\roman*)]
\item\label{it1:generalizedgrad} it holds for all $\theta\in \R^d$ that
  \begin{equation}
  \limsup\limits_{\approximate \to \infty} \Big( \big| \cL_\approximate(\theta) - \cL_0(\theta) \big| + \norm{ (\nabla \cL_\approximate) (\theta) - \cG(\theta)} \Big) =0,
  \end{equation}
\item\label{it2:generalizedgrad}  it holds for all $\theta\in \R^d$ that
  \begin{equation}\label{eq:formulaG}
  \begin{split}
  &\cG_{(i-1) \ell_{k-1}+j + \mathbf{d}_{k-1}} (\theta) \\
  &= \int_{\R^{\ell_0}\times \R^{\ell_L}} (\partial_{\ell_L + (i-1) \ell_{k-1}+j + \mathbf{d}_{k-1}}H)\big(\cN_{0}^{L,\theta}(x),\theta,y\big)+\sum\limits_{h=1}^{\ell_L} (\partial_h H) \big(\cN_{0}^{L,\theta}(x),\theta,y\big)  \\
    &\sum_{
      \substack{v_k,v_{k+1}, \ldots,v_L \in \N, \\ \forall w\in \N\cap[k, L ]\colon v_w\leq\ell_w} 
    }
    \Bigl[
      \fg_{0} (\cN_{0,j}^{\max\{k-1,1\},\theta}(x))\indicator{ (1, L ]}(k)+    
      x_j\indicator{\{1\} }(k)
    \Bigr]
\\
    &\quad\cdot
    \Bigl[ \indicator{ \{ i \} }( v_k ) \Bigr]
    \Bigl[\indicator{ \{ h \} }( v_L ) \Bigr]
    \Bigl[
      \textstyle{\prod}_{p=k+1}^L
      \big( \fw^{p, \theta }_{v_p,v_{p-1} } 
      \big[( \mathscr{d}_\func \fg_0)(\cN_{0,v_{p-1}}^{p-1,\theta}(x))\big]
      \big)
    \Bigr]  \mu(\d x, \d y),
    \end{split}
  \end{equation}
  and
\item\label{it3:generalizedgrad} it holds for all $\theta\in \R^d$ that
\begin{equation}
 \begin{split}
  &\cG_{\ell_k \ell_{k-1}+i + \mathbf{d}_{k-1}} (\theta) \\
  &= \int_{\R^{\ell_0}\times \R^{\ell_L}}  (\partial_{\ell_L+\ell_k \ell_{k-1}+i + \mathbf{d}_{k-1}} H)\big(\cN_{0}^{L,\theta}(x),\theta,y\big)+\sum\limits_{h=1}^{\ell_L}(\partial_h H) \big(\cN_{0}^{L,\theta}(x),\theta,y\big)  \\
  &\sum_{\substack{v_k,v_{k+1}, \ldots,v_L \in \N, \\ \forall w\in \N\cap[k, L ]\colon v_w\leq\ell_w} }
    \Big[\indicator{\{ i \} }(v_k)\Big]
    \Big[\indicator{\{ h \} }(v_L)\Big]\\ &\qquad\qquad\quad
    \Big[\textstyle{\prod}_{p=k+1}^L
    \big( \fw^{p, \theta }_{v_p,v_{p-1} } 
    \big[(\mathscr{d}_\func \fg_0)( \cN_{0,v_{p-1}}^{p-1,\theta}(x))\big]
    \big)\Big]\mu(\d x , \d y)
    \end{split}
\end{equation}
%In particular, the mapping $\big(\R^\fd \ni \theta \mapsto \cG(\theta) \in \R^\fd \big)$  is continuous.
\end{enumerate}
\cfadd{def:LKB_left_right_derivative}\cfout.
\end{athm}

\begin{aproof}
\argument{\cref{it1:derivsofcN} in \cref{lem:derivsofcN}; the assumption that $H \in C^1(\R^{\ell_L}\times\R^{\fd}\times \R^{\ell_L},\R)$}{for all bounded $A,B\subseteq \R^{\ell_L}$, $C\subseteq \R^\fd$ and all $l\in \{(i-1)\ell_{k-1}+j+\bfd_{k-1},\ell_k\ell_{k-1}+i+\bfd_{k-1}\}$ that
\begin{equation}\llabel{eq2}
\begin{split}
 &\sup_{m\in \N}\sup_{(x,y,\vartheta) \in A\times B\times C}\bigl|(\partial_{\ell_L+l} H) \big(\cN_m^{L,\theta}(x),\vartheta,y\big)\bigr| + \textstyle\sum\limits_{h=1}^{\ell_L} \bigl|(\partial_h H) \big(\cN_m^{L,\vartheta}(x),\vartheta, y\big)\bigr|< \infty.
\end{split}
\end{equation}}
\argument{the assumption that $\{x\in \R^{\ell_0}\colon (\exists\, y\in \R^{\ell_L}\colon (x,y)\in \mathrm{supp}(\mu))\}$ is bounded}{that there exist $a,b\in \R$ which satisfies
\begin{equation}\llabel{eqq1}
    \mathrm{supp}(\mu)\subseteq [a,b]^{\ell_0}\times \R^{\ell_L}\dott
\end{equation}}
\argument{\lref{eqq1};}{for all $n\in \N$, $l\in \{(i-1)\ell_{k-1}+j+\bfd_{k-1,\ell_k\ell_{k-1}+i+\bfd_{k-1}}\}$, $\theta\in \R^\fd$ that
\begin{equation}\llabel{eqq2}
\begin{split}
    &\int_{\R^{\ell_0}\times \R^{\ell_L}} \bigl|(\partial_{\ell_L+l} H) \big(\cN_\approximate^{L,\theta}(x),\theta,y\big)\bigr| + \textstyle\sum\limits_{h=1}^{\ell_L} \bigl|(\partial_h H) \big(\cN_\approximate^{L,\theta}(x),\theta, y\big)\bigr|\,\mu(\d x,\d y)\\
    &=\int_{[a,b]^{\ell_0}\times \R^{\ell_L}}\bigl|(\partial_{\ell_L+l} H) \big(\cN_\approximate^{L,\theta}(x),\theta,y\big)\bigr| + \textstyle\sum\limits_{h=1}^{\ell_L} \bigl|(\partial_h H) \big(\cN_\approximate^{L,\theta}(x),\theta, y\big)\bigr|\,\mu(\d x,\d y)\\
    &\leq \int_{[a,b]^{\ell_0}\times \R^{\ell_L}}\sup_{m\in \N}\biggl[\bigl|(\partial_{\ell_L+l} H) \big(\cN_m^{L,\theta}(x),\theta,y\big)\bigr| + \textstyle\sum\limits_{h=1}^{\ell_L} \bigl|(\partial_h H) \big(\cN_m^{L,\theta}(x),\theta, y\big)\bigr|\biggr]\,\mu(\d x,\d y)\\
    &=\int_{[a,b]^{\ell_0}\times \R^{\ell_L}}\biggl[\sup_{m\in \N}\biggl[\bigl|(\partial_{\ell_L+l} H) \big(\cN_m^{L,\theta}(x),\theta,y\big)\bigr| + \textstyle\sum\limits_{h=1}^{\ell_L} \bigl|(\partial_h H) \big(\cN_m^{L,\theta}(x),\theta, y\big)\bigr|\biggr]\biggr]\\
    &\cdot \mathbbm 1_{\mathrm{supp}(\mu)}(x,y)\,\mu(\d x,\d y)
    .
    \end{split}
\end{equation}}
\argument{\cref{eq:def_risk_function}; the assumption that $\mu$ is a finite measure}{that for all $\theta\in \R^\fd$ it holds that
\begin{equation}\llabel{eqq3}
    \int_{\R^{\ell_0} \times \R^{\ell_L}}
 \bigl(\bigl|H\big(0,\theta,y\big)\bigr|+1\bigr) \, \mu(\mathrm{d}x,\d y)<\infty.
\end{equation}}
\argument{\lref{eqq3};\cref{setting: assume};\lref{eq2};\cref{it1:uniformbounds} in \cref{lem:uniformboundsforNandderN}}{that
\begin{equation}\llabel{eqq4}
    \int_{[a,b]^{\ell_0}\times \R^{\ell_L}}\sup_{m\in \N}\biggl[\bigl|(\partial_{\ell_L+l} H) \big(\cN_m^{L,\theta}(x),\theta,y\big)\bigr| + \textstyle\sum\limits_{h=1}^{\ell_L} \bigl|(\partial_h H) \big(\cN_m^{L,\theta}(x),\theta, y\big)\bigr|\biggr]\,\mu(\d x,\d y)<\infty.
\end{equation}}
 \argument{\lref{eqq2};\lref{eqq4};\cref{it2:gcNconvergence} in \cref{prop:gcNconvergence};\cref{lem:uniformboundsforNandderN};\cref{lem:rep_formulaLrgradient};the dominated convergence theorem}{that for all $\theta \in \R^{\fd}$ it holds that
\begin{equation}\label{eq:Lr_convergence}
\limsup\limits_{\approximate\to \infty} |\cL_\approximate(\theta) - \cL_0(\theta)| =0.
\end{equation}}
\argument{the assumption that $H \in C^1(\R^{\ell_L}\times\R^{\fd}\times \R^{\ell_L},\R)$}{for all $\theta=(\theta_1,\dots,\theta_\fd) \in \{    \vartheta \in \R^{ \fd } \colon 
    ( ( \nabla\cL_\approximate )( \vartheta ))_{ \approximate \in \N }\allowbreak
    \text{ is convergent} 
  \}$, $l\in \{(i-1)\ell_{k-1}+j+\bfd_{k-1,\ell_k\ell_{k-1}+i+\bfd_{k-1}}\}$ that 
\begin{equation}\label{eq:G_afterchainrule}
\begin{split}
\cG_l (\theta) &= \lim\limits_{\approximate\to \infty} \frac{\partial}{\partial \theta_l}\cL_\approximate(\theta)\\
&= \lim\limits_{\approximate \to \infty} \Big[ \int_{\R^{\ell_0}\times \R^{\ell_L}}(\partial_{\ell_L+l} H) \big(\cN_\approximate^{L,\theta}(x),\theta,y\big)\\
&+\sum\limits_{h=1}^{\ell_L}(\partial_h H) \big(\cN_\approximate^{L,\theta}(x),\theta,y\big) \Big(\frac{\partial}{\partial \theta_l} \cN_{\approximate,h}^{L,\theta}\Big)(x) \mu(\d x , \d y)\Big]. 
\end{split}
\end{equation}}
\argument{\cref{it2:gcNconvergence} in \cref{prop:gcNconvergence}}{ for all $x =(x_1,\ldots,x_{\ell_0})\in \R^{\ell_0}$, $\theta=(\theta_1,\dots,\theta_\fd) \in \R^\fd$ that
\begin{equation}\label{eq:G_weightslimit}
\begin{split}
\lim\limits_{\approximate\to\infty} &\Big(\sum_{
      \substack{v_k,v_{k+1}, \ldots,v_L \in \N, \\ \forall w\in \N\cap[k, L ]\colon v_w\leq\ell_w} 
    }
    \Bigl[
\fg_\approximate(\cN_{\approximate,j}^{\max\{k-1,1\},\theta}(x))\indicator{ (1, L ]}(k)+    
      x_j\indicator{\{1\} }(k)
    \Bigr]
\\
    &\quad \cdot
    \Bigl[ \indicator{ \{ i \} }( v_k ) \Bigr]
    \Bigl[\indicator{ \{ h \} }( v_L ) \Bigr]
    \Bigl[
      \textstyle{\prod}_{p=k+1}^K
      \big( \fw^{p, \theta }_{v_p,v_{p-1} } 
      \big[(\fg_\approximate)'(\cN_{\approximate,v_{p-1}}^{p-1,\theta}(x))\big]
      \big)
    \Bigr] \Big)
    \\
    &= \sum_{
      \substack{v_k,v_{k+1}, \ldots,v_L \in \N, \\ \forall w\in \N\cap[k, L ]\colon v_w\leq\ell_w} 
    }
    \Bigl[
      \fg_0 (\cN_{0,j}^{\max\{k-1,1\},\theta}(x))\indicator{ (1, L ]}(k)+    
      x_j\indicator{\{1\} }(k)
    \Bigr]
\\
    &\quad \cdot
    \Bigl[ \indicator{ \{ i \} }( v_k ) \Bigr]
    \Bigl[\indicator{ \{ h \} }( v_L ) \Bigr]
    \Bigl[
      \textstyle{\prod}_{p=k+1}^K
      \big( \fw^{p, \theta }_{v_p,v_{p-1} } 
      \big[( \mathscr{d}_\func \fg_0) (\cN_{0,v_{p-1}}^{p-1,\theta}(x))\big]
      \big)
    \Bigr]
    \\
  \end{split}
  \end{equation}
and 
\begin{equation}\label{eq:G_biaslimit}
\begin{split}
&\lim\limits_{\approximate\to\infty}\Big( \sum_{\substack{v_k,v_{k+1}, \ldots,v_K \in \N, \\ \forall w\in \N\cap[k, K ]\colon v_w\leq\ell_w} }
    \Big[\indicator{\{ i \} }(v_k)\Big]
    \Big[\indicator{\{ h \} }(v_K)\Big]
    \Big[\textstyle{\prod}_{p=k+1}^K
    \big( \fw^{p, \theta }_{v_p,v_{p-1} } 
    \big[(\fg_\approximate)'(\cN_{\approximate,v_{p-1}}^{p-1,\theta}(x))\big]
    \big)\Big]\Big) \\
    &= \sum_{\substack{v_k,v_{k+1}, \ldots,v_K \in \N, \\ \forall w\in \N\cap[k, K ]\colon v_w\leq\ell_w} }
    \Big[\indicator{\{ i \} }(v_k)\Big]
    \Big[\indicator{\{ h \} }(v_K)\Big]
    \Big[\textstyle{\prod}_{p=k+1}^K
    \big( \fw^{p, \theta }_{v_p,v_{p-1} } 
    \big[( \mathscr{d}_\func \fg_0)(\cN_{0,v_{p-1}}^{p-1,\theta}(x))\big]
    \big)\Big].
  \end{split}  
\end{equation}}
\argument{\lref{eqq2};\lref{eqq4}; \cref{eq:G_afterchainrule};\cref{eq:G_weightslimit}; \cref{eq:G_biaslimit}; \cref{lem:uniformboundsforNandderN};\cref{lem:rep_formulaLrgradient}; the assumption that $\mu$ is a finite measure; the dominated convergence theorem}{that 
\begin{enumerate}[label=(\Roman*)]
    \item it holds that $\{    \vartheta \in \R^{ \fd } \colon 
    ( ( \nabla\cL_\approximate )( \vartheta ) )_{ \approximate \in \N }
    \text{ is convergent} \allowbreak
  \} = \R^\fd$ and
  \item it holds for all $\theta \in \R^\fd$ that
\begin{equation}\llabel{eq1}
  \limsup\limits_{\approximate \to \infty} \norm{ \nabla \cL_\approximate(\theta) - \cG(\theta)} =0.
\end{equation}
\end{enumerate}}
\argument{\lref{eq1};\cref{eq:Lr_convergence}}{ \cref{it1:generalizedgrad}\dott} 
\argument{\lref{eqq2};\lref{eqq4}; the dominated convergence theorem; \cref{lem:rep_formulaLrgradient};\cref{eq:G_afterchainrule};\cref{eq:G_weightslimit}}{ \cref{it2:generalizedgrad}\dott} 
\argument{\lref{eqq2};\lref{eqq4};\cref{lem:rep_formulaLrgradient};the dominated convergence theorem; \cref{eq:G_afterchainrule};\cref{eq:G_biaslimit}}{ \cref{it3:generalizedgrad}\dott}
%The continuity of the mapping $\big( \R^\fd \ni \theta \mapsto \cG(\theta) \in \R^\fd \big)$ follows from the fact that each coordinate is the composition of continuous functions, see \cref{eq:formulaG}. 
\end{aproof}
\begin{athm}{theorem}{thm:mainthm_sequence: part 1}
%For every $n \in \N$, $f \colon \R^n \to \R$ let $\mathbf{T}_f \colon \R^n \to \R^n$ denote the realization of the Tensorflow gradient of $f$
Let $ \fd, L\in \N $,
$ \ell_0,\ell_1,\dots,\ell_L\in \N $
%$\delta \in (0,\infty)$,
satisfy 
$
  \fd = \sum_{k=1}^L \ell_k ( \ell_{k-1} + 1 )
$, 
for every $\activate\in C(\R,\R)$,
$ 
  \theta = ( \theta_1, \dots, \theta_{ \fd } ) \in \R^{ \fd }$
 let $\mN^{ k, \theta }_{\activate}=(\mN^{ k, \theta }_{\activate,1},\dots,\mN^{ k, \theta }_{\activate,\ell_k}) \colon \R^{ \ell_0 }\allowbreak \to \R^{ \ell_k} $, $k\in \{0,1,\dots,L\}$, satisfy for all $k\in \{0,1,\dots,L-1\}$, $x=(x_1,\dots, x_{\ell_0})\in \R^{\ell_0}$, $i\in\{1,2,\dots,\ell_{k+1}\}$ that
\begin{equation}\llabel{def: realization}
\begin{split}
  \mN^{ k+1, \theta }_{\activate,i}( x ) &= \theta_{\ell_{k+1}\ell_{k}+i+\sum_{h=1}^{k}\ell_h(\ell_{h-1}+1)}\\
  &+\textstyle\sum_{j=1}^{\ell_{k}}\theta_{(i-1)\ell_{k}+j+\sum_{h=1}^{k}\ell_h(\ell_{h-1}+1)}\big(x_j\indicator{\{0\}}(k) 
  +\activate(\mN^{k,\theta}_{\activate,j}(x))\indicator{\N}(k)\big),
  \end{split}
\end{equation}
let $\mu \colon \cB(\R^{\ell_0}\times \R^{\ell_L})\to [0,\infty]$ be a finite measure,
let $H \in C^1(\R^{\ell_L} \times \R^{\ell_L}\times\R^{\fd},\R)$, 
for every $\activate\in C(\R,\R)$ let $\cL_\activate \colon \R^\fd \to \R$ satisfy for all $\theta \in \R^\fd$  that $\int_{\R^{\ell_0}\times \R^{\ell_L}}
  |H(\mathcal{N}_{\activate}^{L,\theta}(x),y,\theta)|\allowbreak\, \mu (\mathrm{d}x,\mathrm{d}y)<\infty$ and
\begin{equation}\llabel{def: risk}
\textstyle 
  \cL_\activate( \theta ) 
  = 
  \int_{\R^{\ell_0}\times \R^{\ell_L}}
  H(\mathcal{N}_{\activate}^{L,\theta}(x),y,\theta)\, \mu (\mathrm{d}x,\mathrm{d}y)
  ,
\end{equation}
assume that $\{x\in \R^{\ell_0}\colon (\exists\, y\in \R^{\ell_L}\colon (x,y)\in \mathrm{supp}(\mu))\}$ is bounded, assume for all $r\in (0,\infty)$, $X\in [-r,r]^{\ell_0}$, $\Theta\in [-r,r]^{\fd}$ that
\begin{equation}\llabel{assume}
   \textstyle \sup\limits_{x\in [-r,r]^{\ell_0}}\sup\limits_{y\in \R^{\ell_L}}\sup\limits_{\theta\in [-r,r]^\fd}\Bigl(\frac{(\|\nabla_x H(x,y,\theta)\|+\|\nabla_\theta H(x,y,\theta)\|)\mathbbm 1_{\mathrm{supp}(\mu)}(x,y)}{1+|H(X,y,\Theta)|}\Bigr)<\infty,
\end{equation} 
let $S\subseteq \R$ be finite, let $\scrA\in C(\R,\R)$ satisfy $\restr{\scrA}{\R \backslash S} \in C^1(\R \backslash S,\R)$, let $\scra \colon \R \to \R$ satisfy $\restr{\scra}{\R \backslash S}=\nabla(\restr{\scrA}{\R \backslash S})$, and let $\ZZ\in \N_0$ satisfy for all $m\in\N$ that 
\begin{equation}
   \textstyle \sup_{x\in [-m,m]} \bigl[|\scra(x)(\ZZ+|\indicator{\{\infty\}}{(|\mathrm{supp}(\mu)|)})\bigr]<\infty.
\end{equation}
Then 
\begin{enumerate}[label=(\roman*)]
   \item \label{item 1: Theorem X part 1} it holds for all $A\in C^1(\R,\R)$ that $\cL_A\in C^1(\R^\fd,\R)$ and
   \item \label{item 2: Theorem X part 1} there exists a unique $\cG \colon \R^\fd \to \R^\fd$ which satisfies for all $\theta \in \R^\fd$ and all $(A_n)_{n\in \N} \subseteq C^1(\R,\R)$ with $\forall \, x \in \R \colon \exists \, m \in \N \colon \textstyle\sum_{n=m}^\infty (|\scrA(x)-A_n(x)|+|\scra(x)-(A_n)'(x)|) = 0$
and 
$\forall \,m \in \N \colon \sup_{n \in \N} \sup_{x \in [-m,m]} \bigl[ ( |A_n(x)|  + |(A_n)'(x)| )(\ZZ+\indicator{\{\infty\}}{(|\mathrm{supp}(\mu)|)})\bigr]<\infty$
that  
\begin{equation}
\textstyle\limsup_{n \to \infty} \norm{(\nabla\cL_{A_n})(\theta) - \cG(\theta)}=0.
\end{equation}
\end{enumerate}
\end{athm}
\begin{aproof}
Throughout this proof let $h\colon S\to \R$ satisfy for all $s\in S$ that $h(s)=\scra(s)$.
\argument{\cref{it1:formulaLrgradient} in \cref{lem:rep_formulaLrgradient}}{\cref{item 1: Theorem X part 1}\dott}
\startnewargseq
\argument{the fact that $\restr{\scrA}{\R \backslash S} \in C^1(\R \backslash S,\R)$;the fact that $\restr{\scra}{\R \backslash S}=\nabla(\restr{\scrA}{\R \backslash S})$}{that
\begin{equation}\llabel{eq1}
    \scrd_\func \scrA =\scra
\end{equation}
(cf.\ \cref{def:LKB_left_right_derivative})\dott}
\argument{\lref{eq1};\cref{thm:generalizedgrad};\cref{lem: existence of A_n}}{\cref{item 2: Theorem X part 1}\dott}
\end{aproof}
\section{Regularity analysis for generalized gradients of the objective function}\label{sec: regularity}
In this section we establish in \cref{prop:loss:gradient:subdiff} and \cref{item 2: Theorem X} in \cref{Theorem X} in \cref{subsec: frechet definition} below, respectively, that the unique generalized gradient function $\cG\colon \R^\fd\to\R$ of the objective function in \cref{item 2: Theorem X part 1} in \cref{thm:mainthm_sequence: part 1} above satisfies that for every $\theta \in \R^\fd$ we have that the generalized gradient $\cG( \theta )$ is a limiting Fréchet subgradient of the objective function $\cL_{ \scrA } \colon \R^\fd\to \R$ at $\theta$. Moreover, as a consequence of this fact, we demonstrate in \cref{cor:cG_equal_to_gradient} and \cref{item 3: Theorem X} in \cref{Theorem X}, respectively, that the unique generalized gradient function $\cG\colon \R^\fd\to\R$ of \cref{item 2: Theorem X part 1} in \cref{thm:mainthm_sequence: part 1} coincides on every open set on which the objective function $\cL_{ \scrA } \colon \R^\fd\to\R$ is continuously differentiable with the usual gradient of $\cL_{ \scrA }$.

Our arguments in this section are based on the arguments in \cite{Aradexistenceofglobmin} (cf.\ \cite{Aradexistenceofglobminv1}) in which a special case of \cref{item 2: Theorem X,item 3: Theorem X} in \cref{Theorem X} has been established. In particular, in \cref{subsec: Local Lipschitz continuity} we generalize the findings in \cite[Lemma 3.1]{Aradexistenceofglobmin}, in \cref{subsec: local boundedness} we generalize \cite[Lemma 3.2]{Aradexistenceofglobmin}, in \cref{subsec: weak differentiability} we generalize \cite[Corollary 3.1]{Aradexistenceofglobmin}, in \cref{subsec: strong differentiability} we generalize \cite[Proposition 3.2]{Aradexistenceofglobmin}, in \cref{subsec: local estimates} we generalize \cite[Lemma 3.5]{Aradexistenceofglobmin}, in \cref{subsec: continuity propeties} we generalize \cite[Lemma 3.6 and Lemma 3.7]{Aradexistenceofglobmin}, in \cref{subsec: frechet definition} we generalize \cite[Proposition 3.3]{Aradexistenceofglobmin}, and in \cref{prop:loss:gradient:subdiff} in \cref{subsec: frechet definition}
 we generalize \cite[Proposition 3.3]{Aradexistenceofglobmin}. The work \cite[Proposition 3.3 and Corollary 3.2]{Aradexistenceofglobmin} (cf.\ \cite[Proposition 3.18 and Corollary 3.19]{Aradexistenceofglobminv1}) has established the special case of \cref{item 2: Theorem X,item 3: Theorem X} in which the activation function is the \ReLU\ activation, in the which the loss function coincides with the mean squared error loss, in which a special case of the approximation procedure is used (cf., \eg, \cite[Setting 2.1]{ScarpaMultiLayers}), in which a special case case of the target function is considered, and in which a special case of the distribution of the input-output data pair is considered.
 \subsection{Local Lipschitz continuity properties of the objective function}\label{subsec: Local Lipschitz continuity}
\begin{athm}{lemma}{bound N}
    Assume \cref{setting:activationapprox}, let $a\in \R$, $b\in (a,\infty)$, and assume for all $m\in \N$ that $\sup_{\approximate \in \N}\sup_{x\in[-m,m]}\bigl[
    \allowbreak|\fg_\approximate(x)|+|(\fg_\approximate)'(x)|\bigr]<\infty$. Then for all $C\in (0,\infty)$, $k\in \{1,2,\dots,L\}$ there exists $\fC_{C,k}\in (0,\infty)$ which satisfies for all $\theta=(\theta_1,\dots,\theta_\fd)\in \R^\fd$, $x\in [a,b]^{\ell_0}$ with $\max_{i\in \{1,2,\dots,\fd\}}|\theta_i|\leq C$ that
    \begin{equation}\llabel{conclude}
    \begin{split}
       &\textstyle\max_{i\in \{1,2,\dots,\ell_{k}\}} |\cN_{0,i}^{k,\theta}(x)|
  \leq \fC_{C,k}.
       \end{split}
    \end{equation}
\end{athm}
\begin{aproof}
    We prove \lref{conclude} by induction on $k\in \{1,2,\dots,L\}$. For the base case $k=1$ observe that \cref{eq:def_NN_realization} proves that for all $C\in (0,\infty)$, $i\in \{1,2,\dots,\ell_1\}$, $\theta=(\theta_1,\dots,\theta_\fd)\in \R^\fd$, $x=(x_1,\dots,x_{\ell_0})\in [a,b]^{\ell_0}$ with $\max_{i\in \{1,2,\dots,\fd\}}|\theta_i|\leq C$ it holds that
    \begin{equation}\llabel{eq1}
    \begin{split}
    \textstyle
&|\cN_{0,i}^{1,\theta}(x)|=\biggl|\theta_{\ell_1\ell_0+i}+\textstyle\sum\limits_{j=1}^{\ell_0}\theta_{(i-1)\ell_0+j}x_j\biggr|\leq |\theta_{\ell_1\ell_0+i}|+\biggl(\textstyle\sum\limits_{j=1}^{\ell_0}|\theta_{(i-1)\ell_0+j}x_j|\biggr)\\
&\leq \bigg[\max_{j\in \{1,2,\dots,\fd\}}|\theta_j|\biggr]+\ell_0\bigg[\max_{j\in \{1,2,\dots,\fd\}}|\theta_j|\bigg]\max\{1,|a|,|b|\}\\
&\leq C+\ell_0C\max\{1,|a|,|b|\}.
\end{split}
    \end{equation}
    For the induction step we assume that there exist $k\in \{1,2,\dots,L-1\}$ and $\fC_{C,k}$, $C\in (0,\infty)$, which satisfy for all $C\in(0,\infty)$, $\theta=(\theta_1,\dots,\theta_\fd)\in \R^{\fd}$, $x\in [a,b]^{\ell_0}$ with $\max_{i\in \{1,2,\dots,\fd\}}|\theta_i|\leq C$ that 
    \begin{equation}\llabel{assume induction}
            \begin{split}
       \textstyle\max_{i\in \{1,2,\dots,\ell_{k}\}} |\cN_{0,i}^{k,\theta}(x)|
  \leq \fC_{C,k}.
       \end{split}
    \end{equation}
\argument{\cref{eq:def_NN_realization};}{that for all $i\in \{1,2,\dots,\ell_{k+1}\}$, $\theta=(\theta_1,\dots,\theta_{\fd})$, $x\in [a,b]^{\ell_0}$ it holds that
\begin{equation}\llabel{eq2}
\begin{split}
    |\cN_{0,i}^{k+1,\theta}(x)|&\leq |\theta_{\ell_{k+1}\ell_{k}+i+\bfd_{k}}|+\biggl[\textstyle\sum\limits_{j=1}^{\ell_{k}}\bigl|\theta_{(i-1)\ell_{k}+j+\bfd_{k}}\fg_0(\cN_{0,j}^{k,\theta}(x))\bigr|\biggr]\dott
    \end{split}
\end{equation}}
\argument{\cref{scrA is continuous};the assumption that for all $m\in \N$ it holds that $\sup_{\approximate \in \N}\allowbreak\sup_{x\in[-m,m]}\allowbreak\bigl[
    \allowbreak|\fg_\approximate(x)|+|(\fg_\approximate)'(x)|\bigr]<\infty$}{that \llabel{arg1} $A_0$ is continuous\dott}
    \argument{\lref{arg1};}{that for all $C\in (0,\infty)$, $x\in \R$ with $|x|\leq \fC_{C,k}$ there exists $\rho\in (0,\infty)$ which satisfies
    \begin{equation}\llabel{eq3}
        |\fg_0(x)|\leq \rho\dott
    \end{equation}}
    \startnewargseq
    \argument{\lref{assume induction};\lref{eq2};\lref{eq3}}{that for all $C\in (0,\infty)$, $i\in \{1,2,\dots,\ell_{k+1}\}$, $\theta=(\theta_1,\dots,\theta_{\fd})$, $x\in [a,b]^{\ell_0}$ with $\max_{i\in \{1,2,\dots,\fd\}}|\theta_i|\leq C$ it holds that
    \begin{equation}\llabel{eq5}
        |\cN_{0,i}^{k+1,\theta}(x)|\leq |\theta_{\ell_{k+1}\ell_{k}+i+\bfd_{k}}|+\biggl[\textstyle\sum\limits_{j=1}^{\ell_{k}}\bigl|\rho\theta_{(i-1)\ell_{k}+j+\bfd_{k}}\bigr|\biggr]\leq C+C\rho\ell_k.
    \end{equation}}
\argument{\lref{eq5};\lref{eq1};induction}{\lref{conclude}\dott}
\end{aproof}
\begin{athm}{lemma}{Local Lipschitz N}
    Assume \cref{setting:activationapprox}, let $a\in \R$, $b\in (a,\infty)$, and assume for all $m\in \N$ that $\sup_{\approximate \in \N}\sup_{x\in[-m,m]}\bigl[
    \allowbreak|\fg_\approximate(x)|+|(\fg_\approximate)'(x)|\bigr]<\infty$.
    Then for all $C\in (0,\infty)$, $k\in \{1,2,\dots,L\}$ there exists $\fC_{C,k}\in (0,\infty)$ such that for all $\theta=(\theta_1,\dots,\theta_\fd)$, $\vartheta=(\vartheta_1,\dots,\vartheta_\fd)\in \R^{\fd}$, $x\in[a,b]^{\ell_0}$ with $\max_{i\in \{1,2,\dots,\fd\}}\max\{|\theta_i|,|\vartheta_i|\}\leq C$ it holds that
    \begin{equation}\llabel{conclude}
           \begin{split}
       & \textstyle\max_{i\in\{1,2,\dots,\ell_{k}\}}|\cN_{0,i}^{k,\theta}(x)-\cN_{0,i}^{k,\vartheta}(x)|
  \leq
  \fC_{C,k}\br*{ 
    \max\nolimits_{ j \in \cu{ 1, 2, \ldots, \bfd_{k } } }
    \abs{ \theta_j - \vartheta_j } 
  }
  .
       \end{split}
    \end{equation}
\end{athm}
\begin{aproof}
    We prove \lref{conclude} by induction on $k\in \{1,2,\dots,L-1\}$. For the base case $k=1$ observe that \cref{eq:def_NN_realization} proves for all $i\in \{1,2,\dots,\ell_1\}$, $\theta=(\theta_1,\dots,\theta_\fd)$, $\vartheta=(\vartheta_1,\dots,\vartheta_\fd)\in \R^\fd$, $x=(x_1,\dots,x_{\ell_0})\in [a,b]^{\ell_0}$ that
    \begin{equation}\llabel{eq1}
    \begin{split}
    \textstyle
&|\cN_{0,i}^{1,\theta}(x)-\cN_{0,i}^{1,\vartheta}(x)|=\biggl|\theta_{\ell_1\ell_0+i}+\sum\limits_{j=1}^{\ell_0}\theta_{(i-1)\ell_0+j}x_j-\biggl(\vartheta_{\ell_1\ell_0+i}+\sum\limits_{j=1}^{\ell_0}\vartheta_{(i-1)\ell_0+j}x_j\biggr)\biggr|\\
&\leq |\theta_{\ell_1\ell_0+i}-\vartheta_{\ell_1\ell_0+i}|+\biggl(\textstyle\sum\limits_{j=1}^{\ell_0}|\theta_{(i-1)\ell_0+j}-\vartheta_{(i-1)\ell_0+j}||x_j|\biggr)\\
&\leq \bigg[\max_{j\in \{1,2,\dots,\ell_1(\ell_0+1)\}}|\theta_j-\vartheta_j|\biggr]+\ell_0\bigg[\max_{j\in \{1,2,\dots,\ell_1(\ell_0+1)\}}|\theta_j-\vartheta_j|\bigg]\max\{|a|,|b|\}\\
&\leq \max\{|a|,|b|\}(\ell_0+1)\biggl[\max_{j\in \{1,2,\dots,\ell_1(\ell_0+1)\}}|\theta_j-\vartheta_j|\biggr]
  .
\end{split}
    \end{equation}
    For the induction step we assume that there exist $k\in \{1,2,\dots,L-1\}$ and $\fC_{C,k}$, $C\in (0,\infty)$, which satisfy for all $C\in (0,\infty)$, $\theta=(\theta_1,\dots,\theta_\fd)$, $\vartheta=(\vartheta_1,\dots,\vartheta_\fd)\in \R^{\fd}$, $x\in [a,b]^{\ell_0}$ with $\max_{i\in \{1,2,\dots,\fd\}}\max\{|\theta_i|,|\vartheta_i|\}\leq C$ that 
    \begin{equation}\llabel{assume induction}
            \begin{split}
       & \textstyle\max_{i\in\{1,2,\dots,\ell_{k}\}}|\cN_{0,i}^{k,\theta}(x)-\cN_{0,i}^{k+1,\vartheta}(x)|\leq \fC_{C,k}\br*{ 
    \max\nolimits_{ j \in \cu{ 1, 2, \ldots, \bfd_{k } } }
    \abs{ \theta_j - \vartheta_j } 
  }
  .
       \end{split}
    \end{equation}
\argument{\cref{eq:def_NN_realization};}{that for all $i\in \{1,2,\dots,\ell_{k+1}\}$, $\theta=(\theta_1,\dots,\theta_{\fd})$, $\vartheta=(\vartheta_1,\dots,\vartheta_\fd)\in \R^{\fd}$, $x\in [a,b]^{\ell_0}$ it holds that
\begin{equation}\llabel{eq2}
\begin{split}
    |\cN_{0,i}^{k+1,\theta}(x)-\cN_{0,i}^{k+1,\vartheta}(x)|&\leq |\theta_{\ell_{k+1}\ell_{k}+i+\bfd_{k}}-\vartheta_{\ell_{k+1}\ell_{k}+i+\bfd_{k}}|\\
&+\biggl[\textstyle\sum\limits_{j=1}^{\ell_{k}}\bigl|(\theta_{(i-1)\ell_{k}+j+\bfd_{k}}-\vartheta_{(i-1)\ell_{k}+j+\bfd_{k}})\fg_0(\cN_{0,j}^{k,\theta}(x))\bigr|\biggr]\\&+\biggl[\textstyle\sum\limits_{j=1}^{\ell_{k}}\bigl|\vartheta_{(i-1)\ell_{k}+j+\bfd_{k}}\bigl(\fg_0(\cN_{0,j}^{k,\theta}(x))-\fg_0(\cN_{0,j}^{k,\vartheta}(x))\bigr)\bigr|\biggr]\dott
    \end{split}
\end{equation}}
\argument{\cref{bound N};}{that for all $C\in (0,\infty)$ there exists $\bfC\in (0,\infty)$ such that for all $\theta=(\theta_1,\dots,\theta_\fd)$, $\vartheta=(\vartheta_1,\dots,\vartheta_\fd)\in \R^{\fd}$, $x\in[a,b]^{\ell_0}$ with $\max_{i\in \{1,2,\dots,\fd\}}\max\{|\theta_i|,|\vartheta_i|\}\allowbreak\leq C$ it holds that
\begin{equation}\llabel{eq3}
    \max_{i\in \{1,2,\dots,\ell_{k}\}} |\cN_{0,i}^{k,\theta}(x)|
  \leq \bfC.
\end{equation}}
\argument{\cref{cor: equivalent sequence};the assumption that for all $m\in \N$ it holds that $\sup_{\approximate \in \N}\sup_{x\in[-m,m]}\bigl[
    \allowbreak|\fg_\approximate(x)|+|(\fg_\approximate)'(x)|\bigr]<\infty$}{that \llabel{argth1} $\fg_0$ is locally Lipschitz continuous\dott}
\argument{\lref{argth1};\lref{eq3}}{that for all $C\in (0,\infty)$ there exists $\bfC\in (0,\infty)$ which satisfies for all $\theta=(\theta_1,\dots,\theta_\fd)$, $\vartheta=(\vartheta_1,\dots,\vartheta_\fd)\in \R^{\fd}$, $x\in[a,b]^{\ell_0}$ with $\max_{i\in \{1,2,\dots,\fd\}}\max\{|\theta_i|,|\vartheta_i|\}\leq C$ that
\begin{equation}\llabel{eq4}
    \max_{i\in \{1,2,\dots,\ell_k\}}|\fg_0(\mN_{0,j}^{k,\theta}(x))|\leq \bfC
\end{equation}
and 
\begin{equation}\llabel{eq5}
    \max_{i\in \{1,2,\dots,\ell_k\}}\bigl|\fg_0(\cN_{0,j}^{k,\theta}(x))-\fg_0(\cN_{0,j}^{k,\vartheta}(x))\bigr|\leq \bfC\bigl|\cN_{0,j}^{k,\theta}(x)-\cN_{0,j}^{k,\vartheta}(x)\bigr|.
\end{equation}}
\startnewargseq
\argument{\lref{assume induction};\lref{eq4};\lref{eq5}}{that for all $C\in (0,\infty)$, $i\in \{1,2,\dots,\ell_{k+1}\}$, $\theta=(\theta_1,\dots,\theta_\fd)$, $\vartheta=(\vartheta_1,\dots,\vartheta_\fd)\in \R^{\fd}$, $x\in[a,b]^{\ell_0}$ with $\max_{i\in \{1,2,\dots,\fd\}}\max\{|\theta_i|,|\vartheta_i|\}\leq C$ that
\begin{equation}\llabel{eq6}
    \begin{split}
        |\cN_{0,i}^{k+1,\theta}(x)-\cN_{0,i}^{k+1,\vartheta}(x)|&\leq \max_{j\in \{1,2,\dots,\bfd_{k+1}\}}|\theta_i-\vartheta_i|+\ell_k\bfC \biggl[\max_{j\in \{1,2,\dots,\bfd_{k+1}\}}|\theta_i-\vartheta_i|\biggr]\\&+
C\ell_k\bfC\fC_{C,k}]\biggl[\max_{j\in \{1,2,\dots,\bfd_{k+1}\}}|\theta_i-\vartheta_i| \biggr]\\
&=(1+\ell_k\bfC+C\ell_k\bfC\fC_{C,k})\biggl[\max_{j\in \{1,2,\dots,\bfd_{k+1}\}}|\theta_i-\vartheta_i|\biggr]\dott
    \end{split}
\end{equation}}
\argument{\lref{eq6};\lref{eq1};induction}{\lref{conclude}\dott}
\end{aproof}
\begin{athm}{lemma}{Local Lipschitz}
    Assume \cref{setting:activationapprox}, assume for all $m\in \N$ that $\sup_{\approximate \in \N}\sup_{x\in[-m,m]}\allowbreak\bigl[|\fg_\approximate(x)|+|(\fg_\approximate)'(x)|\bigr]<\infty$, and let $K\subseteq\R^\fd$ be compact. Then 
there exists $ \fC \in \R $ such that 
for all 
$\theta, \vartheta \in K $ it holds that
\begin{equation}
\label{eq:local_Lip_estimate}
  \abs{
    \cL_{ 0 }( \theta ) - \cL_{ 0 }( \vartheta ) 
  } 
  \le 
  \fC \norm{ \theta - \vartheta } .
\end{equation}
\end{athm}
\begin{aproof}
    Throughout this proof let $a\in \R$, $b\in (a,\infty)$ satisfy $\{x\in \R^{\ell_0}\colon (\exists\, y \in \R^{\ell_L}\colon (x,y)\in \mathrm{supp}(\mu))\}\subseteq [a,b]^{\ell_0}$.
    \argument{\cref{bound N};\cref{Local Lipschitz N}; the assumption that $K$ is compact; the assumption that for all $m\in \N$ it holds that $\sup_{\approximate \in \N}\sup_{x\in[-m,m]}\allowbreak\bigl[|\fg_\approximate(x)|+|(\fg_\approximate)'(x)|\bigr]<\infty$}{that there exists $\rho\in \R$ which satisfies for all $\theta,\vartheta\in K$, $x\in [a,b]^{\ell_0}$ with $\theta\neq \vartheta$ that
    \begin{equation}\llabel{def: rho}
       \max\biggl\{\|\cN_{0}^{L,\theta}(x)\|, \frac{\|\cN_{0}^{L,\theta}(x)-\cN_{0}^{L,\vartheta}(x)\|}{\|\theta-\vartheta\|}\biggr\} \leq \rho\dott
    \end{equation}}
    \startnewargseq
    \argument{\lref{def: rho}; the fact that $H\in C^1(\R^{\ell_L}\times\R^{\ell_L}\times\R^\fd,\R)$; the fundamental theorem of calculus}{that for all $x,y,z\in \R^{\ell_L}$, $\theta,\vartheta\in \R^{\fd}$ it holds that
    \begin{equation}\llabel{def: varrho}
    \begin{split}
     & \textstyle \|H(x,\theta,z)-H(y,\vartheta,z)\|\\
     & \textstyle\leq \biggl(\biggl[\sup\limits_{s,t\in [0,1]}\|(\frac{\partial}{\partial x}H)(x+t(y-x),\vartheta+s(\theta-\vartheta),z)\|\biggr]\\
     &+\textstyle\biggl[\sup\limits_{s,t\in [0,1]}\|(\frac{\partial}{\partial \theta}H)(x+t(y-x),\vartheta+s(\theta-\vartheta),z)\|\biggr]\biggr)
     (\|x-y\|+\|\theta-\vartheta\|)\dott
      \end{split}
    \end{equation}}
\argument{\cref{eq:def_NN_realization};\lref{def: rho};\lref{def: varrho}; the assumption that $K$ is compact; the fact that $\mathrm{supp}(\mu)\subseteq[a,b]^{\ell_0}\times \R^{\ell_L}$}{that there exists $\varrho\in (0,\infty)$ such that for all $\theta,\vartheta\in K$ it holds that
\begin{equation}\llabel{eq1}
\begin{split}
    &\|\cL_{0}(\theta)-\cL_0(\vartheta)\|=\biggl\|\int_{\R^{\ell_0}\times\R^{\ell_L}}\bigl(H(\mN_\approximate^{L,\theta}(x)\theta,y)-H(\mN_0^{L,\vartheta}(x),\vartheta,y)\bigr)\,\mu(\d x,\d y)\biggr\| \\
    &\leq \int_{\R^{\ell_0}\times\R^{\ell_L}}\bigl\| H(\mN_0^{L,\theta}(x),y,\theta)-H(\mN_0^{L,\vartheta}(x),y,\vartheta)\bigr\|\,\mu(\d x,\d y)\\
    \\& \leq (\varrho\|\theta-\vartheta\|)\int_{\R^{\ell_0}\times\R^{\ell_L}} \textstyle\biggl[\sup\limits_{\bfx\in [-\rho,\rho]^{\ell_0}}\sup\limits_{s\in [0,1]}\|(\frac{\partial}{\partial \bfx}H)(\bfx,\theta+s(\vartheta-\theta),y)\|\\
    &\quad+\textstyle\sup\limits_{\bfx\in [-\rho,\rho]^{\ell_0}}\sup\limits_{s\in [0,1]}\|(\frac{\partial}{\partial \theta}H)(\bfx,\theta+s(\vartheta-\theta),y)\|\biggr]\,\mu(\d x,\d y).
    \end{split}
\end{equation}}
\argument{\cref{eq:def_risk_function};the assumption that $\mu$ is a finite measure}{that for all $\bfx\in \R^{\ell_0}$, $\theta\in \R^\fd$ it holds that
\begin{equation}\llabel{eqq3}
    \int_{\R^{\ell_0} \times \R^{\ell_L}}\bigl(
 \bigl|H\big(\bfx,\theta,y\big)\bigr|+1\big) \, \mu(\mathrm{d}x,\d y)<\infty.
\end{equation}}
\argument{\lref{eqq3};\cref{setting: assume}; the assumption that $K$ is compact; the assumption that $H\in C^1(\R^{\ell_L}\times\R^\fd\times\R^{\ell_L},\R)$; the fact that $\mathrm{supp}(\mu)\subseteq[a,b]^{\ell_0}\times \R^{\ell_L}$}{that for all $\theta,\vartheta\in K$ it holds that
\begin{equation}\llabel{eq2}
\begin{split}
   & \int_{\R^{\ell_0}\times\R^{\ell_L}} \textstyle\biggl[\sup\limits_{\bfx\in [-\rho,\rho]^{\ell_0}}\sup\limits_{s\in [0,1]}\|(\frac{\partial}{\partial \bfx}H)(\bfx,\theta+s(\vartheta-\theta),y)\|\\
   &\quad+\textstyle\sup\limits_{\bfx\in [-\rho,\rho]^{\ell_0}}\sup\limits_{s\in [0,1]}\|(\frac{\partial}{\partial \theta}H)(\bfx,\theta+s(\vartheta-\theta),y)\|\biggr]\,\mu(\d x,\d y)\\
   &<\infty.
   \end{split}
\end{equation}}
\argument{\lref{eq2};\lref{eq1}}{\cref{eq:local_Lip_estimate}\dott}
\end{aproof}
\newcommand{\Rr}{\mathbb{R}}
\newcommand{\Reals}{\mathbb{R}}
\DeclarePairedDelimiter{\pr}{ (}{)}
\subsection{Local boundedness of the generalized gradients of the objective function}\label{subsec: local boundedness}
\begin{athm}{lemma}{bound Lr}
     Assume \cref{setting:activationapprox},  assume for all $m\in \N$ that $\sup_{\approximate \in \N}\sup_{x\in[-m,m]}\allowbreak\bigl[|\fg_\approximate(x)|+|(\fg_\approximate)'(x)|\bigr]<\infty$, and
  let $K\subseteq\R^{ \fd } $ be compact.
  Then
  \begin{equation}\label{grad_L_bd} 
  \sup\nolimits_{\theta \in K} \sup\nolimits_{\approximate\in \N} \norm{ ( \nabla\cL_\approximate)( \theta)}
  <\infty.
  \end{equation}
\end{athm}
\begin{aproof}
    Throughout this proof let $a\in \R$, $b\in (a,\infty)$ satisfy $\{x\in \R^{\ell_0}\colon (\exists\, y \in \R^{\ell_L}\colon (x,y)\in \mathrm{supp}(\mu))\}\subseteq [a,b]^{\ell_0}$.
    \argument{\cref{lem:rep_formulaLrgradient};}{that
    \begin{enumerate}[label=(\roman*)]
\item\llabel{item 1} it holds for all $\approximate\in \N$, $k \in \{1,2,\dots,L\}$, $i \in \{1,2,\dots,\ell_k\}$, $j \in \{1,2,\dots,\ell_{k-1}\}$, $\theta=(\theta_1,\dots,\theta_\fd) \in \R^{\fd}$ that
\begin{equation}
\begin{split}
 &\frac{\partial}{\partial \theta_{(i-1) \ell_{k-1}+j + \mathbf{d}_{k-1}}} \cL_\approximate(\theta)\\ &= \int_{[a,b]^{\ell_0}\times \R^{\ell_L}} (\partial_{\ell_L+(i-1) \ell_{k-1}+j + \mathbf{d}_{k-1}} H) \big(\cN_{0}^{L,\theta}(x),\theta,y\big)  + \sum\limits_{h=1}^{\ell_L} (\partial_h H) \big(\cN_{0}^{L,\theta}(x),\theta, y\big)  \\
   &\qquad\sum_{
      \substack{v_k,v_{k+1}, \ldots,v_L \in \N, \\ \forall w\in \N\cap[k, L ]\colon v_w\leq\ell_w} 
    }
    \Bigl[
\fg_\approximate(\cN_{\approximate,j}^{\max\{k-1,1\},\theta}(x))\indicator{ (1, L ]}(k)+    
      x_j\indicator{\{1\} }(k)
    \Bigr]
\\
    &\qquad\qquad\cdot
    \Bigl[ \indicator{ \{ i \} }( v_k ) \Bigr]
    \Bigl[\indicator{ \{ h \} }( v_L ) \Bigr]
    \Bigl[
      \textstyle{\prod}_{p=k+1}^K
      \big( \fw^{p, \theta }_{v_p,v_{p-1} } 
      \big[(\fg_\approximate)'(\cN_{\approximate,v_{p-1}}^{p-1,\theta}(x))\big]
      \big)
    \Bigr]\mu(\d x , \d y),
\end{split}
\end{equation}
and 
\item\llabel{item 2} it holds for all $\approximate\in \N$, $k \in \N \cap \{1,2,\dots,L\}$, $i \in\{1,2,\dots,\ell_k\}$,  $\theta=(\theta_1,\dots,\theta_\fd) \in \R^{\fd}$ that
\begin{equation}
\begin{split}
&\frac{\partial}{\partial \theta_{\ell_k \ell_{k-1}+i + \mathbf{d}_{k-1}}} \cL_\approximate (\theta) \\
&= \int_{[a,b]^{\ell_0}\times \R^{\ell_L}} (\partial_{\ell_L+\ell_k \ell_{k-1}+i + \mathbf{d}_{k-1}} H) \big(\cN_{0}^{L,\theta}(x),\theta ,y\big)+ \sum\limits_{h=1}^{\ell_L}  (\partial_h H) (\cN_{0}^{L,\theta}(x),\theta ,y\big)  \\
&\sum_{\substack{v_k,v_{k+1}, \ldots,v_K \in \N, \\ \forall w\in \N\cap[k, K ]\colon v_w\leq\ell_w} }
    \Big[\indicator{\{ i \} }(v_k)\Big]
    \Big[\indicator{\{ h \} }(v_K)\Big]\\
    &\qquad\Big[\textstyle{\prod}_{p=k+1}^K
    \big( \fw^{p, \theta }_{v_p,v_{p-1} } 
    (\fg_\approximate)'( \cN_{\approximate,v_{p-1}}^{p-1,\theta}(x))
    \big)\Big]\mu(\d x , \d y).
\end{split}
\end{equation}
\end{enumerate}}
\argument{\cref{lem:uniformboundsforNandderN};the assumption that $K$ is compact; the assumption that for all $m\in \N$ it holds that $\sup_{\approximate \in \N}\sup_{x\in[-m,m]}\allowbreak\bigl[|\fg_\approximate(x)|+|(\fg_\approximate)'(x)|\bigr]<\infty$}{that for all $k\in\N$, $i \in \{1,2,\dots,\ell_k\}$ it holds that
\begin{equation}\llabel{eq2}
\sup_{\theta \in K} \sup_{\approximate \in \N} \sup_{x \in [a,b]^{\ell_0}}\big[|\cN_{\approximate,i}^{k,\theta}(x)|+|\fg_\approximate (\cN_{\approximate,i}^{k,\theta}(x))\big| + |(\fg_\approximate)' (\cN_{\approximate,i}^{k,\theta}(x))|\big] <\infty,
\end{equation}}
\argument{\cref{eq:def_risk_function}}{for all $\theta\in \R^\fd$ that
\begin{equation}\llabel{eqq3}
    \int_{\R^{\ell_0} \times \R^{\ell_L}}
 \bigl|H\big(0,\theta,y\big)\bigr| \, \mu(\mathrm{d}x,\d y)<\infty.
\end{equation}}
\argument{\lref{eqq3};\cref{setting: assume};\cref{it1:uniformbounds} in \cref{lem:uniformboundsforNandderN}}{for all $\theta\in \R^\fd$ that
\begin{equation}\llabel{eq3}
    \int_{[a,b]^{\ell_0}\times \R^{\ell_L}}\sup_{m\in \N}\biggl[\bigl|(\partial_{\ell_L+l} H) \big(\cN_m^{L,\theta}(x),\theta,y\big)\bigr| + \textstyle\sum\limits_{h=1}^{\ell_L} \bigl|(\partial_h H) \big(\cN_m^{L,\theta}(x),\theta, y\big)\bigr|\biggr]\,\mu(\d x,\d y)<\infty.
\end{equation}}
\argument{\lref{eq3};\lref{eq2};\lref{item 1};\lref{item 2}}{\cref{grad_L_bd}\dott}
\end{aproof}
\begin{athm}{cor}{lem:approx:gradient:bounded}
Assume \cref{setting:activationapprox}, assume for all $m\in \N$ that $\sup_{\approximate \in \N}\sup_{x\in[-m,m]}\allowbreak\bigl[|\fg_\approximate(x)|+|(\fg_\approximate)'(x)|\bigr]<\infty$, and let $ K \subseteq \R^\fd $ be non-empty and compact.
Then
\begin{equation}
\label{eq:a_priori_bound}
  \sup\nolimits_{ \theta \in K 
  } 
  \sup\nolimits_{ \approximate \in \N
  } 
  \rbr*{ 
    \abs{ 
      \cL_\approximate( \theta ) 
    } 
    +
    \abs{ 
      \cL_{ 0 }( \theta )
    }
    + 
    \norm{ 
      ( \nabla \cL_\approximate )( \theta ) 
    } 
    +
    \norm{
      \cG( \theta )
    }
  } 
  < \infty .
\end{equation}
\end{athm}
\begin{aproof}
\argument{\cref{bound Lr};  \cref{it1:formulaLrgradient} in \cref{lem:rep_formulaLrgradient}}
{that for all $ s \in (0,\infty) $ it holds that \llabel{arg1}
$
  \sup_{ 
    \theta \in \{ \vartheta \in \R^{ \fd } \colon \| \vartheta \| \leq s \}
  } 
  \sup_{ 
    \approximate \in \N
  } 
  \norm{ 
    ( \nabla \cL_\approximate )( \theta ) 
  } 
  < \infty 
$\dott} 
\argument{\lref{arg1}; the fundamental theorem of calculus; the fact that for all $ \approximate \in \N $
it holds that 
$
  \cL_\approximate( 0 ) = \cL_{ 0 }( 0 )
$}
{that for all $ s \in (0,\infty) $ it holds that \llabel{arg2}
$
  \sup_{ 
    \theta \in \{ \vartheta \in \R^{ \fd } \colon \| \vartheta \| \leq s \}
  } 
  \sup_{ 
    \approximate \in \N
  } 
  \rbr*{ 
    \abs{ 
      \cL_\approximate( \theta ) 
    } 
    +
    \norm{ 
      ( \nabla \cL_\approximate )( \theta ) 
    } 
  } 
  < \infty 
$\dott}
\argument{\lref{arg2};\cref{it1:generalizedgrad} in \cref{thm:generalizedgrad}}
{\cref{eq:a_priori_bound}\dott}
\end{aproof}
\subsection{Weak differentiability of the objective function}\label{subsec: weak differentiability}
\begin{athm}{lemma}{weak diff}
    Assume \cref{setting:activationapprox}, assume for all $m\in \N$ that $\sup_{\approximate \in \N}\sup_{x\in[-m,m]}\allowbreak\bigl[|\fg_\approximate(x)|+|(\fg_\approximate)'(x)|\bigr]<\infty$, let 
$ 
  \varphi = 
  ( 
    \varphi( \theta ) 
  )_{ 
    \theta = ( \theta_1, \dots, \theta_{ \fd } ) \in \R^{ \fd }
  } 
 \in C^1( \R^{ \fd }, \R)$ have a compact support, 
and let 
% $ \theta = ( \theta_1, \dots, \theta_{ \fd } ) \in \R^{ \fd } $, 
$ i \in \{ 1, 2, \dots, \fd \} $. 
Then 
$
  \int_{ \R^{ \fd } } 
  |
    \cL_{ 0 }( \theta ) 
    \,
    ( \tfrac{ \partial }{ \partial \theta_i } \varphi )( \theta ) 
  |
  +
  |
    \cG_i( \theta ) 
    \,
    \varphi( \theta )
  |
  \, 
  \d \theta 
  < \infty
$
and 
\begin{equation}
\label{eq:weak_derivative_of_risk_function}
  \int_{ \R^{ \fd } } 
  \cL_{ 0 }( \theta ) 
  \,
  ( \tfrac{ \partial }{ \partial \theta_i } \varphi )( \theta ) \, \d \theta 
  =
  -
  \int_{ \R^{ \fd } } 
  \cG_i( \theta ) 
  \,
  \varphi( \theta )
  \, 
  \d \theta 
  .
\end{equation}
\end{athm}
\begin{aproof}
\argument{the assumption that $ \varphi $ has a compact support;}{that there exists $ R \in (0,\infty) $ which satisfies 
for all 
$ \theta \in \R^\fd \backslash [ - R, R ]^{\fd} $ 
that
\begin{equation}
\llabel{eq:K_compact_set_defining_property}
  \varphi( \theta ) = 0 .
\end{equation}}
\argument{\cref{lem:approx:gradient:bounded};}{that
    \begin{equation}
\llabel{eq:cG_cL_local_boundedness}
 \textstyle \sup\limits_{ \theta\in [-R,R]^{\fd}} 
  \sup\limits_{ \approximate \in \N }
  \bigl(
    | \cL_\approximate( \theta) | 
    +
    | \cL_{ 0 }( \theta ) |
    +
    \| ( \nabla \cL_\approximate )( \theta ) \|
    +
    \| \cG(\theta ) \|
  \bigr)
  < \infty .
\end{equation}
}
\argument{\lref{eq:cG_cL_local_boundedness};\lref{eq:K_compact_set_defining_property}; the assumption that $\varphi$ is compactly supported and continuous differentiable} 
{that for all $ \approximate \in \N $, 
$ \theta=(\theta_1,\dots,\theta_\fd)\in \R^{\fd} $ it holds that
\begin{equation}
\llabel{eq:Lebesgues_Majorante}
\begin{split}
&
  |
    \cL_\approximate( \theta ) 
    (\textstyle\frac{\partial}{\partial \theta_k}\varphi) ( \theta)
  |
  +
  |
    ( \tfrac{ \partial }{ \partial \theta_k } \cL_\approximate )( \theta)
    \varphi( \theta ) 
  |
\\ &
  \leq 
  \biggl[ \textstyle
    \sup\limits_{ \theta \in [ - R, R ]^{\fd} } 
    \sup\limits_{ s \in \N }
    \Bigl(
      \bigl(| \cL_s( \theta) | 
      +
      \| ( \nabla \cL_s )( \theta) \|\bigr)
     \bigl(     
      | \theta | 
      +    
      |  (\textstyle\frac{\partial}{\partial \theta_k}\varphi)(\theta) |\bigr)
    \Bigr)
  \bigg]
  \mathbbm{1}_{ [ - R, R ]^{\fd} }( \theta )
  < \infty 
  .
\end{split}
\end{equation}}
\argument{\lref{eq:Lebesgues_Majorante};\cref{it1:generalizedgrad} in \cref{thm:generalizedgrad}; Lebesgue's dominated convergence theorem;\lref{eq:K_compact_set_defining_property}; 
integration by parts; chain rule}{ that  
\begin{equation}\llabel{eq1}
\begin{split}
&
  \int_{ \R^{\fd} } 
  \cL_{ 0 }(\theta ) 
  \, 
  (\textstyle\frac{\partial}{\partial x_k}\varphi) ( \theta ) 
  \, \d \theta 
  = 
  \displaystyle\lim_{ \approximate \to \infty } 
  \left[ 
    \int_{ \R^{\fd}} 
    \cL_\approximate(\theta) 
    \,
    (\textstyle\frac{\partial}{\partial x_k}\varphi) (\theta) 
    \, \d \theta
  \right]
  \\&=
  \displaystyle
  \lim_{ \approximate \to \infty } 
  \left[ 
    \int_{[-R,R]^{\fd}}
    \cL_\approximate( \theta ) 
    \,
     (\textstyle\frac{\partial}{\partial x_k}\varphi)( \theta ) 
    \, \d \theta 
  \right] = 
  - 
  \left(
    \lim_{ \approximate \to \infty } 
    \left[ 
      \int_{[-R,R]^{\fd}} 
        (\textstyle\frac{\partial}{\partial x_k}\cL_\approximate)(\theta)
      \, \varphi( \theta ) \, \d \theta
    \right]
  \right)
  .
\end{split}
\end{equation}}
\argument{ \lref{eq1};\cref{it1:generalizedgrad} in \cref{thm:generalizedgrad};
\lref{eq:Lebesgues_Majorante};
Lebesgue's dominated convergence theorem }
{that
\begin{equation}\llabel{eq2}
\begin{split}
  &\int_{ \R^{\fd} } 
  \cL_{ 0 }(\theta ) 
  \, 
  (\textstyle\frac{\partial}{\partial x_k}\varphi) ( \theta ) 
  \, \d \theta 
  = - \displaystyle
    \bigg[ 
      \int_{[-R,R]^{\fd}} 
         \Bigl[\lim_{ \approximate \to \infty } ( \tfrac{ \partial }{ \partial \theta_k } \cL_\approximate ) ( \theta) \Bigr]
    \varphi( \theta )\, \d \theta
    \bigg]= 
  - \int_{\R^{\fd}} \fG_k( \theta) \, \varphi ( \theta) \, \d \theta .
\end{split}
\end{equation}}
\argument{\lref{eq2};\lref{eq:K_compact_set_defining_property};\lref{eq:cG_cL_local_boundedness}}{\cref{eq:weak_derivative_of_risk_function}\dott}
\end{aproof}
\subsection{Strong differentiability properties of the objective function}\label{subsec: strong differentiability}
\begin{athm}{lemma}{Existence of E dim d}
   Assume \cref{setting:activationapprox} and assume for all $m\in \N$ that $\sup_{\approximate \in \N}\sup_{x\in[-m,m]}\allowbreak\bigl[|\fg_\approximate(x)|+|(\fg_\approximate)'(x)|\bigr]<\infty$.
        Then there exists $E\in \mathcal B(\R^{\fd})$ such that
        \begin{enumerate}[label=(\roman*)]
            \item \label{existence of E dim d: item 1} it holds that $\int_{\R^{\fd}\backslash E} 1\,\d x=0$
            \item \label{existence of E dim d: item 2} it holds for all $x\in E$ that $\cL_0$ is differentiable at x, and
            \item \label{existence of E dim d: item 3} it holds for all $x=(x_1,\dots,x_{\fd})\in E$ that $(\nabla\cL_0)(x)=\fG(\theta)$.
        \end{enumerate}
\end{athm}
\begin{aproof}
    Throughout this proof let $G=(G_1,\dots,G_{\fd})\colon \R^{\fd}\to\R^{\fd}$ satisfy for all $\theta\in \R^{\fd}$ that
    \begin{equation}\llabel{def: G}
        G(x)=\begin{cases}
            (\nabla \cL_0)(\theta)& \text{ $\cL_0$ is differentiable at $\theta$}\\
             0 & \text{ $\cL_0$ is not differentiable at $\theta$}
        \end{cases}
        .
    \end{equation}
    \argument{\lref{def: G}; the fact that for all measurable $g_n\colon\R^{\fd}\to\R^{\fd}$, $n\in \N$, it holds that $\{\theta\in \R^{\fd}\colon (g_n(\theta))_{n\in \N}$ is a Cauchy sequence$\}$ is measurable; the fact that for all measurable and pointwise convergent $g_n\colon\R^\fd\to\R^\fd$, $n\in \N$, it holds that $\R^{\fd}\ni \theta\mapsto \lim_{n\to\infty}g_n(\theta)\in \R^\fd$ is measurable}{that $G$ is measurable\dott}
    \argument{\cref{Local Lipschitz};}{that \llabel{arg1}$\cL$ is locally Lipschitz continuous\dott}
    \argument{Rademacher's theorem (cf.~Evans~\cite[Theorem 5.8.6]{Evans2010});\lref{arg1}}{that there exists $\cE\in\{ A \in \mathcal B(\R^{\fd})\colon \int_{\R^{\fd}\backslash A}1\, \d x=0\}$ which satisfies for all $x\in \cE$ that \llabel{arg2}$\cL_0$ is differentiable at $x$\dott}
\argument{\lref{arg2};\cite[Lemma 3.6]{Aradexistenceofglobminv1} (applied for $\fd\curvearrowleft \fd$, $E\curvearrowleft \cE$, $f\curvearrowleft \cL_0$, $g\curvearrowleft G$ in the notation of \cite[Lemma 3.6]{Aradexistenceofglobminv1})}{that for all compactly supported $\varphi\in C^\infty (\R^{\fd},\R)$ and all $k\in \{1,2,\dots,\fd\}$ it holds that
$\int_{\R^{\fd}}|\cL_0(\theta)
\textstyle(\frac{\partial}{\partial \theta_k}\varphi)(\theta)
|+|G_{k}(\theta)
\varphi(\theta)|\,\d \theta<\infty$ and
    \begin{equation}\llabel{eq1}
    \begin{split}
\int_{\R^{\fd}}\cL_0(\theta)
\textstyle(\frac{\partial}{\partial \theta_k}\varphi)(\theta)
\,\d \theta= \displaystyle\int_{\R^{\fd}}\cL_0(\theta)
\textstyle(\frac{\partial}{\partial x_k}\varphi)(\theta)
\,\d \theta
=-\displaystyle\int_{\R^{\fd}}G_{k}(\theta)
\varphi(\theta)
    \d \theta.
\end{split}
    \end{equation}} 
    \argument{\cref{weak diff};\lref{def: G}}{that for all compactly supported $\varphi\in C^\infty (\R^{\fd},\R)$ and all $k\in \{1,2,\dots,\fd\}$  it holds that $\int_{\R^{\fd}}|\cL_0(\theta)
\textstyle(\frac{\partial}{\partial x_k}\varphi)(\theta)
|+|\fG_{k}(\theta\allowbreak)
\varphi(\theta)|\,\d \theta<\infty$ and
    \begin{equation}\llabel{eq2}
    \begin{split}
\int_{\R^{\fd}}\cL_0(\theta)
\textstyle(\frac{\partial}{\partial x_k}\varphi)(\theta)
\,\d \theta
=-\displaystyle\int_{\R^{\fd}}\fG_{k}(\theta)
\varphi(\theta)
    \d \theta.
\end{split}
    \end{equation}}
    \argument{\lref{eq1};\lref{eq2}}{that for all compactly supported $\varphi\in C^\infty (\R^{\fd}
    ,\R)$ and all $k\in \{1,2,\dots,\fd\}$  it holds that
    \begin{equation}\llabel{eq3}
        \displaystyle\int_{\R^{\fd}}G_{k}(\theta)
\varphi(\theta)
    \d \theta=\displaystyle\int_{\R^{\theta}}\fG_{k}(\theta)
\varphi(\theta).
    \end{equation}
 }
    \argument{\lref{eq3}; the fundamental lemma of calculus of variations (cf., \eg, \cite[Lemma 3.9]{Aradexistenceofglobminv1})}{that there exists 
$ 
  {\bf E} 
  \in \{ 
     A \in \cB( \R^{\fd} ) \colon 
    \int_{ \R^{\fd} \backslash  A } 1 \, \d x= 0
  \} 
$ 
which satisfies for all $\theta \in {\bf E} $, $k\in \{1,2,\dots,\fd\}$  that 
\begin{equation}
\llabel{eq:G_and_cG_agree}
  G_k( \theta) = \fG_{k}(\theta)
  \dott
\end{equation}}
\startnewargseq
\argument{\lref{def: G};\lref{eq:G_and_cG_agree};the fact that for all 
$ \theta \in \cE $
it holds that 
$ \cL_0 $ 
is differentiable at $\theta $}{that for all 
$ \theta \in ( {\bf E} \cap \cE ) $, $k\in \{1,2,\dots,\fd\}$ 
it holds that
\begin{equation}
\llabel{eq:cG_on_E_cap_cE}
 \textstyle (\frac{\partial}{\partial x_k}\cL_0)(\theta) = \fG_{k}( \theta) .  
\end{equation}
}
\argument{\lref{eq:cG_on_E_cap_cE}; the fact that for all 
$ \theta \in ( {\bf E} \cap \cE ) $
it holds that 
$
  \cL_0
$
is differentiable at $ \theta $; the fact that $\cE,\bfE\in  \{ 
     A \in \cB( \R^{\fd} ) \colon 
    \int_{ \R^{\fd} \backslash  A } 1 \, \d x= 0
  \}$}{\cref{existence of E dim d: item 1}, \cref{existence of E dim d: item 2}, and \cref{existence of E dim d: item 3}\dott}
\end{aproof}
\subsection{Local estimates for the realization functions of ANNs}\label{subsec: local estimates}
\newcommand{\smallsum}{\textstyle\sum}
% The next result, \cref{lem:gradient:left:approx} below, is a generalization of, \eg, \cite[Lemma 3.5]{Aradexistenceofglobmin}.
\begin{athm}{lemma}{lem:gradient:left:approx}
Assume \cref{setting:activationapprox}, let $a\in \R$, $b\in (a,\infty)$,
$ \theta=(\theta_1,\dots,\theta_\fd) \in \R^{ \fd } $,
$ \varepsilon \in (0, \infty) $, $z\in \R$, and assume for all $m\in \N$ that $\sup_{\approximate \in \N}\sup_{x\in[-m,m]}\allowbreak\bigl[|\fg_\approximate(x)|+|(\fg_\approximate)'(x)|\bigr]<\infty$. 
Then there exists a non-empty and open 
$
  U \subseteq \R^{ \fd } 
$ 
such that for all $ \vartheta \in U $, 
$ k \in \cu{ 1, 2, \ldots, L } $, 
$ i \in \cu{ 1, 2, \ldots, \ell_k } $, 
$ x \in [a,b]^{ \ell_0 } $
it holds that 
\begin{equation}\llabel{conclude}
  \norm{ \vartheta - \theta } < \varepsilon 
\qqandqq
  z(\cN_{ 0, i }^{ k, \vartheta }( x ) - \cN_{ 0, i }^{ k, \theta }( x ) )\leq 0.
\end{equation}
\end{athm}
\begin{aproof}
Throughout this proof assume without loss of generality that $z\in\{-1,1\}$.
\argument{\cref{Local Lipschitz N};}{that there exists $C\in (0,\infty)$ which satisfies for all $k\in \{1,2,\dots,L\}$, $\varTheta=(\varTheta_1,\dots,\varTheta_\fd)$, $\vartheta=(\vartheta_1,\dots,\vartheta_\fd)\in \R^{\fd}$, $x\in[a,b]^{\ell_0}$ with $\max_{i\in \{1,2,\dots,\fd\}}\allowbreak\max\{|\varTheta_i|,|\vartheta_i|\}\leq \|\theta\|+\fd^{-1}\varepsilon$ it holds that
    \begin{equation}\llabel{def: C}
           \begin{split}
       & \textstyle\max_{i\in\{1,2,\dots,\ell_{k}\}}|\cN_{0,i}^{k,\varTheta}(x)-\cN_{0,i}^{k,\vartheta}(x)|
  \leq
  C\br*{ 
    \max\nolimits_{ j \in \cu{ 1, 2, \ldots, \bfd_{k } } }
    \abs{ \theta_j - \vartheta_j } 
  }
  .
       \end{split}
    \end{equation}}
In the following let $ \fC_k\in (0, \infty) $, $k\in \N$
satisfy 
for all $ k \in \N$ that
$
  \fC_1 = \max \cu{\ell_0 \abs{a}, \ell_0 \abs{b},  1 }
$ 
and
\begin{equation}
\llabel{eq:def_Ck_constants}
%    \forall \, k \in \cu{1,  \ldots, L - 1} \colon 
   \fC_{ k + 1 } = 
   2 
   \fC_k \max\{1,C\}
   ,
\end{equation}
let $ \delta \in (0, \infty) $ 
satisfy 
$
  \delta = \min\cu{ 1, \varepsilon ( 2 \fC_L \fd )^{ - 1 } } 
$, 
and let 
$ U \subseteq \R^{ \fd } $ 
satisfy
\begin{multline}
\llabel{eq:def_set_U}
  U = 
  \biggl\{
    \vartheta \in \R^{ \fd } \colon 
    \biggl(
      \Bigl[ 
        \forall \, k \in \cu{ 1, \ldots, L }, 
        i \in \cu{ 1, \ldots, \ell_k }, 
        j \in \cu{ 1, \ldots, \ell_{ k - 1 } } \colon 
        \abs{ \w{ k, \vartheta }_{ i, j } - \w{ k, \theta }_{ i, j } } < \delta  
      \Bigr]
%       \Bigr. 
\\
        \qquad 
        \wedge
%         \Bigl. 
        \br*{
          \forall \, k \in \cu{ 1, \ldots, L }, i \in \cu{ 1, \ldots, \ell_k } 
          \colon 
            z(\b{ k, \theta }_i - 2 z\fC_k \delta) 
            <
            z\b{ k, \vartheta }_i 
            <
            z(\b{ k,\theta  }_i -z \fC_k \delta )
        } 
    \biggr)
  \biggr\} 
  .
\end{multline}
\startnewargseq
\argument{\lref{eq:def_set_U}} 
{that $ U \subseteq \R^{ \fd } $ is non-empty and open\dott} 
\argument{\lref{eq:def_Ck_constants}}{ 
for all 
$ k \in \N $ 
that \llabel{arg1}
$
  \fC_{ k + 1 } > 2 \fC_k 
$ 
and 
$ \fC_k \geq 1 $\dott}
\argument{\lref{arg1}; \lref{eq:def_set_U};the fact that $|z|=1$ }
{that for all 
$ 
  \vartheta =(\vartheta_1,\dots,\vartheta_\fd)\in U 
$, 
$ i \in \cu{ 1, 2, \ldots, \fd } $ 
it holds that 
\begin{equation}\llabel{eq1}
  \abs{ \vartheta_i - \theta_i } 
  < \max\{ \delta, 2 \fC_1 \delta, 2 \fC_2 \delta, \dots, 2 \fC_L \delta \} 
  = 2 \fC_L \delta 
  \leq 
  2 \fC_L 
  \bigl( \varepsilon ( 2 \fC_L \fd )^{ - 1 } \bigr)
  = \fd^{ - 1 } \varepsilon 
  \dott
\end{equation}}
\argument{\lref{eq1};}{for all $ \vartheta=(\vartheta_1,\dots,\vartheta_\fd) \in U $ that
\begin{equation}\llabel{eq2}
\textstyle
  \norm{\vartheta - \theta } 
  =
  \bigl[
    \sum_{ i = 1 }^{ \fd }
    \abs{ \vartheta_i - \theta_i }^2
  \bigr]^{ 1 / 2 }
  \le 
  \fd 
  \bigl[
    \max\nolimits_{ i \in \cu{ 1, 2, \ldots, L } } 
    \abs{ \vartheta_i - \theta_i } 
  \bigr]
  < \varepsilon \dott
\end{equation}}
\argument{\cref{wb};\cref{eq:def_NN_realization};\lref{eq:def_set_U}}
{for all 
$ \vartheta \in U $, 
$ i \in \cu{ 1, 2, \dots, \ell_1 } $, 
$ x = ( x_1, \dots, x_{ \ell_0 } ) \in [a,b]^{ \ell_0 } $
that
\begin{equation}
\begin{split}
  z(\cN_{ 0, i }^{ 1, \vartheta }( x ) - 
  \cN_{ 0, i }^{ 1, \theta }( x ) )
&
  =
 z ( 
    \b{ 1, \vartheta }_i - \b{ 1, \theta }_i 
  ) 
  + 
  \smallsum_{ j = 1 }^{ \ell_0 } 
  z( 
    \w{ 1, \vartheta }_{ i, j } - \w{ 1, \theta }_{ i, j } 
  ) x_j 
\\
& 
  < 
  - z^2\fC_1 \delta 
  + 
  \smallsum_{ j = 1 }^{ \ell_0 } 
  |
    ( 
      \w{ 1, \vartheta }_{ i, j } - \w{ 1, \theta }_{ i, j } 
    ) 
    x_j 
  |
\\ &
  \leq
    - \fC_1 \delta 
    + \delta \bigl( \smallsum_{ j = 1 }^{ \ell_0 } \abs{ x_j } \bigr)
  \leq 
    - \fC_1 \delta + \ell_0 \delta \max\cu{ \abs{a}, \abs{b} } 
  \le 0 \dott
\end{split}
\end{equation}}
It thus remains to prove an analogous inequality for the subsequent layers. 
For this let
$ 
  \vartheta \in U 
$,
$
  k \in \N \cap (0,L) 
%   \cu{ 1, 2, \ldots, L - 1 } 
$, 
$
  i \in \cu{ 1, 2, \ldots, \ell_{ k + 1 } }
$, 
$ x \in [a,b]^{ \ell_0 } $, 
let 
$ 
  \bfd \in \N 
$
satisfy 
$
  \bfd = \ell_{ k + 1 } \ell_k + 1 + \sum_{ j = 1 }^{ k } \ell_j ( \ell_{ j - 1 } + 1 ) 
$, 
and let 
$
  \psi \in \R^{ \fd } 
$
satisfy 
\begin{equation}
\label{eq:def_psi_zwischenvector}
  \psi = 
  ( 
    \vartheta_1, \vartheta_2, \ldots, \vartheta_{ \bfd - 1 }, \theta_{ \bfd }, 
    \theta_{ \bfd + 1 }, \dots, \theta_{ \fd }
  )
  .
\end{equation}
\argument{\cref{wb};\cref{eq:def_NN_realization}; \lref{eq:def_set_U}; the fact that $\vartheta\in U$} 
{ that
\begin{equation}
\llabel{eq:vartheta_estimate_from_above_general_k}
\begin{split}
  z\cN_{ 0, i }^{ k + 1, \vartheta }(x) 
& = 
 z \cN_{ 0, i }^{ k + 1, \psi }( x ) 
  + 
  z( 
    \b{ k + 1, \vartheta }_i - \b{ k + 1, \psi }_i 
  ) 
\\ & =
  z\cN_{ 0, i }^{ k + 1, \psi }( x ) 
  + 
 z ( 
    \b{ k + 1, \vartheta }_i - \b{ k + 1, \theta }_i 
  ) 
< 
 z \cN_{ 0, i }^{ k + 1, \psi }( x ) - \fC_{ k + 1 } \delta 
  .
\end{split}
\end{equation}}
\argument{\lref{def: C};\lref{eq1}}{ that
\begin{equation}
\llabel{eq:DNN_lipschitz_estimate}
\begin{split}
&
  \abs{ 
    \cN_{ 0, i }^{ k + 1, \theta }( x ) 
    - 
    \cN_{ 0, i }^{ k + 1, \psi }( x ) 
  } 
\leq C
  \br*{ 
    \max\nolimits_{ i \in \cu{ 1, 2, \ldots, \bfd_{k+1} } } 
    \abs{ \theta_i - \psi_i } 
  }
  .
\end{split}
\end{equation}}
\argument{\lref{eq:def_set_U};}{that 
\begin{equation}\llabel{eq3}
  \max_{ i \in \cu{ 1, 2, \ldots, \bfd_{k+1} } } 
  \abs{ \theta_i - \psi_i } 
  =
  \max_{ i \in \cu{ 1, 2, \ldots, \bfd - 1 } } 
  \abs{ \theta_i - \vartheta_i } 
  \leq
  \max\{ \delta, 2 \fC_1 \delta, 2 \fC_2 \delta, \dots, 2 \fC_k \delta \}
  = 
  2 \fC_k \delta 
  .
\end{equation}}
\argument{\lref{eq3};\lref{eq:def_Ck_constants};\lref{eq:DNN_lipschitz_estimate}} {that 
\begin{equation}\llabel{eq5}
\begin{split}
&
  \abs{ 
    \cN_{ 0, i }^{ k + 1, \theta }( x ) 
    - 
    \cN_{ 0, i }^{ k + 1, \psi }( x ) 
  } \leq
    2 \fC_k C \delta \leq
  \fC_{ k + 1 } \delta \dott
\end{split}
\end{equation}}
\argument{\lref{eq5}; \lref{eq:vartheta_estimate_from_above_general_k}} 
{that
\begin{equation}
\begin{split}
 z \cN_{ 0, i }^{ k + 1, \vartheta }(x) 
& 
<  
  z\cN_{ 0, i }^{ k + 1, \psi }( x ) - \fC_{ k + 1 } \delta 
=
  z\cN_{ 0, i }^{ k + 1, \theta }( x ) 
  + 
 z \bigl( 
    \cN_{ 0, i }^{ k + 1, \psi }(x) - \cN_{ 0, i }^{ k + 1, \theta }( x )  
  \bigr)
  - \fC_{k+1} \delta 
\\
&  
\leq
  z\cN_{ 0, i }^{ k + 1, \theta }( x ) 
  + \abs{ \cN_{ 0, i }^{ k + 1, \psi }(x) - \cN_{ 0, i }^{ k + 1, \theta }( x ) } 
  - \fC_{k+1} \delta 
\leq 
  \cN_{ 0, i }^{ k + 1, \theta }( x ) \dott
\end{split}
\end{equation}}
\end{aproof}
\subsection{Continuity properties of generalized gradients}\label{subsec: continuity propeties}
% The next results, \cref{lem:gradient:convergence,lem:gradient:approx:sequence}, are generalizations of, \eg, \cite[Lemma 3.6 and Lemma 3.7]{Aradexistenceofglobmin}.
\begin{athm}{lemma}{lem:gradient:convergence}[Continuity points of the generalized gradient function]
Assume \cref{setting:activationapprox}, assume for all $m\in \N$ that $\sup_{\approximate \in \N}\sup_{x\in[-m,m]}\allowbreak\bigl[|\fg_\approximate(x)|+|(\fg_\approximate)'(x)|\bigr]<\infty$,
and let 
$ 
  \theta = ( \theta_n )_{ n \in \N_0 } \colon \N_0 \to \R^{ \fd } 
$
satisfy 
for all 
$ k \in \cu{ 1, 2, \ldots, L } $, 
$ i \in \cu{ 1, 2, \ldots, \ell_k } $, 
$ x \in [a,b]^{ \ell_0 } $
that 
\begin{equation}
\llabel{eq:limit_theta_assumption}
  \limsup\nolimits_{ n \to \infty } 
  \bigl(
    \| \theta_n - \theta_0 \| 
    +
    |
      (\mathscr{d}_\func \fg_0)( 
        \cN_{ 0, i }^{ k, \theta_n }( x ) 
      ) 
      - 
      (\scrd_\func\fg_0)( 
        \cN_{ 0, i }^{ k, \theta_0 }( x ) 
      ) 
    | 
  \bigr)
  = 0
  .
\end{equation}
Then 
$
  \limsup_{ n \to \infty } 
  \| \cG( \theta_n ) - \cG ( \theta_0 ) \| = 0 
$.
\end{athm}
\begin{aproof}
\argument{\cref{Local Lipschitz N};}
{that 
for all $ k \in \{ 1, 2, \dots, L \} $, 
$ j \in \{ 1, 2, \dots, \ell_k \} $
it holds that 
\begin{equation}
\llabel{eq:continuous_realization_function}
\textstyle
  \limsup_{ n \to \infty }
  \sup_{ x \in [a,b]^{ \ell_0 } }
  |
    \cN_{ 0, j }^{ k, \theta_n }( x ) 
    -
    \cN_{ 0, j }^{ k, \theta_0 }( x ) 
  |
  = 0 .
\end{equation}}
\argument{
\lref{eq:limit_theta_assumption};
\lref{eq:continuous_realization_function}; 
 \cref{thm:generalizedgrad}; 
Lebesgue's dominated convergence theorem} {that
$
  \limsup_{ n \to \infty } \allowbreak
  \| \cG( \theta_n ) - \cG( \theta_0 ) \| = 0
$\dott}
\end{aproof}
\begin{athm}{lemma}{lem:gradient:approx:sequence}
Assume \cref{setting:activationapprox}, let $ \theta \in \R^{ \fd } $, $ E \in \cB( \R^{ \fd } ) $ satisfy 
$
  \int_{ \R^{ \fd } \backslash E } 1 \, \d \vartheta = 0
$, assume for all $m\in \N$ that $\sup_{\approximate \in \N}\sup_{x\in[-m,m]}\allowbreak\bigl[|\fg_\approximate(x)|+|(\fg_\approximate)'(x)|\bigr]<\infty$, and assume $\min_{ z \in \{ -1, 1 \}} \bigl( \sum_{ x \in S } \limsup_{ h \searrow 0 }\allowbreak | (\scrd_\func A_0)( x + z h ) - (\scrd_\func A_0)( x ) | \bigr) = 0$.
Then there exists 
$
  \vartheta = ( \vartheta_n )_{ n \in \N } \colon \N \to E
$ 
such that
\begin{equation}
\llabel{eq:vartheta_to_prove_construction_sequence}
  \limsup\nolimits_{
    n \to \infty
  } 
  \bigl( 
    \norm{ \vartheta_n - \theta } 
    + 
    \norm{ \cG( \vartheta_n ) - \cG( \theta ) } 
  \bigr)
  = 0 .
\end{equation}
\end{athm}
\begin{aproof}
\argument{the assumption that $\min_{ z \in \{ -1, 1 \}} \bigl( \sum_{ x \in S } \limsup_{ h \searrow 0 }\allowbreak | (\scrd_\func A_0)( x + z h ) - (\scrd_\func A_0)( x ) | \bigr) = 0$;the assumption that $A_0|_{\R\backslash S}\in C(\R\backslash S,\R)$}{that there exists $z\in \R$ which satisfies for all $x\in \R$ that
\begin{equation}\llabel{def: z}
     \limsup_{ h \searrow 0 }\allowbreak | (\scrd_\func A_0)( x + z h ) - (\scrd_\func A_0)( x ) |=0.
\end{equation}}
\startnewargseq
\argument{\cref{lem:gradient:left:approx};}{ 
that there exist 
non-empty and open $ U_n \subseteq \R^{ \fd } $, $ n \in \N $, 
which satisfy for all 
$ n \in \N $, 
$ \vartheta \in U_n $,
$ k \in \cu{ 1, 2, \ldots, L} $, 
$ i \in \cu{ 1, 2, \ldots, \ell_k } $, 
$ x \in [a,b]^{ \ell_0 } $ 
that 
\begin{equation}
\llabel{eq:property_sets_Un}
  \norm{ \vartheta - \theta } < \tfrac{ 1 }{ n } 
\qqandqq
 z \cN_{ 0, i }^{ k, \vartheta }(x) 
  \le 
  z\cN_{ 0, i }^{ k, \theta }( x )\dott
\end{equation}}
\startnewargseq
\argument{the assumption that 
$
  \int_{ \R^{ \fd } \backslash E } 1 \, \d \vartheta = 0
$ ;}
{that for all $ n \in \N $ it holds that 
$
  ( U_n \cap E ) \not= \emptyset
$\dott} 
In the following let 
$
  \vartheta = ( \vartheta_n )_{ n \in \N } \colon \N \to E 
$ 
satisfy for all $ n \in \N $ that 
\begin{equation}
\llabel{eq:construction_of_sequence_vartheta}
  \vartheta_n \in U_n
  .
\end{equation}
\startnewargseq
\argument{\lref{eq:property_sets_Un};}{
that for all $ n \in \N $ it holds
that \llabel{arg1}
$
  \norm{ \vartheta_n - \theta } < \frac{ 1 }{ n } 
$\dott} 
\argument{\lref{arg1};}
{that 
\begin{equation}
\llabel{eq:vartheta_n_is_convergent}
\textstyle
  \limsup_{ n \to \infty } \| \vartheta_n - \theta \| = 0 
  \dott
\end{equation}}
\argument{\lref{eq:vartheta_n_is_convergent};
\cref{Local Lipschitz} (applied for every $ k \in \{ 1, 2, \dots, L \} $ 
with 
$
  L \with k
$
in the notation of \cref{Local Lipschitz})}
{that for all 
$
  k \in \cu{ 1, 2, \ldots, L }
$, 
$
  i \in \cu{ 1, 2, \ldots, \ell_k } 
$, 
$
  x \in [a,b]^{ \ell_0 }
$
it holds that
\begin{equation}
\llabel{eq:realization_functions_sequence_is_convergent}
\textstyle
  \limsup_{ n \to \infty } 
  |
    \cN_{ 0, i }^{ k, \vartheta_n }( x ) 
    -
    \cN_{ 0, i }^{ k, \theta }( x ) 
  |
  = 0
  \dott
\end{equation}}
\argument{
\lref{eq:property_sets_Un};\lref{eq:construction_of_sequence_vartheta};}
{for all \llabel{arg3}
$
  n \in \N
$, 
$
  k \in \cu{ 1, 2, \ldots, L }
$, 
$
  i \in \cu{ 1, 2, \ldots, \ell_k }
$,  
$
  x \in [a,b]^{ \ell_0 }
$
that 
$
  z\cN_{ 0, i }^{ k, \vartheta_n }( x ) 
  \le 
  z\cN_{ 0, i }^{ k, \theta }( x ) 
$\dott} 
\argument{\lref{def: z};\lref{arg3};
\lref{eq:realization_functions_sequence_is_convergent}}
{that for all 
$ 
  x \in [a,b]^{ \ell_0 } 
$, 
$
  k \in \cu{ 1, 2, \ldots, L } 
$, 
$
  i \in \cu{ 1, 2, \ldots, \ell_k } 
$ 
it holds that 
\begin{equation}\llabel{arg4}
  \limsup_{ n \to \infty } \allowbreak
  |
   (\scrd_\func \fg_0)( 
      \cN_{ 0, i }^{ k, \vartheta_n }( x ) 
    ) 
   \allowbreak -
     (\scrd_\func \fg_0)( 
      \cN_{ 0, i }^{ k, \theta }( x ) 
    )
  | = 0\dott
  \end{equation}}
\argument{\lref{arg4};\cref{lem:gradient:convergence};\lref{eq:vartheta_n_is_convergent}}
{that \llabel{arg5}
$
  \limsup_{ n \to \infty } \allowbreak
  \| \cG( \vartheta_n ) - \cG( \theta ) \| = 0
$\dott} 
\argument{\lref{arg5};\lref{eq:vartheta_n_is_convergent}}
{\lref{eq:vartheta_to_prove_construction_sequence}\dott} 
\end{aproof}
\subsection{Generalized gradients as limiting Fréchet subgradients}
\label{subsec: frechet definition}
In the next notion, \cref{def:limit:subdiff} below, we recall the concepts of Fr\'{e}chet subgradients and limiting Fréchet subgradients (cf., \eg, \cite[Definition 2.10]{Bolte2006} and \cite[Definition 8.3]{TyrrelVariation}). The precise form of \cref{def:limit:subdiff} is a slightly modified version of \cite[Definition 3.1]{Aradexistenceofglobmin}.
\begin{definition}[Fr\'{e}chet subgradients and limiting Fréchet subgradients]
\label{def:limit:subdiff}
Let $ n \in \N $, $ f \in C( \R^n, \R) $, $ x \in \R^n $.
Then we denote by 
$ (\cD f)(x) \subseteq \R^n $ 
the set given by
\begin{equation}
  ( \cD f)( x )
  = 
  \cu*{ 
    y \in \R^n \colon 
    \left[
      \liminf_{\R^n \backslash \cu{  0 } \ni h \to 0 } 
      \rbr*{ \frac{f(x + h ) - f ( x ) - \spro{y , h } }{\norm{h}} } \geq 0  
    \right]
  } 
\end{equation}
and we denote by 
$
  (\mathbb D f)(x) \subseteq \R^n 
$ 
the set given by
\begin{equation}
\label{def:limit:subdiff:eq}
  (\mathbb D f)( x ) =
  \textstyle\bigcap_{ \varepsilon \in (0, \infty) } 
  \overline{
    \br*{
      \cup_{ 
        y \in 
        \cu{ 
          z \in \R^n \colon \norm{ x - z } < \varepsilon 
        }
      } 
      ( \cD f )( y ) 
    } 
  }
  .
\end{equation}
\end{definition}
% The next result, \cref{prop:loss:gradient:subdiff} below, is a generalization of, \eg, \cite[Proposition 3.3]{Aradexistenceofglobmin}.
\begin{athm}{prop}{prop:loss:gradient:subdiff}
\cfadd{def:limit:subdiff}
Assume \cref{setting:activationapprox}, assume for all $m\in \N$ that $\sup_{\approximate \in \N}\sup_{x\in[-m,m]}\allowbreak\bigl[|\fg_\approximate(x)|+|(\fg_\approximate)'(x)|\bigr]<\infty$, assume $\min_{ z \in \{ -1, 1 \}} \bigl( \sum_{ x \in S } \limsup_{ h \searrow 0 }\allowbreak | (\scrd_\func A_0)( x + z h ) - (\scrd_\func A_0)( x ) | \bigr) = 0$, and let $ \theta \in \R^{ \fd } $.
Then 
$
  \cG( \theta ) \in ( \mathbb D \cL_{ 0 } )( \theta )
$
\cfload.
\end{athm}
\begin{aproof}
\argument{\cref{Existence of E dim d};}{that 
there exists 
$ E \in \cB( \R^{ \fd } ) $ 
which satisfies  
$
  \int_{ \R^{ \fd } \backslash E } 1 \, \d \vartheta = 0 
$, 
which satisfies for all $ \vartheta \in E $ 
that $\cL_0$ is differentiable at $\vartheta$, 
and which satisfies for all 
$\vartheta \in E$ that 
\begin{equation}
\label{eq:differentiable_property}
  ( \nabla \cL_{ 0 } )( \vartheta ) 
  = \cG( \vartheta )
  .
\end{equation}}
\startnewargseq
\argument{\cref{eq:differentiable_property}; \cite[Lemma 3.8]{Aradexistenceofglobmin}}
{that for all $ \vartheta \in E $ 
it holds that 
\begin{equation}
\label{eq:generalized_gradient_subdifferential}
  \cG( \vartheta ) \in ( \cD \cL_{ 0 } )( \vartheta )
  .
\end{equation}}
\argument{the fact that 
$ \int_{ \R^{ \fd } \backslash E } 1 \, \d \vartheta = 0 $; \cref{lem:gradient:approx:sequence}} { 
that there exists 
$
  \vartheta = ( \vartheta_n )_{ n \in \N } \colon \N \to E 
$ 
which satisfies 
\begin{equation}
\label{eq:proof_construction_of_vartheta_sequence}
\textstyle
  \limsup_{ n \to \infty } 
  \bigl(
    \| \vartheta_n - \theta \| 
    + 
    \| \cG( \vartheta_n ) - \cG( \theta ) \|
  \bigr)
  = 0
  \dott
\end{equation}}
\argument{
\cref{eq:proof_construction_of_vartheta_sequence,eq:generalized_gradient_subdifferential};} 
{that 
$
  \cG( \theta ) 
  \in 
  ( \mathbb D \cL_{ 0 } )( \theta )
$\dott}
\end{aproof}

\cfclear
\begin{athm}{cor}{cor:cG_equal_to_gradient}
\cfadd{def:limit:subdiff}
Assume \cref{setting:activationapprox},  assume for all $m\in \N$ that $\sup_{\approximate \in \N}\sup_{x\in[-m,m]}\allowbreak\bigl[|\fg_\approximate(x)|+|(\fg_\approximate)'(x)|\bigr]<\infty$, and assume $\min_{ z \in \{ -1, 1 \}} \bigl( \sum_{ x \in S } \limsup_{ h \searrow 0 }\allowbreak | (\scrd_\func A_0)( x + z h ) - (\scrd_\func A_0)( x ) | \bigr) = 0$. 
Then it holds for all 
$
  \theta \in 
  (\cup_{ 
    U \subseteq \R^{ \fd }, 
    \, U \text{ is open}, \, 
    ( \cL_{ 0 } )|_U \in C^1( U, \R )
  }
  U)
$
that 
\begin{equation}\llabel{conclude}
  \cG( \theta ) = ( \nabla \cL_{ 0 } )( \theta )
\end{equation}
\cfload.
\end{athm}
\begin{aproof}
\argument{ \cite[item (iv) in Lemma 3.8]{Aradexistenceofglobmin} 
(applied with 
$
  n \with \fd 
$,
$  
  f \with \cL_{ 0 }
$
in the notation of \cite[Lemma 3.8]{Aradexistenceofglobmin}); 
\cref{prop:loss:gradient:subdiff}}
{that for all open $ U \subseteq \R^{ \fd } $ 
and all $ \theta \in U $
with 
$ ( \cL_{ 0 } )|_U \in C^1( U, \R) $
it holds that 
\begin{equation}\llabel{eq1}
  \cG( \theta ) 
  \in 
  ( \mathbb D \cL_{ 0 } )( \theta )
  =
  \{
    ( \nabla \cL_{ 0 } )( \theta )
  \}  
  \dott
\end{equation}}
\argument{\lref{eq1};}{ for all open $ U \subseteq \R^{ \fd } $ 
and all $ \theta \in U $
with 
$ ( \cL_{ 0 } )|_U \in C^1( U, \R) $
that 
$
  \cG( \theta ) = 
  ( \nabla \cL_{ 0 } )( \theta )
$\dott}
\end{aproof}
\begin{athm}{theorem}{Theorem X}
%For every $n \in \N$, $f \colon \R^n \to \R$ let $\mathbf{T}_f \colon \R^n \to \R^n$ denote the realization of the Tensorflow gradient of $f$
Let $ \fd, L\in \N $,
$ \ell_0,\ell_1,\dots,\ell_L\in \N $
%$\delta \in (0,\infty)$,
satisfy 
$
  \fd = \sum_{k=1}^L \ell_k ( \ell_{k-1} + 1 )
$, 
for every $\activate\in C(\R,\R)$,
$ 
  \theta = ( \theta_1, \dots, \theta_{ \fd } ) \in \R^{ \fd }$
 let $\mN^{ k, \theta }_{\activate}=(\mN^{ k, \theta }_{\activate,1},\dots,\mN^{ k, \theta }_{\activate,\ell_k}) \colon \R^{ \ell_0 }\allowbreak \to \R^{ \ell_k} $, $k\in \{0,1,\dots,L\}$, satisfy for all $k\in \{0,1,\dots,L-1\}$, $x=(x_1,\dots, x_{\ell_0})\in \R^{\ell_0}$, $i\in\{1,2,\dots,\ell_{k+1}\}$ that
\begin{equation}
\begin{split}
  \mN^{ k+1, \theta }_{\activate,i}( x ) &= \theta_{\ell_{k+1}\ell_{k}+i+\sum_{h=1}^{k}\ell_h(\ell_{h-1}+1)}\\
  &+\textstyle\sum_{j=1}^{\ell_{k}}\theta_{(i-1)\ell_{k}+j+\sum_{h=1}^{k}\ell_h(\ell_{h-1}+1)}\big(x_j\indicator{\{0\}}(k) 
  +\activate(\mN^{k,\theta}_{\activate,j}(x))\indicator{\N}(k)\big),
  \end{split}
\end{equation}
let $\mu \colon \cB(\R^{\ell_0}\times \R^{\ell_L})\to [0,\infty]$ be a finite measure,
let $H \in C^1(\R^{\ell_L} \times \R^{\ell_L}\times\R^{\fd},\R)$, for every $\activate \in C(\R,\R)$ let $\cL_\activate \colon \R^\fd \to \R$ satisfy for all $\theta \in \R^\fd$  that $\int_{\R^{\ell_0}\times \R^{\ell_L}}
  |H(\mathcal{N}_{\activate}^{L,\theta}(x),y,\theta)|\allowbreak\, \mu (\mathrm{d}x,\mathrm{d}y)<\infty$ and
\begin{equation}
\textstyle 
  \cL_\activate( \theta ) 
  = 
  \int_{\R^{\ell_0}\times \R^{\ell_L}}
  H(\mathcal{N}_{\activate}^{L,\theta}(x),y,\theta)\, \mu (\mathrm{d}x,\mathrm{d}y)
  ,
\end{equation}
assume that $\{x\in \R^{\ell_0}\colon (\exists\, y\in \R^{\ell_L}\colon (x,y)\in \mathrm{supp}(\mu))\}$ is bounded, assume for all $r\in (0,\infty)$, $X\in [-r,r]^{\ell_0}$, $\Theta\in [-r,r]^{\fd}$ that
\begin{equation}
   \textstyle \sup\limits_{x\in [-r,r]^{\ell_0}}\sup\limits_{y\in \R^{\ell_L}}\sup\limits_{\theta\in [-r,r]^\fd}\Bigl(\frac{(\|\nabla_x H(x,y,\theta)\|+\|\nabla_\theta H(x,y,\theta)\|)\mathbbm 1_{\mathrm{supp}(\mu)}(x,y)}{1+|H(X,y,\Theta)|}\Bigr)<\infty,
\end{equation} 
let $S\subseteq \R$ be finite, let $\scrA\in C(\R,\R)$ satisfy $\restr{\scrA}{\R \backslash S} \in C^1(\R \backslash S,\R)$, let $\scra \colon \R \to \R$ satisfy $\restr{\scra}{\R \backslash S}=\nabla(\restr{\scrA}{\R \backslash S})$, assume $\min_{ z \in \{ -1, 1 \}} \bigl( \sum_{ x \in S } \limsup_{ h \searrow 0 }\allowbreak | \scra( x + z h ) - \scra( x ) | \bigr) = 0$, and let $\ZZ\in \N_0$.
Then 
\begin{enumerate}[label=(\roman*)]
\item \label{item 0: Theorem X} it holds for all $A\in C^1(\R,\R)$ that $\cL_A\in C^1(\R^\fd,\R)$,
\item \label{item 1: Theorem X} there exists a unique $\cG \colon \R^\fd \to \R^\fd$ which satisfies for all $\theta \in \R^\fd$ and all $(A_n)_{n\in \N} \subseteq C^1(\R,\R)$ with $\forall \, x \in \R \colon \exists \, m \in \N \colon \sum_{n=m}^\infty (|\scrA(x)-A_n(x)|+|\scra(x)-(A_n)'(x)|) = 0$ and $\forall \,m \in \N \colon \sup_{n \in \N} \sup_{x \in [-m,m]} \bigl[(|A_n(x)|  + |(A_n)'(x)| )(\ZZ+\mathbbm 1_{\{\infty\}}(\mathrm{supp}(\mu))\bigr]<\infty$ that  
\begin{equation}
\textstyle\limsup_{n \to \infty} \norm{(\nabla\cL_{A_n})(\theta) - \cG(\theta)}=0,
\end{equation}
\item \label{item 2: Theorem X} it holds for all $\theta\in \R^\fd$ that $\cG( \theta ) \in ( \mathbb D \cL_{ 0 } )( \theta )$, and
\item \label{item 3: Theorem X}  it holds for all open $U\subseteq \R^d$ with $\cL_\scrA|_{U}\in  C^1( U, \R )$ that
    \begin{equation}
        \cG|_{U}=\nabla (\cL_\scrA|_{U})
        \end{equation}
\end{enumerate}
(cf. \cref{def:limit:subdiff}).
\end{athm}
\begin{aproof}
% \argument{the assumption that for all $m\in \N$ it holds that $\sup_{x\in [-m,m]}\allowbreak [|\scra(x)|(\ZZ+\indicator{\{\infty\}}{(|\mathrm{supp}(\mu)|)})]<\infty$; the fact that $z\geq 0$}{that for all $m\in \N$ it holds that 
% \begin{equation}\llabel{eq1}
%     \sup_{x\in [-m,m]} [|\scra(x)|(\indicator{\{\infty\}}{(|\mathrm{supp}(\mu)|)})]<\infty.
% \end{equation}}
\argument{\cref{item 1: Theorem X part 1} in \cref{thm:mainthm_sequence: part 1}}{\cref{item 0: Theorem X}\dott}
\startnewargseq
\argument{the assumption that $\min_{ z \in \{ -1, 1 \}} \bigl( \sum_{ x \in S } \limsup_{ h \searrow 0 }\allowbreak | \scra( x + z h ) - \scra( x ) | \bigr) = 0$; the assumption that $\restr{\scrA}{\R \backslash S} \in C^1(\R \backslash S,\R)$; the assumption that $\restr{\scra}{\R \backslash S}=\nabla(\restr{\scrA}{\R \backslash S})$}{that \llabel{arg1} $\scra$ is locally bounded\dott}
\argument{\lref{arg1};}
{for all $m\in \N$ that
\begin{equation}\llabel{eq1}
     \textstyle \sup_{x\in [-m,m]} \bigl[|\scra(x)(\ZZ+|\indicator{\{\infty\}}{(|\mathrm{supp}(\mu)|)})\bigr]<\infty.
\end{equation}}
    \argument{\lref{eq1};\cref{item 2: Theorem X part 1} in \cref{thm:mainthm_sequence: part 1};}{\cref{item 1: Theorem X}\dott}
    \startnewargseq
    \argument{\cref{lem: existence of A_n};the fact that $\scra$ is locally bounded;}{that there exist $A_n \in C^1(\R,\R)$, $n\in \N$, which satisfy
    \begin{enumerate}[label=(\Roman*)]
        \item \llabel{it1} it holds for all $ x \in \R$ that there exists $m \in \N$ such that $\sum_{n=m}^\infty (|\scrA(x)-A_n(x)|+|\scra(x)-(A_n)'(x)|) = 0$ and
        \item \llabel{it2} it holds for all $m \in \N $ that $\sup_{n \in \N} \sup_{x \in [-m,m]} ( |A_n(x)|  + |(A_n)'(x)| )\allowbreak<\infty$.
 \end{enumerate}}
 \startnewargseq
\argument{\lref{it1};\lref{it2};\cref{prop:loss:gradient:subdiff}}{\cref{item 2: Theorem X}\dott}
    \startnewargseq
\argument{\lref{it1};\lref{it2};\cref{cor:cG_equal_to_gradient}}{\cref{item 3: Theorem X}\dott}
\end{aproof}
%giu cho ky
% \section{Example}\label{sec: example}
% \begin{example}
%     Let $m,n\in \N$, $\bfC\colon (0,1)^n\times \R^n\to\R$ satisfy for all $x=(x_1,\dots,x_n)$, $y\in (y_1,\dots,y_n)\in \R^n$ that 
%     \begin{equation}
%         \bfC(x,y)=\sum_{i=1}^n \ln(x_i)y_i
%     \end{equation} 
%      and let $\bfA\colon \R^n\to\R^n$ satisfy for all $x=(x_1,\dots,x_n)\in \R^n$ that
%     \begin{equation}
%         \textstyle A(x)=\biggl(\frac{\exp(x_1)}{\sum_{i=1}^n\exp(x_i)},\frac{\exp(x_2)}{\sum_{i=1}^n\exp(x_i)},\dots,\frac{\exp(x_n)}{\sum_{i=1}^n\exp(x_i)}\biggr).
%     \end{equation}
%      Then there exists $H\in C^1(\R^n\times\R^n\times \R^m,\R)$ satisfy for all $x,y\in \R^n$, $z\in \R^m$ that 
%     \begin{equation}
%         H(x,y,z)=\bfC(A(x),y).
%     \end{equation}
% \end{example}
\subsubsection*{Acknowledgements}
This work has been partially funded by the European Union (ERC, MONTECARLO, 101045811). The views and the opinions expressed in this work are however those of the authors only and do not necessarily reflect those of the European Union or the European Research Council (ERC). Neither the European Union nor the granting authority can be held responsible for them. Moreover, we gratefully acknowledge the Cluster of Excellence EXC 2044-390685587, Mathematics Münster: Dynamics-Geometry-Structure funded by the Deutsche Forschungsgemeinschaft (DFG, German Research Foundation).
\bibliographystyle{acm}
%alpha
\bibliography{bibfile}

\end{document}